\documentclass[10pt,journal,compsoc]{IEEEtran}
\usepackage{indentfirst}
\usepackage{booktabs} 
\usepackage{xcolor}
\definecolor{softblue}{RGB}{100,149,237} 
\usepackage{threeparttable}
\usepackage{ulem} 
\usepackage{bm}
\usepackage{graphicx}
\usepackage{xcolor}
\usepackage[colorlinks=true,
            linkcolor=blue!40,
            citecolor=blue!40,
            urlcolor=blue!40]{hyperref}
\usepackage{multirow}

\usepackage{graphicx}  
\usepackage{makecell}

\usepackage[nottoc]{tocbibind} 

\usepackage{enumitem} 

\usepackage{dblfloatfix} 
\usepackage{cuted}       
\usepackage{balance}     

\usepackage{placeins}
\usepackage{float}

\ifCLASSOPTIONcompsoc
  \usepackage[nocompress]{cite}
\else
  \usepackage{cite}
\fi
\usepackage{etoolbox}
\patchcmd{\thebibliography}{\sloppy}{\sloppy\raggedright}{}{}
\makeatletter
\renewcommand{\@biblabel}[1]{[#1]}
\renewcommand{\emph}[1]{\textit{#1}}
\makeatother

\ifCLASSINFOpdf
\else
\fi
\usepackage{amsmath}
\usepackage{amssymb}
\usepackage{lscape}
\usepackage{adjustbox}

\ifCLASSOPTIONcompsoc
 \usepackage[caption=false,font=footnotesize,labelfont=sf,textfont=sf]{subfig}
\else
 \usepackage[caption=false,font=footnotesize]{subfig}
\fi
\begin{document}
%
\title{STELLA: Guiding Large Language Models for Time Series Forecasting with Semantic Abstractions}


\author{Junjie Fan,~Hongye Zhao,~
        Linduo Wei,~Jiayu Rao,~
        ~Guijia Li,~Jiaxin Yuan,~
        Wenqi Xu, and~Yong Qi$^*$

\IEEEcompsocitemizethanks{
\IEEEcompsocthanksitem J. Fan and Y. Qi are with the School of Intellectual Property, Nanjing University of Science and Technology, Xiaolingwei 200, Nanjing, 210094, China. 
Email: \texttt{junjiefan.edu@gmail.com}, \texttt{qyong@njust.edu.cn}

\IEEEcompsocthanksitem H. Zhao, L. Wei, and J. Yuan are with the School of Economics and Management, Nanjing University of Science and Technology, Xiaolingwei 200, Nanjing, 210094, China. 
Email: \texttt{323107010199@njust.edu.cn}, \texttt{weilinduo@njust.edu.cn}, \texttt{318107010178@njust.edu.cn}

\IEEEcompsocthanksitem J. Rao, G. Li, and W. Xu are with the School of Computer Science and Engineering, Nanjing University of Science and Technology, Xiaolingwei 200, Nanjing, 210094, China. 
Email: \texttt{jiayurao@njust.edu.cn}, \texttt{124106010774@njust.edu.cn}, \texttt{546153140@qq.com}

\IEEEcompsocthanksitem Yong Qi is the corresponding author.
}

}

\IEEEtitleabstractindextext{%
\begin{abstract}

Adapting Large Language Models (LLMs) for time series forecasting has recently emerged as a prominent research direction. However, a fundamental deficiency pervades existing methods: they fail to effectively perform \textbf{information enhancement} for the raw series, leaving the formidable reasoning capabilities of LLMs underutilized. Specifically, prevailing prompting strategies often rely on retrieving correlational context from static semantic pools, establishing a \textit{correlational} link rather than a \textit{generative} interpretation of a series' dynamic behavior. Consequently, they fall short of providing the structured \textbf{supplementary information} (offering global context) and \textbf{complementary information} (revealing instance-specific patterns) that are critical for guiding the LLM.

To this end, we propose \textbf{STELLA} (\textbf{S}emantic-\textbf{T}emporal \textbf{A}lignment with \textbf{L}language \textbf{A}bstractions), a novel framework designed to systematically mine, generate, and inject these two key types of information. The core logic of STELLA is its innovative \textbf{dynamic semantic abstraction mechanism}. This mechanism first decouples the input series into its fundamental components (trend, seasonality, and residual). Subsequently, it generatively translates the intrinsic behavioral features of these components (e.g., periodicity, slope) into a set of \textbf{Hierarchical Semantic Anchors}. These anchors comprise a \textbf{Corpus-level Semantic Prior (CSP)} for global supplementary information and a \textbf{Fine-grained Behavioral Prompt (FBP)} for instance-level complementary information. By leveraging these dynamically generated anchors as prefix-prompts, STELLA precisely guides the LLM to model the intrinsic dynamics of the time series.Extensive experiments on eight benchmark datasets demonstrate that the proposed STELLA consistently outperforms state-of-the-art methods in both long- and short-term forecasting tasks, while also showing superior generalization in zero-shot and few-shot settings. Furthermore, our ablation studies and visualization analyses substantiate the necessity and effectiveness of our core design---guidance through dynamically generated semantic anchors.

\end{abstract}

}

\maketitle

\IEEEdisplaynontitleabstractindextext

%
\IEEEpeerreviewmaketitle

\IEEEraisesectionheading{\section{Introduction}\label{sec:introduction}}


%
%
%
%

In recent years, foundation models, particularly Large Language Models (LLMs), have achieved revolutionary breakthroughs in natural language processing~\cite{achiam2023gpt, touvron2023llama, touvron2023llama2}. Their powerful sequence modeling and contextual reasoning capabilities are driving a paradigm shift across numerous related fields, demonstrating immense potential in complex, structured domains such as finance, healthcare, and beyond~\cite{singhal2023large, cui2024survey, li2023large}. Time series forecasting, a critical task for applications like energy management, traffic planning, and meteorology~\cite{gao2020robusttad, liu2023sadi, friedman1962interpolation, bose2017probabilistic, angryk2020multivariate}, is at the forefront of this transformation. At present, the academic community is exploring two primary paths for developing foundation models for time series.

The first path involves building general-purpose foundation models for time series analysis from scratch. Significant efforts have been made in this direction~\cite{wu2022timesnet, garza2023timegpt, rasul2023lag}, with works like TimesNet~\cite{wu2022timesnet} and TimeGPT-1~\cite{garza2023timegpt} attempting to design universal backbones to handle time series from diverse domains. However, this path is fraught with formidable challenges: time series data exhibit a wide variety of formats and domains, and are ubiquitously affected by non-stationarity and concept drift. These factors make it exceedingly difficult to construct a single pre-trained model that can generalize across all scenarios.

The second, and increasingly prominent, path is to \textbf{adapt} existing pre-trained LLMs. Leveraging the robust knowledge base acquired from massive text corpora, LLMs are expected to become proficient at time series analysis with minimal fine-tuning or prompt-based learning~\cite{zhou2023one}. Despite its promise, we argue that current adaptation methods suffer from a series of deep-seated issues. First, many approaches fail to leverage the core intrinsic patterns—such as seasonality, trend, and residuals—that distinguish time series from generic sequential data~\cite{harvey1990forecasting}. Indeed, the backbone architectures and prompting techniques in existing language models are not inherently designed to capture the evolution of these temporal patterns, unlike specialized models such as N-BEATS~\cite{oreshkin2019n} and AutoFormer~\cite{wu2021autoformer}, which treat them as fundamental. This deficiency manifests in several ways. The prevalent practice of applying a patching operation on the raw series~\cite{nie2022time} not only struggles to capture the subtle variations of these key components but also leads to a significant distributional discrepancy between the resulting numerical tokens and the LLM's native text tokens. Second, while prompting has proven to be an effective paradigm for guiding LLMs~\cite{ouyang2022training}, existing strategies remain coarse in their generation of semantic information. For instance, recent work like S²IP-LLM~\cite{pan2024s} attempts to generate soft prompts by aligning time series embeddings with pre-defined ``semantic anchors.'' However, this retrieval-based method, which relies on similarity matching, essentially finds the most relevant tag from a static semantic pool. It establishes a \textit{correlational} context rather than a \textit{generative} interpretation. Consequently, it fails to generate true \textbf{complementary information}—a fine-grained, semantic description dynamically abstracted from the unique behavior of each individual time series instance.

These shortcomings point to a more fundamental problem: the failure to effectively \textbf{augment information} for the raw time series. Multimodal research has long established that a model's powerful performance often stems from the effective fusion of different modalities of information, particularly when the model can leverage both \textbf{supplementary information} (providing independent context) and \textbf{complementary information} (offering different perspectives on the same subject)~\cite{baltruvsaitis2018multimodal}. As noted by \cite{baltruvsaitis2018multimodal}, leveraging complementary information allows for the capture of patterns invisible in a single modality, while building models that effectively utilize supplementary information is even more challenging. We posit that the performance bottleneck of current LLM-based forecasters is rooted in this information-level deficit. This gives rise to the central question of our work: Can we design a unified framework that systematically mines and injects structured supplementary and complementary information from the text modality to fundamentally enhance the LLM's understanding and forecasting of time series?

To answer this question, we propose \textbf{STELLA} (Semantic-Temporal Alignment with Language Abstractions), a novel framework that actively guides an LLM by generating and utilizing \textbf{Hierarchical Semantic Anchors}. The design philosophy of STELLA is that the most potent guidance arises from the synergy of supplementary context and complementary behavioral interpretation, both distilled into semantic anchors that an LLM can directly leverage. The STELLA architecture is meticulously constructed around this philosophy. First, a Neural STL module decouples the input series into its fundamental trend, seasonal, and residual components. Subsequently, our innovative Semantic Anchor Module (SAM) takes center stage. On one hand, it generates a Corpus-level Semantic Prior (CSP) from dataset metadata to provide global, independent \textbf{supplementary information}. On the other hand, for each decomposed component, it computes behavioral features over the entire input window (e.g., overall slope, dominant periodicity), then textualizes and distills them into a compact Fine-grained Behavioral Prompt (FBP). The CSP and FBP jointly constitute the dynamically generated semantic anchors. In this manner, STELLA uses these anchors as \textbf{prefix-prompts} to provide strong contextual indicators for the numerical embeddings, thereby guiding the LLM's reasoning process.

The main contributions of this paper can be summarized as follows:
\begin{itemize}
    \item From the perspective of \textbf{information enhancement}, we are the first to systematically argue that \textbf{supplementary and complementary information} are key to resolving the current bottleneck in LLM-based time series forecasting. Based on this, we establish a novel theoretical direction for the field centered on \textbf{Semantic-Guided} learning.
    
    \item We design and implement \textbf{STELLA}, an innovative, interpretable, generative forecasting model. At its core is a novel \textbf{dynamic semantic abstraction mechanism} that generatively translates the behavioral features of decomposed time series components into hierarchical semantic anchors containing both supplementary and complementary information, achieving precise guidance for the LLM.
    
    \item We conduct extensive experiments on eight real-world benchmark datasets. The results demonstrate that STELLA not only achieves state-of-the-art performance in both long- and short-term forecasting tasks but also exhibits remarkable generalization capabilities in zero-shot and few-shot scenarios, proving its strong transfer potential to unseen domains.
\end{itemize}

\section{Related Work}

\subsection{Deep Models for Time Series Forecasting}

The field of time series forecasting has evolved from traditional statistical methods, such as ARIMA~\cite{box2015time} and Prophet~\cite{taylor2018forecasting}, to a diverse landscape of deep learning architectures. Early deep learning approaches were largely dominated by Recurrent Neural Network (RNN) based models, including LSTMs and GRUs, which were designed to capture auto-regressive temporal dynamics~\cite{qin2017dual, lai2018modeling}. Concurrently, other specialized architectures like Graph Neural Networks (GNNs) were leveraged to explicitly model the dependencies among multiple time series variables~\cite{cao2020spectral, wu2020connecting}.

Subsequently, Transformer-based models~\cite{vaswani2017attention} emerged as a dominant paradigm, largely due to the self-attention mechanism's exceptional ability to model long-range dependencies. A multitude of variants have since been proposed, exploring different architectural innovations. A significant line of work focuses on integrating decomposition principles directly into the architecture to better handle the complex patterns in time series. Models like Autoformer~\cite{wu2021autoformer} introduced an auto-correlation mechanism, while FEDformer~\cite{zhou2022fedformer} enhanced decomposition with frequency-domain analysis, and a recent framework further explored periodicity decoupling~\cite{dai2024periodicity}. Another research direction has focused on more effectively modeling dependencies across both the time and variable dimensions, with notable examples including Crossformer~\cite{zhang2023crossformer} and CARD~\cite{xue2023card}. Other hybrid approaches, such as ETSformer~\cite{woo2022etsformer}, sought to combine the strengths of Transformers with classical exponential smoothing methods. In parallel, PatchTST~\cite{nie2022time} demonstrated that partitioning a time series into patches as input tokens can significantly improve performance, establishing patching as a powerful representation technique.

However, the dominance of these architecturally complex models has been challenged by recent studies showing that simpler models can achieve comparable, if not superior, state-of-the-art performance. This includes both time-domain approaches, such as MLP-based architectures~\cite{zeng2023transformers}, and lightweight frequency-domain models like FITS, which operates with merely 10k parameters~\cite{xu2023fits}. This trend suggests that many sophisticated models, in their singular pursuit of modeling temporal dependencies, may have overlooked the fundamental intrinsic properties of time series. Ultimately, while these specialized deep forecasters, whether architecturally simple or complex, perform well on in-domain benchmarks, they often lack the flexibility and generalization capabilities required to adapt to the diverse and non-stationary nature of real-world data from varying domains. This critical gap highlights the need for a more generalizable paradigm, motivating the exploration of large-scale pre-trained models.

\subsection{Large Language Models for Time Series Forecasting}

To surmount the generalization challenges inherent in conventional deep learning models for time series forecasting, the research community has begun to explore a promising new frontier: leveraging the formidable knowledge transfer and reasoning capabilities of Large Language Models (LLMs)~\cite{jiang2024empowering}. The groundbreaking success of models such as the GPT~\cite{radford2019language} and LLaMA~\cite{touvron2023llama, touvron2023llama2} series in Natural Language Processing (NLP) and Computer Vision (CV) has established a powerful precedent~\cite{raffel2020exploring, brown2020language, ouyang2022training}. This paradigm is predicated on the principle that robust representations learned from vast, diverse corpora can be adapted to elicit substantial performance gains on downstream tasks, offering a novel pathway to address the generalization deficits in time series analysis~\cite{tang2022domain}.

Early explorations of adapting LLMs to the time series domain primarily bifurcated into two approaches. The first, \textit{textualization}, converts numerical series into text sequences for zero-shot inference by an LLM~\cite{xue2023promptcast}. This method, however, is often constrained by the semantic gap between numerical values and natural language tokens. The second, more dominant approach involves tokenizing time series into patches to subsequently fine-tune a pre-trained LLM on this new modality~\cite{zhou2023one, chang2023llm4ts}. While these paradigms validated the feasibility of employing LLMs, they fundamentally treated the model as a generic sequence processor and thus failed to adequately account for the unique characteristics of time series or the inherent modality mismatch~\cite{shin2020autoprompt}.

To bridge this modality gap more effectively, a sophisticated line of research has emerged that leverages prompting mechanisms to connect numerical dynamics with the LLM's semantic space. Prompt tuning, an efficient adaptation method for foundation models~\cite{lester2021power, li2021prefix}, has demonstrated remarkable efficacy beyond its NLP origins. Foundational models like CLIP have shown that textual prompts can achieve state-of-the-art performance in image classification~\cite{radford2021learning}. Concurrently, in continual learning, concepts such as prompt pools~\cite{wang2022learning} and complementary prompts~\cite{wang2022dualprompt} have enabled models to acquire new tasks without rehearsal. These cross-domain successes underscore the pivotal role of well-designed prompts in unlocking the full potential of pre-trained models.

Inspired by these advancements, a suite of prompt-based methods has been developed for time series. A significant research direction aligns time series representations with text prototypes, as seen in TEST~\cite{sun2023test}, Time-LLM~\cite{jin2023time}, and other recent works~\cite{li2024frozen}. Others incorporate time series decomposition and retrieval-based prompts to handle non-stationarity, such as TEMPO~\cite{cao2023tempo}. Notably, S²IP-LLM~\cite{pan2024s} aligns time series embeddings with pre-defined ``semantic anchors'' retrieved from the LLM's word embedding space. Despite representing significant progress, these methods share a fundamental limitation: they rely on \textbf{retrieving or matching pre-defined, static textual descriptions}. This establishes a merely \textit{correlational} context rather than a \textit{generative} understanding of a series' intrinsic dynamics. Consequently, they struggle to capture and leverage the fine-grained, \textbf{instance-specific complementary information} that is critical for nuanced forecasting.

In stark contrast, our work introduces the principle of \textbf{generative semantic guidance}. Our proposed \textbf{STELLA model}, through a \textbf{dynamic semantic abstraction mechanism}, \textbf{endogenously generates} hierarchical anchors from a series' intrinsic properties, thereby explicitly disentangling global \textbf{supplementary information} (via the Causal Semantic Prompt, CSP) and instance-specific \textbf{complementary information} (via the Fluctuation-aware Behavioral Prompt, FBP). This approach forges a more intrinsic and robust alignment between the numerical dynamics of time series and the semantic reasoning of LLMs, aiming to systematically address the key limitations of prior art.

\section{Methodology}
\begin{figure*}[!t]
  \centering
  \includegraphics[width=0.8\textwidth]{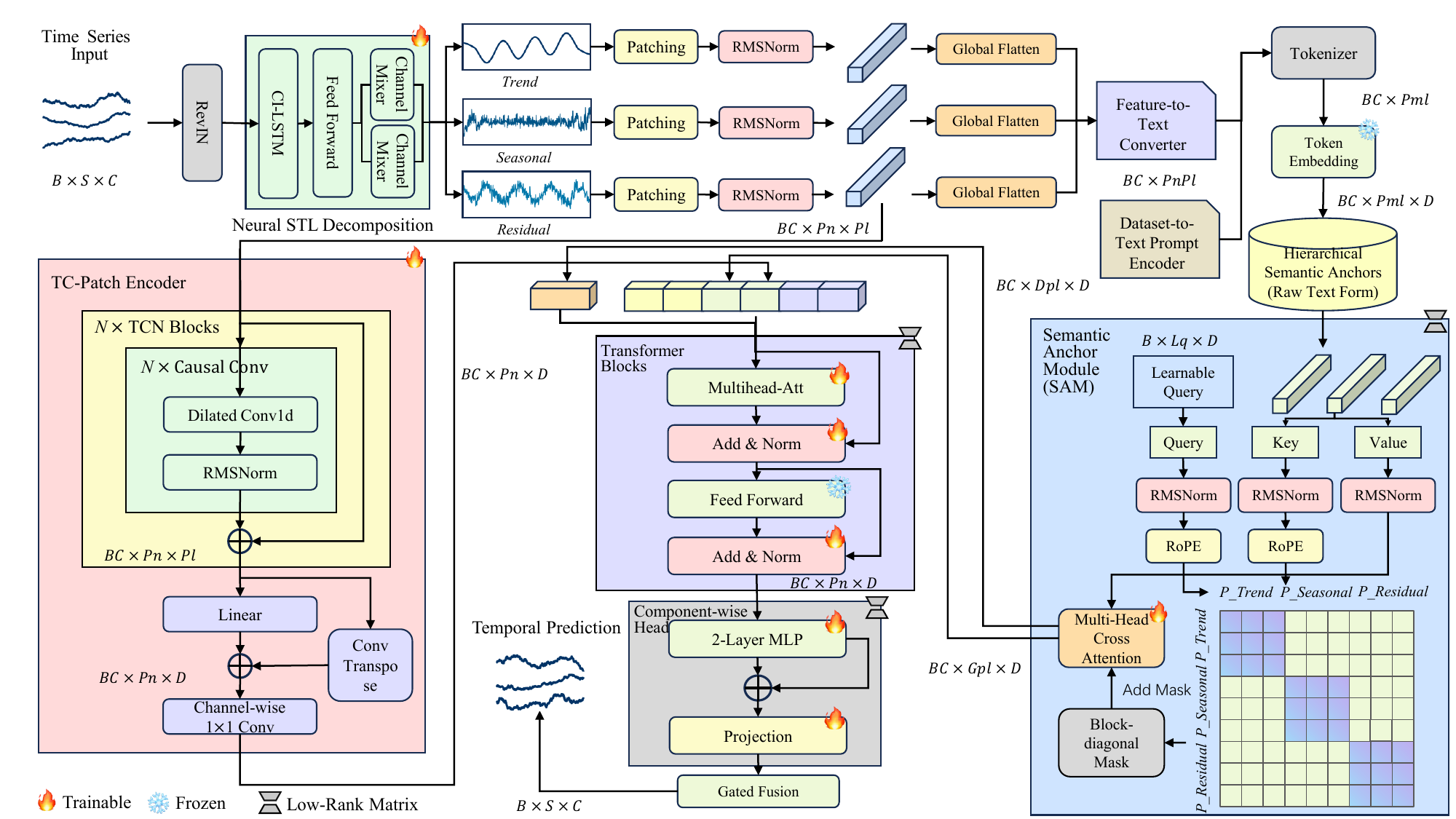} 
  \caption{An overview of the STELLA framework. STELLA first decomposes the input series and then utilizes a dual-path architecture: a TC-Patch Encoder generates numerical embeddings, while a Semantic Anchor Module generates hierarchical prompts (CSP and FBP) to guide a frozen LLM backbone for forecasting.}
  \label{fig:overview-image}
\end{figure*}

\textbf{Overview:} As shown in Fig.~\ref{fig:overview-image}, the \textbf{STELLA} framework operates in three stages. First, in the structural representation stage, the input series $\mathbf{X}$ is normalized and factorized into temporal components $(\mathbf{T}, \mathbf{S}, \mathbf{R})$ by a Neural STL module. A dual-path architecture then processes each component: one path yields a numerical embedding $\mathbf{E}^{(k)}$ via a TC-Patch encoder, while the parallel path utilizes our \textbf{Semantic Anchor Module (SAM)} to abstract behavioral signatures into compact, learnable prompt vectors. These vectors serve as two potent semantic anchors: a global Corpus-level Semantic Prior ($\mathbf{P}_{\mathrm{CSP}}$) and an instance-specific Fine-grained Behavioral Prompt ($\mathbf{P}^{(k)}_{\mathrm{FBP}}$). Second, these anchors are prepended to their corresponding embeddings to form a unified input sequence $\mathbf{X}_{\mathrm{in}}$. This sequence is fed to the LLM backbone, where causal self-attention aligns the guiding anchors with the numerical representations. Finally, a synthesis stage decodes component-wise forecasts $\widehat{\mathbf{Y}}^{(k)}$ from the LLM's latent outputs, which are then dynamically integrated by a gated fusion head to render the final forecast $\hat{\mathbf{Y}}$.

\subsection{Problem Definition}
The core objective of multivariate time series forecasting is to accurately predict the future values of multiple variables over the next $H$ time steps, based on the historical observations from the preceding $S$ steps.
Formally, let $\mathbf{x}_t \in \mathbb{R}^{C}$ denote the $C$-dimensional observation at time $t$, and let 
$\mathbf{X}_{t-S:t-1} = [\mathbf{x}_{t-S}, \ldots, \mathbf{x}_{t-1}] \in \mathbb{R}^{S \times C}$ be the input matrix.
We define a forecasting function $F_\Phi$ with learnable parameters $\Phi$ such that
\begin{equation}
\hat{\mathbf{Y}}_{t:t+H-1} \;=\; F_\Phi\!\big(\mathbf{X}_{t-S:t-1}\big),
\tag{1}
\end{equation}
where $\hat{\mathbf{Y}}_{t:t+H-1} = [\hat{\mathbf{x}}_{t}, \ldots, \hat{\mathbf{x}}_{t+H-1}] \in \mathbb{R}^{H \times C}$ are the predictions of the next $H$ steps. 
In training, we use mini-batches arranged as $\mathbf{X}\in\mathbb{R}^{B\times S\times C}$ and targets $\mathbf{Y}\in\mathbb{R}^{B\times H\times C}$.

\subsection{Channel Strategy: Mixing vs.~Independence}
\label{sec:channel_strategy}

Modeling cross-channel interactions in multivariate time-series forecasting (MTSF) is non-trivial, especially when the channel cardinality $C$ varies across datasets and variable semantics differ across domains.
The central question is how to represent and exploit inter-variable relations without sacrificing robustness under domain shift.

A widely used design space separates approaches by how they handle cross-channel information: \emph{channel-independent (CI)} and \emph{channel-mixing (CM)}.
CI encodes each variable separately and, by construction, does not model inter-channel dependencies.
Although such simplification may appear suboptimal when complex cross-variable relations drive predictability, empirical studies have reported \emph{strong gains for channel independence} in practice, indicating robustness to changes in $C$ and reduced cross-domain interference~\cite{zeng2023transformers,nie2022time}.
In contrast, CM jointly embeds all variables so that each channel is informed by the others.
By integrating interdependencies among multiple time-series variables, CM often yields richer and more accurate representations of underlying patterns and trends within a fixed domain~\cite{wu2022timesnet,wu2021autoformer,zhou2021informer}.
However, implementations that rely on a single shared embedding to mediate inter-channel relations may be insufficient to capture mutual information and can sometimes introduce noise; robustness under \emph{domain shift} (e.g., varying $C$ or differing variable semantics) may also be limited~\cite{wang2025csformer}.

\textbf{Our stance.}
To reconcile these trade-offs, we adopt an \emph{independence--then--mixing} regime.
Within the STL module, temporal-feature extraction is performed in a \emph{channel-independent} manner to obtain per-variable trend/seasonal/residual factors, which are then mapped to tokens; the subsequent decoder-only Transformer \emph{mixes} channels via self-attention to model cross-variable dependencies.
In summary, this CI--then--CM pipeline effectively leverages cross-variable information while fully exploiting temporal structure.
This design balances robustness to varying $C$ and domain shift with the expressive power needed to model inter-variable structure.
See Sec.~\ref{sec:neural_stl} for implementation details.

\subsection{Normalization Design: Input ReVIN \& Backbone RMSNorm}
\label{sec:norm}

In real-world deployments, exogenous factors often induce non-stationary distribution shift across time and variables, biasing models trained under mismatched distributions.
To mitigate such shifts while preserving exact invertibility and scale information, we adopt \emph{reversible instance normalization} (ReVIN)~\cite{kim2021reversible} at the input.

Given a mini-batch $\mathbf{X}\!\in\!\mathbb{R}^{B\times S\times C}$, we compute per-instance, per-channel statistics over the $S$ time steps:
\begin{equation}
\boldsymbol{\mu}=\mathrm{Mean}_{S}(\mathbf{X}),\quad
\boldsymbol{\sigma}=\sqrt{\mathrm{Var}_{S}(\mathbf{X})}
\;\in\;\mathbb{R}^{B\times 1\times C}.
\tag{2}
\end{equation}
With learnable affine parameters $\boldsymbol{\gamma},\boldsymbol{\beta}\!\in\!\mathbb{R}^{1\times 1\times C}$ (broadcast over time), ReVIN maps each series to a normalized space:
\begin{equation}
\tilde{\mathbf{X}}
=\boldsymbol{\gamma}\odot
\frac{\mathbf{X}-\boldsymbol{\mu}}{\boldsymbol{\sigma}+\epsilon}
+\boldsymbol{\beta},
\quad
\tilde{\mathbf{X}}\in\mathbb{R}^{B\times S\times C},
\tag{3}
\end{equation}
where $\epsilon\!>\!0$ prevents division by zero and $\odot$ denotes element-wise product.
This instance- and channel-wise transformation reduces artificial disparities among univariate variables and improves robustness under domain shift.

The forecasting backbone operates on $\tilde{\mathbf{X}}$ to produce normalized predictions $\hat{\mathbf{Y}}\in\mathbb{R}^{B\times H\times C}$, which are then returned to the original scale via the exact inverse:
\begin{equation}
\breve{\mathbf{Y}}
=\boldsymbol{\sigma}\odot
\frac{\hat{\mathbf{Y}}-\boldsymbol{\beta}}{\boldsymbol{\gamma}+\epsilon}
+\boldsymbol{\mu},
\quad
\breve{\mathbf{Y}}\in\mathbb{R}^{B\times H\times C}.
\label{eq:revindenorm}
\tag{4}
\end{equation}
Here $\boldsymbol{\mu},\boldsymbol{\sigma}$ are computed from the corresponding input window $\mathbf{X}$ for each instance and channel, and are broadcast across the $H$ forecast steps.
ReVIN thus interfaces cleanly with subsequent tokenization and decoding, offering domain-shift robustness without discarding scale information.

Additionally, we employ RMSNorm in a pre-norm configuration across all Transformer modules  and the MLP decoder/head, aligning with LLaMA-style designs and improving training stability under varying sequence lengths and batch sizes.
For a hidden vector $\mathbf{h}\!\in\!\mathbb{R}^{d}$ with trainable gain $\mathbf{g}\!\in\!\mathbb{R}^{d}$,
\begin{equation}
\mathrm{RMSNorm}(\mathbf{h})
=\frac{\mathbf{h}}{\sqrt{\tfrac{1}{d}\sum_{i=1}^{d}h_i^2+\epsilon}}\odot \mathbf{g}.
\tag{5}
\end{equation}
This root-mean-square normalization (without centering) fits the pre-norm residual design we adopt across all Transformer blocks and heads.

\subsection{Neural STL Decomposition}
\label{sec:neural_stl}
An \emph{a priori} segregation of multivariate time series into trend, seasonal, and residual strata is a long-established principle for enhancing model-specificity and data efficiency \cite{cleveland1990stl}. This principle becomes particularly consequential within Transformer architectures. Given that self-attention layers approximate orthogonality-constrained operators, they are, in principle, ill-suited to separate non-orthogonal temporal constituents \cite{cao2023tempo}. Classical approaches typically address this with hand-crafted filters, such as centered moving averages, to derive a trend surrogate. In lieu of such predefined operators, we posit that an explicit, learnable, and pre-attentive factorization is more effective. Accordingly, our module, inspired by basis expansion methods \cite{oreshkin2019n}, parameterizes this factorization over a fixed local window. This prior not only stabilizes the decomposition but also aligns its granularity with our downstream patching mechanism. This entire decomposition is performed on the normalized input, and only thereafter do we proceed to the tokenization and component-specific encoding stages.

Given the ReVIN-normalized input $\tilde{\mathbf{X}}\!\in\!\mathbb{R}^{B\times S\times C}$, our module actualizes an additive decomposition:
\begin{equation}
\label{eq:stl_decomposition}
\tilde{\mathbf{X}}
\;=\;
\mathbf{T}
\;+\;
\mathbf{S}
\;+\;
\mathbf{R},
\end{equation}
where $\mathbf{T} \in \mathbb{R}^{B\times S\times C}$ constitutes the low-frequency trend, $\mathbf{S} \in \mathbb{R}^{B\times S\times C}$ encapsulates recurrent periodicities, and $\mathbf{R} \in \mathbb{R}^{B\times S\times C}$ absorbs the irregular, high-frequency complement. We implement this decomposition via a \textbf{Channel-Independent (CI) to Channel-Mixing (CM)} paradigm, which first processes channel-wise temporal dynamics in isolation before modeling their cross-channel interdependencies.

The CI stage generates a "proto-trend" tensor $\mathbf{Z}$, which serves as an intermediate representation. For each channel $c \in \{1, \ldots, C\}$, a dedicated LSTM encoder produces temporally-contextualized embeddings from its local history. A shared linear contraction and a non-linear activation $\phi$ then distill these embeddings into a univariate feature sequence:
\begin{equation}
\label{eq:ci_encoder}
\begin{gathered}
\mathbf{H}^{(c)}=\mathrm{LSTM}_c\!\big(\tilde{\mathbf{X}}_{:,:,c}\big)\in\mathbb{R}^{B\times S\times d_h}, \\
\mathbf{z}^{(c)}=\phi\!\big(\mathbf{H}^{(c)}\mathbf{w}_\ell+\mathbf{b}_\ell\big)\in\mathbb{R}^{B\times S\times 1}.
\end{gathered}
\end{equation}
Here, the contraction parameters $\mathbf{w}_\ell \in \mathbb{R}^{d_h}$ and $\mathbf{b}_\ell \in \mathbb{R}$ are shared across all channels. The final proto-trend tensor is formed by stacking these sequences: $\mathbf{Z}=\mathrm{Stack}_{c=1}^{C}(\mathbf{z}^{(c)})\in\mathbb{R}^{B\times S\times C}$.

The CM stage synthesizes the final components. The trend component $\mathbf{T}$ is effected by a time-shared channel mixer, $\mathrm{MLP}_T^{\text{chan}}$, which realizes an instantaneous, non-linear coupling across the channel dimension of $\mathbf{Z}$ at each time step:
\begin{equation}
\label{eq:trend_synthesis}
\mathbf{T}_{b,t,:}
\;=\;
\mathrm{MLP}_T^{\text{chan}}\!\big(\mathbf{Z}_{b,t,:}\big), \quad \mathrm{MLP}_T^{\text{chan}}:\mathbb{R}^{C}\!\to\!\mathbb{R}^{C}.
\end{equation}
The seasonal component $\mathbf{S}$ is generated analogously by a distinct mixer, $\mathrm{MLP}_S^{\text{chan}}$, operating on the detrended series:
\begin{equation}
\label{eq:season_synthesis}
\begin{gathered}
\mathbf{S}_{\mathrm{pre}} = \tilde{\mathbf{X}}-\mathbf{T}, \\
\mathbf{S}_{b,t,:} = \mathrm{MLP}_S^{\text{chan}}\!\big(\mathbf{S}_{\mathrm{pre},\,b,t,:}\big).
\end{gathered}
\end{equation}
Finally, the residual component $\mathbf{R}$ is determined by enforcing additive closure on the normalized scale:
\begin{equation}
\label{eq:residual_completion}
\mathbf{R}
\;=\;
\tilde{\mathbf{X}}
\;-\;
\mathbf{T}
\;-\;
\mathbf{S}.
\end{equation}

This CI$\to$CM staging—wherein channel-independent recurrence generates temporally-contextualized embeddings before time-shared mixers realize instantaneous cross-channel couplings—is a deliberate architectural choice. It preserves essential symmetries such as temporal locality and permutation invariance across the time axis. Critically, this explicit, pre-attentive factorization offloads the formidable task of disentangling non-orthogonal temporal strata from the self-attention layers. By operationally decomposing the main task into simpler, more tractable sub-problems, it allows the subsequent Transformer blocks to focus their capacity on modeling the dynamics of cleaner, pre-processed component series.

\subsection{Temporal Convolutional Patch Encoder (TC--Patch)}

This work presents the Temporal Convolutional Patch Encoder (TC-Patch), a novel architecture for learning expressive, temporally-localized representations from sequential data. Drawing inspiration from Vision Transformers \cite{dosovitskiy2020image} and PatchTST \cite{nie2022time}, we employ a patching strategy to segment adjacent time steps into patches that serve as input tokens. This approach serves the dual purpose of capturing local contextual information while substantially reducing the sequence length, thereby mitigating the computational complexity inherent to long sequences. Crucially, patching is applied subsequent to the Neural STL decomposition—a pivotal design choice that enables the encoder to model the distinct local dynamics of the extracted trend, seasonality, and residual components in isolation, rather than attempting to disentangle these confounded patterns from the raw series.

For each component $\mathbf{Z}\in\{\mathbf{T},\mathbf{S},\mathbf{R}\}$, we first partition the temporal axis into $P_n$ non-overlapping patches of length $P_\ell$ with a stride of $s_p$, where $P_n = \lfloor (S-P_\ell)/s_p\rfloor + 1$. This procedure reshapes the input into a 4D tensor $\mathbf{Z}^{\mathrm{patch}} \in \mathbb{R}^{B\times C\times P_n\times P_\ell}$. Subsequently, patches are flattened and normalized via RMSNorm to form the initial token sequences:
\begin{equation}
\label{eq:tokenization}
\mathbf{Z}^{\mathrm{tok}} = \mathrm{RMSNorm}\!\left( \mathrm{Reshape}_{(BC)\times P_n \times P_\ell} \big(\mathbf{Z}^{\mathrm{patch}}\big) \right).
\end{equation}
The TC-Patch architecture then processes these tokens through two main stages: a TCN-based backbone for intra-patch feature extraction and a projection head for dimensional alignment.

\textbf{Intra-Patch TCN Backbone.}
The backbone of our encoder is a Temporal Convolutional Network (TCN) \cite{bai2018empirical}, a class of architectures proven effective for sequence modeling. Its efficacy is predicated on two core principles: causal and dilated convolutions. Causal convolutions enforce strict temporal causality, ensuring an output at time step $t$ is exclusively dependent on inputs at or before $t$, thereby precluding information leakage from future time steps. Dilated convolutions introduce a fixed step between adjacent kernel taps, a mechanism that exponentially expands the receptive field with network depth without a commensurate increase in computational cost. A dilated causal convolution on a 1D sequence $\mathbf{X}$ with a filter $\mathbf{W}$ of size $K$ and dilation $d$ is defined as:
$$(\mathbf{X} *_{d} \mathbf{W})_t = \sum_{k=0}^{K-1} \mathbf{W}_k \cdot \mathbf{X}_{t - d \cdot k}.$$

Our core contribution is the novel manner in which we apply the TCN backbone. Specifically, we remap the axes of the 3D tensor from Eq.~\ref{eq:tokenization}, treating the composite batch-channel dimension ($BC$) as the batch dimension, the number of patches ($P_n$) as the channel dimension, and the patch length ($P_\ell$) as the sequence length. This remapping strategy not only confers a significant computational advantage by enabling parallel processing, but more critically, its primary role is to transform the time series into a structured token sequence. This output format is designed to align seamlessly with the autoregressive next-token prediction paradigm of a subsequent Large Language Model (LLM).

The TCN backbone comprises $L_{\mathrm{tcn}}$ residual layers, each containing $M$ sub-blocks. A residual connection \cite{he2016deep} envelops a sequence of \emph{Conv1D(causal, dilated) $\rightarrow$ RMSNorm $\rightarrow$ Activation $\phi$ $\rightarrow$ Dropout}. The dilation rate increases exponentially with depth ($d_m = 2^{m-1}$), yielding a vast receptive field. For layer $l$ and sub-block $m$, the operations are:

\begin{equation}
\begin{aligned}
\mathbf{Z}^{(0)} &= \mathbf{Z}^{\mathrm{tok}}, \\
\mathbf{U}^{(l,m)} &= \mathrm{Dropout}_{p}\!\Big(
\phi\!\big(\mathrm{RMSNorm}( \\
&\quad\quad \mathrm{Conv1D}_{\mathrm{causal},\,d_m}(\mathbf{Z}^{(l,m-1)}))\big)\Big), \\
\mathbf{Z}^{(l,m)} &= \mathbf{Z}^{(l,m-1)} + \mathbf{U}^{(l,m)}, \\
&\quad m=1,\dots,M, \\
\mathbf{H} &= \mathbf{Z}^{(L_{\mathrm{tcn}}, M)} \in \mathbb{R}^{(BC)\times P_n \times P_\ell}.
\end{aligned}
\tag{7}
\end{equation}

We apply weight normalization \cite{salimans2016weight} to all 1D convolutional layers and initialize their weights using Kaiming initialization \cite{he2015delving}. Standard dropout is applied for regularization. The backbone strictly preserves the sequence length $P_\ell$.

\textbf{Projection Head.}
Following the TCN backbone, a projection head maps the learned intra-patch features of length $P_\ell$ to the model's primary hidden dimension $D$. This is a requisite step for aligning the feature dimension with subsequent architectural components. The tensor $\mathbf{H}$ is first reshaped into $\tilde{\mathbf{H}} \in \mathbb{R}^{(BC \cdot P_n) \times P_\ell}$. A position-wise linear layer then projects the feature dimension from $P_\ell$ to $D$. It incorporates a residual connection wherein a standard 1D transposed convolution ($\mathrm{TConv}$) upsamples the original signal to match the target dimension $D$.
\begin{equation}
\begin{aligned}
\hat{\mathbf{H}} \;&=\; \phi(\tilde{\mathbf{H}})\,\mathbf{W}^{\top},\quad
\mathbf{W}\in\mathbb{R}^{D\times P_\ell}, \\
\bar{\mathbf{H}} \;&=\; \hat{\mathbf{H}}
\;+\;
\mathrm{TConv}_{P_\ell\rightarrow D}(\tilde{\mathbf{H}}).
\end{aligned}
\tag{8}
\end{equation}

Finally, we apply a $1 \times 1$ depthwise convolution ($\mathrm{DWConv}_{1\times 1}$). The choice of a depthwise convolution is deliberate: by operating channel-wise (i.e., patch-wise in our remapped context), it preserves the representational independence of each patch. This allows for per-patch feature recalibration across the hidden dimension $D$ without inducing cross-patch information leakage. The final output is:
\begin{equation}
\begin{aligned}
\mathbf{E} \;=\;& \mathrm{DWConv}_{1\times 1}\!
\Big(\mathrm{Reshape}_{(BC)\times P_n \times D}(\bar{\mathbf{H}})\Big) \\
&\;\in\; \mathbb{R}^{(BC)\times P_n \times D}.
\end{aligned}
\tag{9}
\end{equation}

\textbf{Component-Factored Instantiation.}
To capture the distinct characteristics inherent in the decomposed time series, we instantiate three independent TC-Patch encoders, one for each of the trend ($\mathbf{T}$), seasonality ($\mathbf{S}$), and residual ($\mathbf{R}$) components. This yields three distinct sets of embeddings: $\mathbf{E}^{T}$, $\mathbf{E}^{S}$, and $\mathbf{E}^{R}$. Crucially, these encoders do not share weights; each possesses a unique set of parameters ($\Theta_T, \Theta_S, \Theta_R$), thereby enabling specialized feature extraction tailored to the specific dynamics of each component, culminating in a more expressive and disentangled final representation.

\subsection{Semantic Anchor Module (SAM) for Cross-Modal Alignment}
\label{sec:cross_modal_alignment}

The component-specific embeddings ($\mathbf{E}^{T}$, $\mathbf{E}^{S}$, $\mathbf{E}^{R}$) provide a disentangled numerical representation. However, to effectively guide a Large Language Model (LLM), we must bridge the modality gap by injecting task-specific knowledge. Prompting offers a powerful paradigm for this, creating a structured framework that aligns the model's capabilities with desired objectives. Early research predominantly focused on enhancing pre-trained models by fine-tuning with fixed "hard" prompts \cite{brown2020language}. Subsequently, a spectrum of techniques has emerged, including "semi-soft" prompts that transform a fixed text template into a learnable vector to balance interpretability and adaptability \cite{cao2023tempo}.

Building on these insights, we introduce a sophisticated hierarchical mechanism for semantic injection. The architectural principle of employing complementary prompts—one for general and one for specific guidance—finds parallels in domains like continual learning, where prompts are used to switch between discrete tasks \cite{wang2022dualprompt}. Our work, however, introduces a novel formulation for the distinct challenge of time series forecasting. We pair a global prior with a fully dynamic prompt conditioned on the unique behavioral signature of each individual \textbf{instance}. This hierarchical mechanism is implemented within our proposed \textbf{Semantic Anchor Module (SAM)}. The SAM is responsible for generating two distinct semantic anchors: a \textbf{Fine-grained Behavioral Prompt (FBP)} and a \textbf{Corpus-level Semantic Prior (CSP)}.

\textbf{Fine-grained Behavioral Prompt (FBP). } The FBP furnishes the model with a potent, data-driven inductive bias that captures the unique behavior of each time series component. The generation process involves two stages: textualization of behavior and learnable abstraction.

First, for each component series $\mathbf{Z}_{b,c, :}^{(k)} \in \mathbb{R}^S$, we generate a "behavioral signature" by transforming its raw numerical properties into stable, textual cues. This feature set is defined as:

\begin{align}
\mathrm{Feature}_{b,c}^{(k)} 
&= \Big[
    \underbrace{\min, \max, \dots, \mathrm{var}}_{\text{Standard Stats}}, \notag\\
&\quad\underbrace{\mathrm{Categorical\text{-}Trend}, \mathrm{Descriptive\text{-}Lags}}_{\text{Textual Cues}}
\Big].
\end{align}

The textual cues are derived by first computing a numerical metric, which is then converted into a human-readable string. For instance, a continuous linear regression slope is discretized into one of five categories (e.g., `stable`, `slightly increasing`), and the top-$K$ autocorrelation lags are formatted into descriptive sentences. The Feature-to-Text process yields a rich textual description of the component's behavior, which is used to populate a template $\mathcal{T}^{(k)}$.

Second, this textual description is distilled into a compact, learnable prompt. The text is passed through the LLM's frozen tokenizer and embedding layers to produce key-value pairs ($\mathbf{K}^{(k)}, \mathbf{V}^{(k)}$). A cross-attention module then uses a set of learnable queries ($\mathbf{Q}^{(k)}$) to attend to this text and abstract the most salient behavioral information into the final prompt vector:
\begin{equation}
\begin{aligned}
\mathbf{P}^{(k)}_{\mathrm{FBP}} &= \mathrm{CA}\big(\mathbf{Q}^{(k)},\mathbf{K}^{(k)},\mathbf{V}^{(k)};\mathbf{M}_{\mathrm{BD}}\big) \in \mathbb{R}^{(BC)\times G_{pl}\times D}.
\end{aligned}
\tag{10}
\end{equation}
A block-diagonal mask ($\mathbf{M}_{\mathrm{BD}}$) preserves component-wise disentanglement. This entire process ensures the FBP is inherently adaptive to non-stationary characteristics and distributional shifts \cite{fan2023dish}. The resulting prompt is then prepended to its corresponding token sequence:
\begin{equation}
\widetilde{\mathbf{E}}^{(k)}
=\big[\,\mathbf{P}^{(k)}_{\mathrm{FBP}}\,;\,\mathbf{E}^{(k)}\,\big]
\in \mathbb{R}^{(BC)\times (G_{pl}+P_n)\times D}.
\tag{11}
\end{equation}

\textbf{Corpus-level Semantic Prior (CSP).} Complementing the adaptive FBP, the CSP provides a stable, high-level inductive bias that contextualizes the entire dataset. It is realized as a textual description, $\mathcal{T}_{\mathrm{data}}$, containing global metadata (e.g., data domain, sampling frequency). We treat this information as a semantic \textbf{prior}, offering the interpretable guidance of a "hard" prompt but encoded into a flexible vector representation for the model. A dedicated learnable query, $\mathbf{Q}_{\mathrm{data}}$, distills this corpus-level information into a single, dense vector via cross-attention:
\begin{equation}
\begin{aligned}
\mathbf{P}_{\mathrm{CSP}}
&= \mathrm{CA}\!\big(\mathbf{Q}_{\mathrm{data}},\mathbf{K}_{\mathrm{data}},
\mathbf{V}_{\mathrm{data}};\mathbf{0}\big) \in \mathbb{R}^{B\times G_{pl}^{\mathrm{data}}\times D}.
\end{aligned}
\tag{12}
\end{equation}

\textbf{Final Input Sequence Assembly.} The final input stream is constructed by hierarchically arranging the semantic prior and the behaviorally-prompted tokens. The CSP is broadcast and placed at the head of the sequence, followed by each component's token sequence, which has been prepended with its corresponding FBP.
\begin{equation}
\begin{aligned}
\mathbf{X}_{\mathrm{in}}
=\Big[ &\underbrace{\mathbf{P}_{\mathrm{CSP}}^{\uparrow C}}_{\text{Corpus Prior}}\ ;\
\underbrace{\mathbf{P}^{(T)}_{\mathrm{FBP}};\mathbf{E}^{(T)}}_{\text{Trend Behavior}}\ ; \\
&\underbrace{\mathbf{P}^{(S)}_{\mathrm{FBP}};\mathbf{E}^{(S)}}_{\text{Season Behavior}}\ ;\
\underbrace{\mathbf{P}^{(R)}_{\mathrm{FBP}};\mathbf{E}^{(R)}}_{\text{Residual Behavior}}\ \Big].
\end{aligned}
\tag{13}
\end{equation}

This structured assembly provides the model with a rich, multi-layered inductive bias, combining global context with adaptive, instance-specific behavioral cues.

\subsection{Prompt-Conditioned Transformer Backbone and Gated Decoding}
\label{sec:backbone_decoding}

\textbf{Transformer Backbone.}
The core of our model is a decoder-only Transformer backbone with $L$ layers, adhering to the LLaMA-style architecture. We select this architecture due to its autoregressive nature and causal self-attention mechanism, which are inherently aligned with the temporal ordering of time series data. We adopt a unified backbone for the entire input sequence to encourage the model to learn potential cross-component interactions within a shared representational space. A key design choice is the ordering of the input stream: by placing the global Corpus-level Semantic Prior (CSP) before the time series tokens, the causal attention mechanism ensures that the model is conditioned on high-level domain context as it processes the specific behavioral characteristics of the instance. To preserve the rich prior knowledge of the pre-trained model and enhance training efficiency, we employ the parameter-efficient fine-tuning strategy of Low-Rank Adaptation (LoRA) \cite{hu2022lora}. The input token stream $\mathbf{X}_{\mathrm{in}}$ is processed by each layer as follows:
\begin{equation}
\begin{aligned}
\mathbf{H}^{(l)}_{\mathrm{msa}}
&= \mathbf{X}^{(l-1)} \;+\;
\mathrm{MSA}\!\big(\mathrm{RMSNorm}(\mathbf{X}^{(l-1)})\big),\\
\mathbf{X}^{(l)}
&= \mathbf{H}^{(l)}_{\mathrm{msa}} \;+\;
\mathrm{FFN}\!\big(\mathrm{RMSNorm}(\mathbf{H}^{(l)}_{\mathrm{msa}})\big).
\end{aligned}
\tag{14}
\end{equation}
After the final layer, we discard all prompt tokens, retaining only the output tokens corresponding to the time series patches. This yields three distinct component representations: $\mathbf{Z}^{(T)},\mathbf{Z}^{(S)},\mathbf{Z}^{(R)}\in\mathbb{R}^{(BC)\times P_n\times D}$.

\textbf{Component-wise Decoding Head.}
From these component representations, the next step is to generate the final forecast. To ensure robustness against the error accumulation common in iterative forecasting, we adopt a \textbf{direct forecasting} paradigm, where the entire forecast horizon is generated in a single forward pass \cite{nie2022time,wu2022timesnet,zeng2023transformers}. This, however, presents a challenge: how to effectively decode a fixed-length numerical forecast from a variable-length sequence of high-dimensional patch tokens. While a simple linear projection is a common baseline, it may be insufficient to model the complex relationships across the token sequence \cite{liu2024unitime}.

To this end, we design an expressive decoding head that posits a more accurate forecast can be achieved by \textbf{globally aggregating} information from the entire sequence of patch tokens after a stable non-linear transformation. For each component, we employ a two-layer residual MLP, shared across patches, before a final aggregation and projection layer:
\begin{equation}
\begin{aligned}
\widehat{\mathbf{Y}}^{(k)}
&= \mathrm{Head}^{(k)}(\mathbf{Z}^{(k)}) = \mathrm{Proj}_H\!\Big( \mathbf{Z}^{(k)} \;+\; \mathrm{MLP}_{\text{res}}(\mathbf{Z}^{(k)}) \Big),
\end{aligned}
\tag{15}
\end{equation}
where the final projection layer, $\mathrm{Proj}_H(\cdot)$, is responsible for synthesizing information across all $P_n$ patch tokens and mapping the resulting features to the desired forecast horizon $H$. This process yields the component-wise forecasts $\widehat{\mathbf{Y}}^{(k)}\in\mathbb{R}^{B\times H\times C}$.

\textbf{Gated Fusion for Dynamic Recomposition.}
A critical step is to recompose the component forecasts into a final prediction. While a simple summation ($\widehat{\mathbf{Y}}^{(T)} + \widehat{\mathbf{Y}}^{(S)} + \widehat{\mathbf{Y}}^{(R)}$) is a common approach, it rigidly assumes a linear and static relationship between the components. To instead model their complex, non-linear, and time-varying interactions, we introduce a \textbf{Gated Fusion} mechanism.

To capture the unique dynamics within each time series variate, our mechanism generates component weights independently for each channel. For a given channel $c$, the process is as follows:
First, a contextual feature vector $\mathbf{F}_{c} \in \mathbb{R}^{B \times H}$ is created by summing the three component forecasts for that channel:
\begin{equation}
    \mathbf{F}_{c} = \widehat{\mathbf{Y}}^{(T)}_{:,:,c} + \widehat{\mathbf{Y}}^{(S)}_{:,:,c} + \widehat{\mathbf{Y}}^{(R)}_{:,:,c}.
\end{equation}
This feature vector, which summarizes the overall predicted signal for the channel, is then passed through a shared weight generation network, $\mathrm{MLP}_{\text{gate}}$, to produce a dynamic adjustment term.

Our fusion employs a **residual weighting scheme**, where this dynamic term is added to a set of learnable, static base weights $\mathbf{W}_{\text{base}} \in \mathbb{R}^{3}$. This allows the model to learn a stable default weighting while still adapting to instance-specific patterns. The final weights $\mathbf{G}_c$ for channel $c$ are computed as:
\begin{equation}
\begin{aligned}
\Delta\mathbf{G}_{c} &= \mathrm{MLP}_{\text{gate}}(\mathbf{F}_{c}) \in \mathbb{R}^{B \times H \times 3}, \\
\mathbf{G}_{c} &= \mathbf{W}_{\text{base}} + \Delta\mathbf{G}_{c}.
\end{aligned}
\tag{16}
\end{equation}
Here, the base weights $\mathbf{W}_{\text{base}}$ are broadcast to match the shape of the dynamic adjustment $\Delta\mathbf{G}_{c}$. Finally, the recomposed forecast is obtained by taking the weighted sum of the components for each channel:
\begin{equation}
\widehat{\mathbf{Y}}_{:,:,c} = \sum_{k\in\{T,S,R\}} \left( \mathbf{G}_{c} \right)_{:,:,k} \odot \widehat{\mathbf{Y}}^{(k)}_{:,:,c}.
\tag{17}
\end{equation}
This channel-aware, residual gating not only provides a more expressive fusion but also enhances interpretability; by inspecting the final gate values $\mathbf{G}_c$, one can analyze how the model dynamically prioritizes trend, seasonality, and residuals for each individual variate in the forecast.

\section{Experiments}

In our experiments, we compare the proposed STELLA with a variety of baseline methods across multiple public datasets. We demonstrate the effectiveness of STELLA on a range of time series tasks, including long-term forecasting (Section 4.1), short-term forecasting (Section 4.2), few-shot forecasting (Section 4.3), and zero-shot forecasting (Section 4.4). Furthermore, we present ablation studies and parameter sensitivity analyses in Section 4.5. Finally, we visualize the semantically-guided time series representations to qualitatively assess the capability of STELLA. All experiments are conducted under a unified pipeline, following the experimental configurations described in \cite{wu2022timesnet}.

\textbf{Baselines.} We carefully select representative baselines from the recent time series forecasting landscape, including the following categories: (1) LLMs-based models: GPT4TS \cite{zhou2023one} and TimeLLM \cite{jin2023time}; (2) Transformer-based models: PatchTST \cite{nie2022time}, iTransformer \cite{liu2023itransformer}, Crossformer \cite{zhang2023crossformer}, ETSformer \cite{woo2022etsformer}, FEDformer \cite{zhou2022fedformer}, Autoformer \cite{wu2021autoformer}, Informer \cite{zhou2021informer}, NSformer \cite{liu2022non} and Reformer \cite{kitaev2020reformer}; (3) CNN-based models: TCN \cite{bai2018empirical}, TimesNet \cite{wu2022timesnet} and MICN \cite{wang2023micn}; (4) MLP-based models: DLinear \cite{zeng2023transformers} and TiDE \cite{das2023long}. Besides, N-HiTS \cite{challu2023nhits}, N-BEATS \cite{oreshkin2019n}, LSTM \cite{hochreiter1997long} and LSSL \cite{gu2021efficiently} are included for short-term forecasting.

\textbf{Implementation Details.} We adopt the methodology outlined in \cite{wu2022timesnet} to ensure a fair comparison. Specifically, we use the pre-trained LLaMA2-7B \cite{touvron2023llama} as our base large language model (LLM), selecting its first 6 Transformer layers as the backbone. Optimization is carried out using the Adam optimizer \cite{kingma2014adam}. For long-term forecasting, we apply either the L1 loss (MAE) or L2 loss (MSE), depending on the characteristics of the dataset. For short-term forecasting, the loss is computed using SMAPE. All experiments are conducted using 8 NVIDIA L20 GPUs (48 GB each) and 8 NVIDIA A800 GPUs (80 GB each). Further details are provided in Appendix~B.1.

\subsection{Long-term Forecasting}

\textbf{Setups.} We conduct experiments on seven widely-used real-world datasets, including the Electricity Transformer Temperature (ETT) dataset and its four subsets (\textbf{ETTh1}, \textbf{ETTh2}, \textbf{ETTm1}, \textbf{ETTm2}), Weather, Exchange, and Illness datasets. Detailed descriptions of the datasets are provided in \textbf{Appendix~B.3}. For a fair comparison, the input time series length $T$ is fixed at 96, and we adopt four distinct prediction horizons $H \in \{96, 192, 336, 720\}$. Consistent with prior works, the Mean Square Error (\textbf{MSE}) and Mean Absolute Error (\textbf{MAE}) are chosen as evaluation metrics. 

\textbf{Results.} The comprehensive long-term forecasting results are presented in Table~\ref{tab:longterm_summary}. Our method consistently delivers state-of-the-art performance, achieving the top results in 60 evaluation settings, while the closest competing baseline achieves the best performance in only 8 cases. Notably, compared to the recently published state-of-the-art Transformer-based models PatchTST and Crossformer, our method achieves relative reductions in MSE/MAE of 16.06\%/9.63\% and 55.91\%/38.44\%, respectively. When compared to the representative LLM-based method GPT4TS, our approach yields reductions of 20.25\% in MSE and 10.87\% in MAE. Against the CNN-based model MICN, our method reduces MSE/MAE by 24.61\%/20.78\%, and compared to the MLP-based method DLinear, we achieve improvements of 23.36\% in MSE and 17.55\% in MAE. Moreover, our method exhibits substantial improvements over other baselines, with performance gains exceeding 15\% in most cases.

\begin{table*}[!ht]
\centering
\caption{Multivariate long-term forecasting results. The input sequence length $T$ is set to 36 for the Illness dataset and 96 for all other baselines. The predictive lengths are set to $\{24, 36, 48, 60\}$ for Illness, and $\{96, 192, 336, 720\}$ for the other datasets. All reported values are averaged over the four prediction lengths used in each case. The \textbf{\textcolor{red}{best}} and \textcolor{softblue}{\uline{second-best}} results for each forecasting horizon are highlighted in \textbf{\textcolor{red}{bold}} and \textcolor{softblue}{\uline{underlined}}, respectively. Complete results are provided in Appendix~C.}
\label{tab:longterm_summary}
\scriptsize
\setlength{\tabcolsep}{3pt}
\renewcommand{\arraystretch}{1.2}
\begin{tabular}{ccccccccccccccccc}
\toprule
\multicolumn{1}{c}{Models} & \multicolumn{2}{c}{ETTm1} & \multicolumn{2}{c}{ETTm2} & \multicolumn{2}{c}{ETTh1} & \multicolumn{2}{c}{ETTh2} & \multicolumn{2}{c}{Weather} & \multicolumn{2}{c}{Exchange} & \multicolumn{2}{c}{Illness} \\
\cmidrule(r){2-3} \cmidrule(r){4-5} \cmidrule(r){6-7} \cmidrule(r){8-9}
\cmidrule(r){10-11} \cmidrule(r){12-13} \cmidrule(r){14-15}
\multicolumn{1}{c}{Metric} & MSE & MAE & MSE & MAE & MSE & MAE & MSE & MAE & MSE & MAE & MSE & MAE & MSE & MAE \\
\midrule
\textbf{STELLA (Ours)}      & \textbf{\textcolor{red}{0.379}} & \textbf{\textcolor{red}{0.388}} & \textbf{\textcolor{red}{0.275}} & \textbf{\textcolor{red}{0.317}} & \textbf{\textcolor{red}{0.416}} & \textbf{\textcolor{red}{0.425}} & \textbf{\textcolor{red}{0.364}} & \textbf{\textcolor{red}{0.391}} & \textcolor{softblue}{\uline{0.243}} & \textbf{\textcolor{red}{0.265}} & \textcolor{softblue}{\uline{0.339}} & \textbf{\textcolor{red}{0.394}} & \textbf{\textcolor{red}{1.819}} & \textbf{\textcolor{red}{0.814}} \\
\midrule
GPT4TS\cite{zhou2023one}                   & 0.396 & \textcolor{softblue}{\uline{0.401}} & 0.294 & 0.329 & 0.457 & 0.450 & 0.389 & 0.414 & 0.279 & 0.297 & 0.371 & 0.409 & 2.623 & 1.060 \\
TimeLLM\cite{jin2023time}                  & 0.410 & 0.409 & 0.297 & 0.341 & 0.460 & 0.449 & 0.381 & 0.408 & 0.264 & 0.284 & 0.359* & \textcolor{softblue}{\uline{0.403*}} & \textcolor{softblue}{\uline{2.000*}} & \textcolor{softblue}{\uline{0.880*}} \\
PatchTST\cite{nie2022time}                 & \textcolor{softblue}{\uline{0.392}} & 0.402 & \textcolor{softblue}{\uline{0.285}} & 0.328 & 0.463 & 0.449 & 0.395 & 0.414 & 0.257 & 0.280 & 0.390 & 0.429 & 2.388 & 1.011 \\
iTransformer\cite{liu2023itransformer}             & 0.407 & 0.411 & 0.292 & 0.335 & 0.455 & 0.448 & \textcolor{softblue}{\uline{0.381}} & \textcolor{softblue}{\uline{0.405}} & 0.257 & \textcolor{softblue}{\uline{0.279}} & 0.360 & 0.403 & 2.183* & 0.943* \\
Crossformer\cite{zhang2023crossformer}              & 0.502 & 0.503 & 1.217 & 0.708 & 0.620 & 0.572 & 0.942 & 0.684 & 0.259 & 0.315 & 0.727* & 0.653* & 4.431* & 1.429* \\
FEDformer\cite{zhou2022fedformer}                & 0.448 & 0.452 & 0.305 & 0.349 & \textcolor{softblue}{\uline{0.440}} & 0.460 & 0.437 & 0.449 & 0.309 & 0.360 & 0.519 & 0.500 & 2.847 & 1.144 \\
Autoformer\cite{wu2021autoformer}               & 0.588 & 0.517 & 0.327 & 0.371 & 0.496 & 0.487 & 0.450 & 0.459 & 0.338 & 0.382 & 0.613 & 0.539 & 3.006 & 1.161 \\
Informer\cite{zhou2021informer}                 & 0.961 & 0.734 & 1.410 & 0.810 & 1.040 & 0.795 & 4.431 & 1.729 & 0.634 & 0.548 & 1.550 & 0.998 & 5.137 & 1.544 \\
NSformer\cite{liu2022non}               & 0.481 & 0.456 & 0.306 & 0.347 & 0.570 & 0.537 & 0.526 & 0.516 & 0.288 & 0.314 & 0.461 & 0.454 & 2.077 & 0.914 \\
Reformer\cite{kitaev2020reformer}                 & 0.799 & 0.671 & 1.479 & 0.915 & 1.029 & 0.805 & 6.736 & 2.191 & 0.803 & 0.656 & 1.280 & 0.932 & 4.724 & 1.445 \\
TimesNet\cite{wu2022timesnet}                 & 0.400 & 0.406 & 0.291 & 0.333 & 0.458 & 0.450 & 0.414 & 0.427 & 0.259 & 0.287 & 0.416 & 0.443 & 2.139 & 0.931 \\
MICN\cite{wang2023micn}                     & \textcolor{softblue}{\uline{0.392}} & 0.414 & 0.328 & 0.382 & 0.558 & 0.535 & 0.587 & 0.525 & \textbf{\textcolor{red}{0.242}} & 0.299 & \textbf{\textcolor{red}{0.315}} & 0.404 & 2.664 & 1.221 \\
DLinear\cite{zeng2023transformers}                  & 0.403 & 0.407 & 0.350 & 0.401 & 0.456 & 0.452 & 0.559 & 0.551 & 0.265 & 0.317 & 0.354 & 0.414 & 2.616 & 1.090 \\
TiDE\cite{das2023long}                     & 0.412 & 0.406 & 0.289 & \textcolor{softblue}{\uline{0.326}} & 0.445 & \textcolor{softblue}{\uline{0.432}} & 0.611 & 0.550 & 0.271 & 0.320 & \textcolor{softblue}{\uline{0.339*}} & \textcolor{softblue}{\uline{0.400*}} & 4.020* & 1.454* \\
\bottomrule
\end{tabular}
\vspace{2pt}
\parbox{\textwidth}{\scriptsize\textit{* indicates that the model did not report results on this dataset in the original paper. The results were reproduced using the code provided by the authors, ensuring that the experimental settings (including hyperparameters, data splits, etc.) were consistent with those described in the original paper. Other results are from TimesNet \cite{wu2022timesnet} and iTransformer \cite{liu2023itransformer}.}}
\end{table*}

\subsection{Short-term Forecasting}

\textbf{Setup.} We evaluate the effectiveness of \textsc{STELLA} in the short-term forecasting setting using the M4 datasets \cite{makridakis2018m4}, which consist of a diverse collection of marketing time series sampled at varying frequencies. Comprehensive details of the datasets are provided in Appendix~B.3.In this setup, the prediction horizons are considerably shorter than those in the long-term forecasting scenario, ranging from 6 to 48. Consistent with the experimental configurations in \cite{jin2023time, zhou2023one}, the input sequence lengths are set to be twice the size of the prediction horizons.To assess forecasting performance, we adopt three standard evaluation metrics: symmetric mean absolute percentage error (\textbf{SMAPE}), mean absolute scaled error (\textbf{MASE}), and overall weighted average (\textbf{OWA}). Definitions and computation details of these metrics are provided in Appendix~B.4.

\textbf{Results.} As shown in Table~\ref{tab:shortterm_summary}, our method delivers consistently strong performance in short-term forecasting across multiple evaluation metrics. Notably, it achieves the best results in all 15 out of 15 evaluation categories, clearly surpassing all baseline models. In particular, when compared to TimesNet—the current state-of-the-art in short-term forecasting—our model attains an overall performance improvement of \textbf{1\%}. This improvement is calculated based on the relative reduction in the average \textbf{OWA}, the primary evaluation metric in the M4 benchmark, which combines both \textbf{SMAPE} and \textbf{MASE} through a weighted aggregation scheme.

\begin{table*}[!ht]
    \centering
    \caption{Short-term time series forecasting results on the M4 dataset. The forecasting horizons are in the range of $[6, 48]$, and lower values indicate better performance. More detailed short-term forecasting results are provided in Appendix D.}
    \label{tab:shortterm_summary}
    \footnotesize
    \setlength{\heavyrulewidth}{1.1pt}  
    \setlength{\lightrulewidth}{0.5pt}  
    \setlength{\cmidrulewidth}{0.5pt}   
    \resizebox{\textwidth}{!}{
    \renewcommand{\arraystretch}{1.3}
    \begin{tabular}{@{}p{0.2cm}|c|ccccccccccccc|}
        \toprule 
        \multicolumn{2}{c|}{Models} & \makecell{\textbf{STELLA}\\\textbf{(Ours)}} & \makecell{TimeLLM\\ \cite{jin2023time}} & \makecell{GPT4TS\\ \cite{zhou2023one}} & \makecell{PatchTST\\ \cite{nie2022time}} & \makecell{ETSformer\\ \cite{woo2022etsformer}} & \makecell{FEDformer\\ \cite{zhou2022fedformer}} & \makecell{Autoformer\\ \cite{wu2021autoformer}} & \makecell{TimesNet\\ \cite{wu2022timesnet}} & \makecell{TCN\\ \cite{bai2018empirical}} & \makecell{N-HiTS\\ \cite{challu2023nhits}} & \makecell{N-BEATS \\ \cite{oreshkin2019n}} & \makecell{DLinear\\ \cite{zeng2023transformers}} \\
        \midrule[\heavyrulewidth] 
        \multirow{3}{*}{\rotatebox{90}{\scriptsize Yearly}} & SMAPE & \textbf{\textcolor{red}{13.348}} & 13.419 & 13.531 & 13.477 & 18.009 & 13.728 & 13.974 & \textcolor{softblue}{\uline{13.387}} & 14.920 & 13.418 & 13.436 & 16.965 \\
                                  & MASE  & \textbf{\textcolor{red}{2.978}}  & 3.005  & 3.015  & 3.019  & 4.487  & 3.048  & 3.134  & \textcolor{softblue}{\uline{2.996}}  & 3.364  & 3.045  & 3.043  & 4.283 \\
                                  & OWA   & \textbf{\textcolor{red}{0.783}}  & 0.789  & 0.793  & 0.792  & 1.115  & 0.803  & 0.822  & \textcolor{softblue}{\uline{0.786}}  & 0.880  & 0.793  & 0.794  & 1.058 \\
        \midrule 
        \multirow{3}{*}{\rotatebox{90}{\scriptsize Quarterly}} & SMAPE & \textbf{\textcolor{red}{9.997}} & 10.110 & 10.177 & 10.380 & 13.376 & 10.792 & 11.338 & \textcolor{softblue}{\uline{10.100}} & 11.122 & 10.202 & 10.124 & 12.145 \\
                                   & MASE  & \textbf{\textcolor{red}{1.164}} & 1.178  & 1.194  & 1.233  & 1.906  & 1.283  & 1.365  & 1.182  & 1.360  & 1.194  & \textcolor{softblue}{\uline{1.169}}  & 1.520 \\
                                   & OWA   & \textbf{\textcolor{red}{0.878}} & 0.889  & 0.898  & 0.921  & 1.302  & 0.958  & 1.012  & 0.890  & 1.001  & 0.899  & \textcolor{softblue}{\uline{0.886}}  & 1.106 \\
        \midrule
        \multirow{3}{*}{\rotatebox{90}{\scriptsize Monthly}} & SMAPE & \textbf{\textcolor{red}{12.608}} & 12.980 & 12.894 & 12.959 & 14.588 & 14.260 & 13.958 & 12.679 & 15.626 & 12.791 & \textcolor{softblue}{\uline{12.677}} & 13.514 \\
                                 & MASE  & \textbf{\textcolor{red}{0.932}}  & 0.963  & 0.956  & 0.970  & 1.368  & 1.102  & 1.103  & \textcolor{softblue}{\uline{0.933}}  & 1.274  & 0.969  & 0.937  & 1.037 \\
                                 & OWA   & \textbf{\textcolor{red}{0.875}}  & 0.903  & 0.897  & 0.905  & 1.149  & 1.012  & 1.002  & \textcolor{softblue}{\uline{0.878}}  & 1.141  & 0.899  & 0.880  & 0.956 \\
        \midrule
        \multirow{3}{*}{\rotatebox{90}{\scriptsize Others}} & SMAPE & \textbf{\textcolor{red}{4.651}}  & \textcolor{softblue}{\uline{4.795}}  & 4.940  & 4.952  & 7.267  & 4.954  & 5.485  & 4.891  & 7.186  & 5.061  & 4.925  & 6.709 \\
                                & MASE  & \textbf{\textcolor{red}{3.112}}  & \textcolor{softblue}{\uline{3.178}}  & 3.228  & 3.347  & 5.240  & 3.264  & 3.865  & 3.302  & 4.677  & 3.216  & 3.391  & 4.953 \\
                                & OWA   & \textbf{\textcolor{red}{0.980}}  & \textcolor{softblue}{\uline{1.006}}  & 1.029  & 1.049  & 1.591  & 1.036  & 1.187  & 1.035  & 1.494  & 1.040  & 1.053  & 1.487 \\
        \midrule
        \multirow{3}{*}{\rotatebox{90}{\scriptsize Average}} & SMAPE & \textbf{\textcolor{red}{11.754}} & 11.983 & 11.991 & 12.059 & 14.718 & 12.840 & 12.909 & \textcolor{softblue}{\uline{11.829}} & 13.961 & 11.927 & 11.851 & 13.639 \\
                                 & MASE  & \textbf{\textcolor{red}{1.567}}  & 1.595  & 1.600  & 1.623  & 2.408  & 1.701  & 1.771  & \textcolor{softblue}{\uline{1.585}}  & 1.945  & 1.613  & 1.599  & 2.095 \\
                                 & OWA   & \textbf{\textcolor{red}{0.843}}  & 0.859  & 0.861  & 0.869  & 1.172  & 0.918  & 0.939  & \textcolor{softblue}{\uline{0.851}}  & 1.023  & 0.861  & 0.855  & 1.051 \\
        \bottomrule
    \end{tabular}}
\end{table*}

\subsection{Few-shot Learning}
\textbf{Setups.} Following the experimental protocol of \cite{zhou2023one}, we assess the performance of our model in the few-shot forecasting setting. Experiments are conducted on four \textbf{ETT} datasets, where only the first 10\% of the training data is used for each dataset. This constrained data scenario presents a substantial challenge, designed to evaluate the model's ability to learn effectively under limited supervision.

\textbf{Results.} Under the 10\% training data setting, our method consistently demonstrates state-of-the-art performance across multiple datasets. As shown in Table~\ref{tab:fewshot_summary}, it achieves the best results in 23 out of 40 evaluations and ranks second in 9 additional cases. In contrast, the closest competing baseline, TiDE, obtains the top result in only 9 evaluations. Compared to TiDE, our method achieves a reduction of 7.58\% in MSE and 4.85\% in MAE. Furthermore, when compared to a representative LLM-based method, GPT4TS, our approach yields relative improvements of 14.40\% in MSE and 7.37\% in MAE.

\begin{table*}[!ht]
    \centering
    \caption{Few-shot learning performance on the \textbf{ETT} datasets using 10\% of the training data. Results are averaged across four prediction horizons, $H \in \{96, 192, 336, 720\}$. Comprehensive results are presented in Appendix E.}
    \label{tab:fewshot_summary}
    \small
    \setlength{\heavyrulewidth}{1.1pt}  
    \setlength{\lightrulewidth}{0.5pt}  
    \setlength{\cmidrulewidth}{0.5pt}   
    \resizebox{\textwidth}{!}{  
    \begin{tabular}{ccccccccccccccccccccc}
    \toprule
        Models &  \multicolumn{2}{c}{\makecell{\textbf{STELLA}\\\textbf{(Ours)}}}  & \multicolumn{2}{c}{\makecell{TimeLLM\\ \cite{jin2023time}}}  &  \multicolumn{2}{c}{\makecell{GPT4TS\\ \cite{zhou2023one}}} &  \multicolumn{2}{c}{\makecell{PatchTST\\ \cite{nie2022time}}} &  \multicolumn{2}{c}{\makecell {Crossformer\\ \cite{zhang2023crossformer}}}  &  \multicolumn{2}{c}{\makecell{FEDformer\\ \cite{zhou2022fedformer}}} &  \multicolumn{2}{c}{\makecell{TimesNet\\ \cite{wu2022timesnet}}} &  \multicolumn{2}{c}{\makecell{MICN \\ \cite{wang2023micn}}} &  \multicolumn{2}{c}{\makecell{DLinear\\ \cite{zeng2023transformers}}}  &  \multicolumn{2}{c}{\makecell{TiDE\\ \cite{das2023long}}}  \\
        \cmidrule(lr){2-3}  \cmidrule(lr){4-5}  \cmidrule(lr){6-7}  \cmidrule(lr){8-9}   \cmidrule(lr){10-11} \cmidrule(lr){12-13} \cmidrule(lr){14-15} \cmidrule(lr){16-17} \cmidrule(lr){18-19} \cmidrule(lr){20-21}
        Metric & MSE & MAE & MSE & MAE & MSE & MAE & MSE & MAE & MSE & MAE & MSE & MAE & MSE & MAE & MSE & MAE & MSE & MAE & MSE & MAE \\
        \midrule[\heavyrulewidth] 
        ETTm1 & \textcolor{softblue}{\uline{0.543}}  & \textcolor{softblue}{\uline{0.473}}  & 0.636 & 0.512 & 0.608 & 0.500 & 0.557 & 0.483 & 1.340 & 0.848 & 0.696 & 0.572 & 0.673 & 0.534 & 0.970 & 0.674 & 0.567 & 0.499 & \textbf{\textcolor{red}{0.515}} & \textbf{\textcolor{red}{0.469}} \\
        \midrule
        ETTm2 & \textbf{\textcolor{red}{0.291}} & \textbf{\textcolor{red}{0.332}}  & 0.308 & 0.343 & 0.303 & 0.336 & \textcolor{softblue}{\uline{0.295}} & \textcolor{softblue}{\uline{0.334}} & 1.985 & 1.048 & 0.356 & 0.392 & 0.321 & 0.354 & 1.073 & 0.716 & 0.329 & 0.382 & 0.303 & 0.337 \\
        \midrule
        ETTh1 & \textbf{\textcolor{red}{0.617}}  & \textbf{\textcolor{red}{0.522}}  & 0.765 & 0.584 & 0.689 & 0.555 & 0.683 & 0.645 & 1.744 & 0.914 & 0.750  & 0.607 & 0.865 & 0.625 & 1.405 & 0.814 & \textcolor{softblue}{\uline{0.647}} & \textcolor{softblue}{\uline{0.552}} & 0.779 & 0.604 \\
        \midrule
        ETTh2 & \textbf{\textcolor{red}{0.414}}  & \textbf{\textcolor{red}{0.421}}  & 0.589 & 0.498 & 0.579 & 0.497 & 0.550  & 0.487 & 3.139 & 1.378 & 0.553 & 0.525 & 0.476 & 0.463 & 2.533 & 1.158 & 0.441 & 0.458 & \textcolor{softblue}{\uline{0.421}} & \textcolor{softblue}{\uline{0.428}} \\
    \bottomrule
  \end{tabular}}
\end{table*}

\subsection{Zero-shot Learning}

\textbf{Setups.} We adopt a zero-shot forecasting setting to evaluate the cross-domain generalization ability of our model. In this setup, the model is trained on one ETT dataset and directly evaluated on a different target dataset, without access to its training samples. We consider four ETT datasets: \textbf{ETTh1}, \textbf{ETTh2}, \textbf{ETTm1}, and \textbf{ETTm2}. The notation “A $\rightarrow$ B” indicates that the model is trained on dataset A and evaluated on dataset B (e.g., \texttt{ETTh1} $\rightarrow$ \texttt{ETTm2}). This zero-shot protocol follows the experimental design of \cite{jin2023time}. Results are averaged over four prediction lengths, $H \in \{96, 192, 336, 720\}$, and full results are provided in Appendix~F.3.

\textbf{Results.} As shown in Table~\ref{tab:zeroshot_summary}, \textbf{STELLA} significantly outperforms nine state-of-the-art time series models in terms of zero-shot adaptation. Notably, it achieves the best performance across \textbf{all 40 out of 40} evaluation settings, clearly surpassing all baselines. The second-best model, TimeLLM, reaches the top rank in only one case. On average, \textbf{STELLA} yields a \textbf{24.12\% reduction in MSE} and a \textbf{15.61\% reduction in MAE} compared to all baseline methods.Furthermore, when compared specifically to TimeLLM---a strong representative LLM-based approach---our method achieves \textbf{relative improvements of 11.79\% in MSE} and \textbf{4.49\% in MAE}. We also observe substantial gains over Transformer-based models; for instance, STELLA surpasses \textbf{PatchTST}, one of the strongest non-LLM baselines, by \textbf{17.11\% in MSE} and \textbf{8.59\% in MAE}. Additionally, LLM-based models in general outperform most Transformer-based baselines, likely due to the transferable knowledge acquired through pretraining on large-scale sequential corpora. This highlights the advantages of leveraging LLMs for zero-shot time series forecasting.

\begin{table*}[!ht]
    \centering
    \caption{Zero-shot forecasting results on ETT datasets. Each row represents a cross-dataset evaluation setting, where the model is trained on one ETT dataset and evaluated on another (e.g., \texttt{ETTh1} $\rightarrow$ \texttt{ETTm2}). All results are averaged over four prediction lengths, $H \in \{96, 192, 336, 720\}$. See Appendix F for the complete results.}
    \label{tab:zeroshot_summary}
    \small
    \setlength{\heavyrulewidth}{1.1pt}  
    \setlength{\lightrulewidth}{0.5pt} 
    \setlength{\cmidrulewidth}{0.5pt}  
    \resizebox{\textwidth}{!}{  
    \begin{tabular}{ccccccccccccccccccccc}
    \toprule
      Models &  \multicolumn{2}{c}{\makecell{\textbf{STELLA}\\\textbf{(Ours)}}}  & \multicolumn{2}{c}{\makecell{TimeLLM\\ \cite{jin2023time}}}  & \multicolumn{2}{c}{\makecell{GPT4TS\\ \cite{zhou2023one}}} & \multicolumn{2}{c}{\makecell{PatchTST\\ \cite{nie2022time}}}  & \multicolumn{2}{c}{\makecell {Crossformer\\ \cite{zhang2023crossformer}}} &  \multicolumn{2}{c}{\makecell{FEDformer\\ \cite{zhou2022fedformer}}} & \multicolumn{2}{c}{\makecell{TimesNet\\ \cite{wu2022timesnet}}}  & \multicolumn{2}{c}{\makecell{MICN \\ \cite{wang2023micn}}} & \multicolumn{2}{c}{\makecell{DLinear\\ \cite{zeng2023transformers}}}   & \multicolumn{2}{c}{\makecell{TiDE\\ \cite{das2023long}}} \\ 
      \cmidrule(lr){2-3}  \cmidrule(lr){4-5}  \cmidrule(lr){6-7}  \cmidrule(lr){8-9}   \cmidrule(lr){10-11} \cmidrule(lr){12-13} \cmidrule(lr){14-15} \cmidrule(lr){16-17} \cmidrule(lr){18-19} \cmidrule(lr){20-21}
      Metric & MSE & MAE & MSE & MAE & MSE & MAE & MSE & MAE & MSE & MAE & MSE & MAE & MSE & MAE & MSE & MAE & MSE & MAE & MSE & MAE \\ 
      \midrule[\heavyrulewidth] 
      h1->m1 & \textbf{\textcolor{red}{0.719}}  & \textbf{\textcolor{red}{0.561}}  & 0.847 & 0.6 & 0.798 & \textcolor{softblue}{\uline{0.574}} & 0.894 & 0.610  & 0.999 & 0.736 & 0.765 & 0.588 & 0.794 & 0.575 & 1.439 & 0.870  & \textcolor{softblue}{\uline{0.760}}  & 0.577 & 0.774 & \textcolor{softblue}{\uline{0.574}} \\ 
      \midrule
      h1->m2 & \textbf{\textcolor{red}{0.312}}  & \textbf{\textcolor{red}{0.353}}  & \textcolor{softblue}{\uline{0.315}} & \textcolor{softblue}{\uline{0.357}} & 0.317 & 0.359 & 0.318 & 0.362 & 1.120  & 0.789 & 0.357 & 0.403 & 0.339 & 0.370  & 2.428 & 1.236 & 0.399 & 0.439 & 0.314 & 0.355 \\ 
      \midrule
      h2->m1 & \textbf{\textcolor{red}{0.730}}  & \textbf{\textcolor{red}{0.562}}  & 0.868 & 0.595 & 0.920  & 0.610  & 0.871 & 0.596 & 1.195 & 0.711 & \textcolor{softblue}{\uline{0.741}} & \textcolor{softblue}{\uline{0.588}} & 1.286 & 0.705 & 0.764 & 0.601 & 0.778 & 0.594 & 0.841 & 0.590  \\ 
      \midrule
      h2->m2 & \textbf{\textcolor{red}{0.313}}  & \textbf{\textcolor{red}{0.354}}  & 0.322 & \textcolor{softblue}{\uline{0.363}} & 0.331 & 0.371 & 0.420  & 0.433 & 2.043 & 1.124 & 0.365 & 0.405 & 0.361 & 0.390  & 0.527 & 0.519 & 0.496 & 0.496 & \textcolor{softblue}{\uline{0.321}} & 0.364 \\ 
    \bottomrule
  \end{tabular}}
\end{table*}

\subsection{Method Analysis}
\textbf{Ablation Study.}
To validate our architectural design and dissect the contribution of each key component, we conduct a comprehensive ablation study on the ETTh1, ETTm1, and Exchange datasets. We systematically deconstruct our full STELLA model by creating four variants, each ablating a single module: \textbf{(i) w/o N-STL}, which removes the Neural STL decomposition; \textbf{(ii) w/o TC-P}, which replaces our Temporal Convolutional Patch Encoder with a standard linear projection; \textbf{(iii) w/o FBP}, which ablates the instance-specific behavioral prompt; and \textbf{(iv) w/o CSP}, which removes the corpus-level semantic prior. The results, summarized in Table~\ref{tab:ablation}, unequivocally affirm that each component is indispensable for achieving the model's optimal performance.

The most significant performance degradation stems from the removal of the \textbf{Neural STL (N-STL)} module. On the volatile Exchange dataset, for instance, its absence increases the average MSE by 9.4

The efficacy of our \textbf{Temporal Convolutional Patch Encoder (TC-Patch)} is similarly pronounced. When replaced by a simpler linear projection (w/o TC-P), the model's performance consistently deteriorates, with the average MSE on Exchange rising by 6.5

Furthermore, the hierarchical prompting mechanism, comprising the \textbf{FBP} and \textbf{CSP}, serves as a critical layer of semantic guidance that unlocks further performance enhancements. The results confirm that both prompts are instrumental to the model's success. Removing the Corpus-level Semantic Prior (w/o CSP) degrades performance, underscoring the value of a stable, high-level inductive bias regarding the data's domain. Likewise, ablating the Fine-grained Behavioral Prompt (w/o FBP) leads to a consistent performance drop, validating the importance of conditioning the model on adaptive, instance-specific behavioral signatures. These prompting modules are not mere supplements; they are integral components that refine the model's understanding and are essential for pushing the forecasting accuracy to state-of-the-art levels.

In conclusion, our ablation study validates a clear and effective design hierarchy. The N-STL and TC-Patch modules collaborate to build a robust and efficient temporal representation foundation. This foundation is then intelligently conditioned and refined by the hierarchical FBP and CSP prompts. It is this synergistic interplay—from coarse-grained decomposition to fine-grained semantic guidance—that underpins the superior performance of the complete STELLA model.

\begin{table}[H]
    \centering
    \caption{Ablation study confirming the synergistic contribution of STELLA’s core components. Performance, measured in MSE and MAE, consistently degrades upon the removal of the Neural STL (N-STL), the TC-Patch encoder (TC-P), or the prompting mechanisms (FBP, CSP), validating our integrated design.}
    \label{tab:ablation}
    \setlength{\heavyrulewidth}{1.1pt}
    \setlength{\lightrulewidth}{0.5pt}
    \setlength{\cmidrulewidth}{0.5pt}
    \resizebox{\columnwidth}{!}{
    \begin{tabular}{cc|cc|cc|cc|cc|cc}
    \toprule
        \multicolumn{2}{c}{Models} 
        & \multicolumn{2}{c}{\makecell{\textbf{STELLA}\\\textbf{(Ours)}}}   
        & \multicolumn{2}{c}{\makecell{w/o\\TC-P}} 
        & \multicolumn{2}{c}{\makecell{w/o\\N-STL}} 
        & \multicolumn{2}{c}{\makecell{w/o\\FBP}} 
        & \multicolumn{2}{c}{\makecell{w/o\\CSP}} \\
    \toprule
        \multicolumn{2}{c}{Metric} 
        & MSE & MAE 
        & MSE & MAE 
        & MSE & MAE 
        & MSE & MAE 
        & MSE & MAE \\
    \toprule
        \multirow{5}{*}{\rotatebox{90}{ETTh1}}  
        & 96 & \textbf{\textcolor{red}{0.368}} & \textbf{\textcolor{red}{0.390}} & 0.370 & \textbf{\textcolor{red}{0.390}} & 0.386 & 0.400 & 0.371 & 0.392 & 0.373 & 0.393 \\ 
        ~ & 192 & \textbf{\textcolor{red}{0.406}} & 0.421 & 0.418 & 0.424 & 0.422 & 0.428 & 0.415 & 0.425 & 0.416 & \textbf{\textcolor{red}{0.416}} \\ 
        ~ & 336 & \textbf{\textcolor{red}{0.446}} & \textbf{\textcolor{red}{0.442}} & 0.457 & 0.448 & 0.453 & 0.446 & 0.448 & 0.447 & 0.451 & 0.443 \\ 
        ~ & 720 & \textbf{\textcolor{red}{0.445}} & \textbf{\textcolor{red}{0.448}} & 0.456 & 0.455 & 0.450 & 0.453 & 0.449 & 0.451 & 0.455 & 0.454 \\ 
        \cmidrule(lr){2-12}
        ~ & Avg. & \textbf{\textcolor{red}{0.416}} & \textbf{\textcolor{red}{0.425}} & 0.425 & 0.430 & 0.428 & 0.432 & 0.421 & 0.429 & 0.424 & 0.426 \\ 
        \midrule
        \multirow{5}{*}{\rotatebox{90}{ETTm1}}  
        & 96 & \textbf{\textcolor{red}{0.310}} & \textbf{\textcolor{red}{0.343}} & 0.311 & \textbf{\textcolor{red}{0.343}} & 0.329 & 0.359 & 0.317 & 0.348 & 0.313 & 0.344 \\ 
        ~ & 192 & \textbf{\textcolor{red}{0.358}} & \textbf{\textcolor{red}{0.371}} & \textbf{\textcolor{red}{0.358}} & \textbf{\textcolor{red}{0.371}} & 0.372 & 0.382 & 0.362 & 0.374 & 0.359 & 0.372 \\ 
        ~ & 336 & \textbf{\textcolor{red}{0.391}} & \textbf{\textcolor{red}{0.404}} & 0.392 & 0.405 & 0.405 & 0.412 & 0.395 & 0.405 & 0.396 & 0.418 \\ 
        ~ & 720 & 0.460 & \textbf{\textcolor{red}{0.431}} & \textbf{\textcolor{red}{0.456}} & 0.435 & 0.465 & 0.436 & 0.464 & 0.435 & 0.458 & \textbf{\textcolor{red}{0.431}} \\ 
        \cmidrule(lr){2-12}
        ~ & Avg. & 0.380 & \textbf{\textcolor{red}{0.387}} & \textbf{\textcolor{red}{0.379}} & 0.389 & 0.393 & 0.397 & 0.385 & 0.391 & 0.382 & 0.391 \\ 
        \midrule
        \multirow{5}{*}{\rotatebox{90}{Exchange}}  
        & 96 & \textbf{\textcolor{red}{0.084}} & \textbf{\textcolor{red}{0.204}} & 0.085 & \textbf{\textcolor{red}{0.204}} & 0.100 & 0.226 & 0.085 & 0.207 & 0.086 & 0.208 \\ 
        ~ & 192 & \textbf{\textcolor{red}{0.172}} & \textbf{\textcolor{red}{0.298}} & 0.186 & 0.307 & 0.194 & 0.317 & 0.173 & 0.301 & 0.177 & 0.303 \\ 
        ~ & 336 & 0.315 & \textbf{\textcolor{red}{0.407}} & 0.329 & 0.416 & 0.338 & 0.421 & \textbf{\textcolor{red}{0.312}} & 0.411 & 0.324 & 0.411 \\ 
        ~ & 720 & \textbf{\textcolor{red}{0.785}} & \textbf{\textcolor{red}{0.666}} & 0.844 & 0.692 & 0.850 & 0.700 & 0.804 & 0.674 & 0.819 & 0.681 \\ 
        \cmidrule(lr){2-12}
        ~ & Avg. & \textbf{\textcolor{red}{0.339}} & \textbf{\textcolor{red}{0.394}} & 0.361 & 0.405 & 0.371 & 0.416 & 0.343 & 0.398 & 0.352 & 0.401 \\ 
    \bottomrule
     \end{tabular}}
\end{table}

\textbf{Parameter Sensitivity.}
To understand the impact of prompt length, we analyzed the performance of STELLA with varying lengths for both the CSP and FBP. Our analysis reveals a clear trade-off: forecasting accuracy generally improves as prompt length increases up to an optimal threshold (10-20 for CSP, 12-24 for FBP), beyond which performance declines due to the introduction of semantic redundancy or attentional noise. This underscores the importance of our carefully tuned prompt design. \textbf{The full experimental details and results are presented in Appendix G, including Figure 6.}

\begin{figure}[!tbp]
    \centering
    \subfloat[Disentangled Component Representations\label{fig:umap_components}]{%
        \resizebox{0.94\columnwidth}{!}{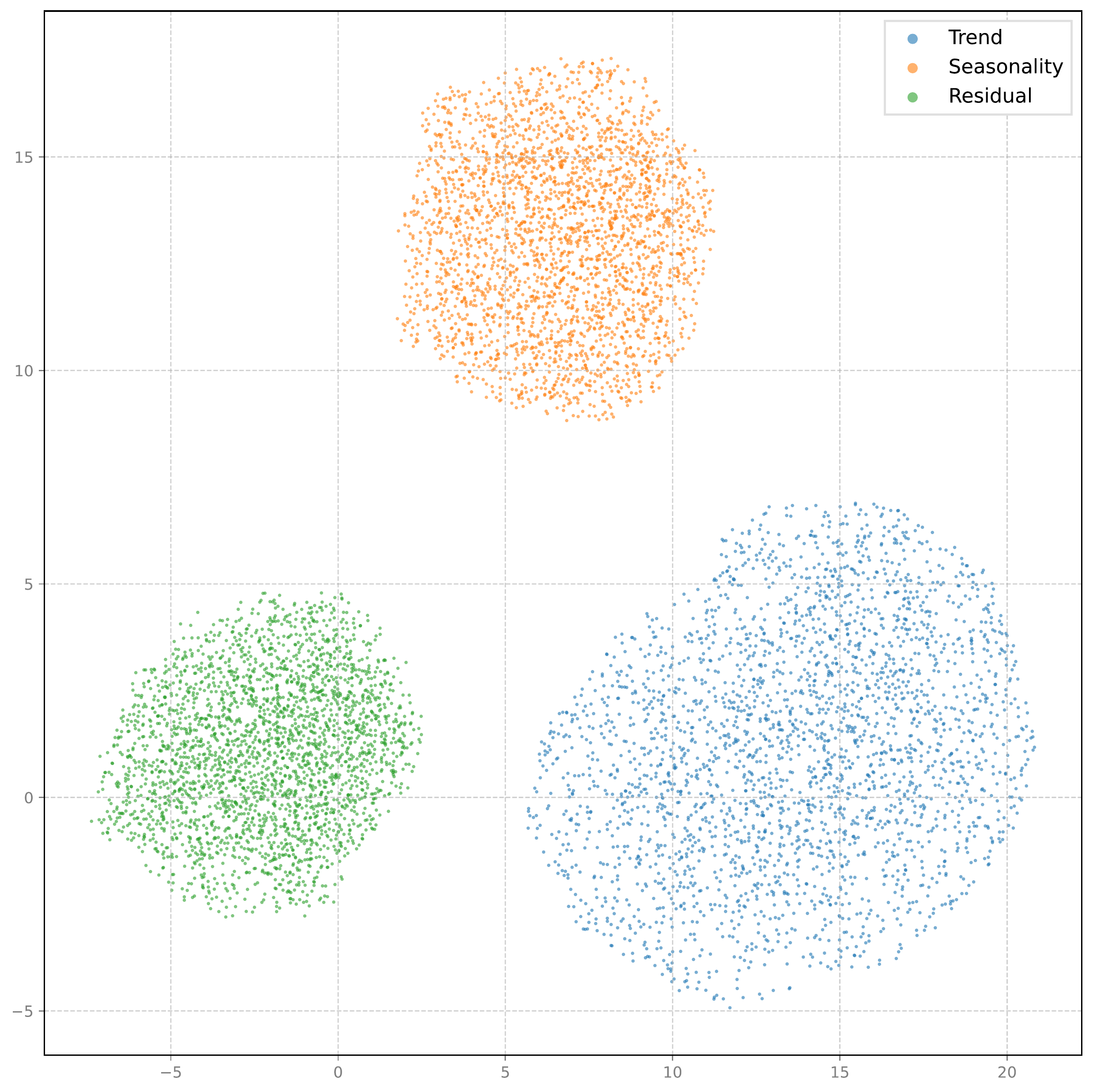}
    }
    \hfil 
    \subfloat[Hierarchical Semantic Anchors\label{fig:umap_anchors}]{%
        \resizebox{0.94\columnwidth}{!}{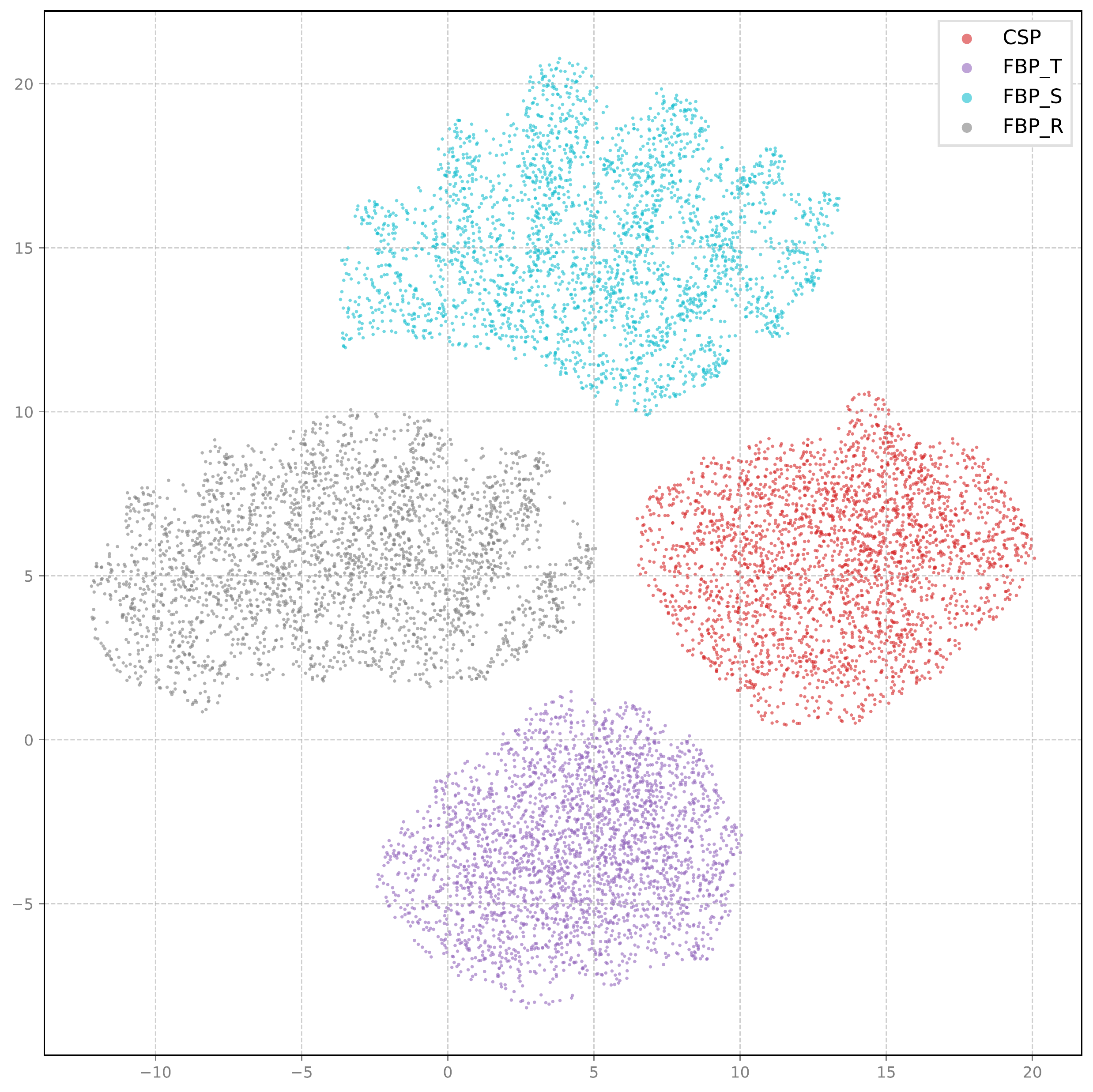}
    }
    
    \caption{UMAP visualization of \textbf{Semantic-Driven Disentanglement}. The disentanglement of final component representations (a) is shown to be a direct result of the clear separation of their guiding \textbf{Hierarchical Semantic Anchors} (b).}
    \label{fig:disentanglement_viz_main}
\end{figure}

\textbf{Disentanglement Effect Investigation.} A core premise of STELLA is that explicit \textbf{semantic guidance} compels the LLM to learn a meaningfully structured and disentangled latent space. To verify this, we visualize the manifold of two key sets of embeddings using Uniform Manifold Approximation and Projection (UMAP)~\cite{mcinnes2018umap}. The first set contains the final output representations of the three time series components ($\mathbf{Z}^{(T)}, \mathbf{Z}^{(S)}, \mathbf{Z}^{(R)}$) after being processed by the LLM (Figure~\ref{fig:umap_components}). The second set comprises the embeddings of our proposed \textbf{Hierarchical Semantic Anchors} (the global CSP and the instance-specific FBPs), which serve as the guiding signals (Figure~\ref{fig:umap_anchors}). 
As depicted in Figure~\ref{fig:disentanglement_viz_main}, the projection reveals a striking outcome. The component representations converge into highly separated and compact regions, demonstrating that the LLM has successfully learned to distinguish their intrinsic patterns. Crucially, the semantic anchors that guide them also form distinct clusters, showcasing their unique semantic roles. This dual clustering provides compelling evidence of a successful \textbf{semantic-to-temporal alignment}: the clear separation learned for the temporal components is a direct reflection of the distinctness of their guiding semantic anchors. These findings are further corroborated by PCA and t-SNE analyses in Appendix H. This result confirms that our architecture does not merely process time series, but learns to organize them according to the rich, hierarchical information provided by our dynamically generated anchors, achieving true semantic-driven disentanglement.


    
    

\section{Conclusion}
In this work, we introduced STELLA, a novel framework that tackles the fundamental information enhancement bottleneck in adapting Large Language Models (LLMs) for time series forecasting. We reframe the task from generic sequence processing to a semantic-guided paradigm, where the key is to systematically inject the structured supplementary and complementary information that current methods lack. To this end, STELLA employs a unique dynamic semantic abstraction mechanism: in lieu of static, retrieval-based prompting, it first decouples the intrinsic patterns of a series and then generatively translates their behavioral signatures into Hierarchical Semantic Anchors---a global Corpus-level Semantic Prior (CSP) and an instance-specific Fine-grained Behavioral Prompt (FBP). Comprehensive experiments not only demonstrate STELLA's state-of-the-art performance across eight benchmarks but, critically, also its superior generalization in zero-shot and few-shot scenarios. Ultimately, STELLA validates a key thesis: the path to unlocking the full potential of LLMs for quantitative reasoning lies not in passive adaptation, but in actively constructing and conditioning on dynamic, structured semantic knowledge---a promising frontier for future research.\looseness=-1


%



%



\FloatBarrier 

\bibliographystyle{IEEEtran}
\bibliography{references}

\onecolumn

%
\ifCLASSINFOpdf
\else
\fi
\hyphenation{op-tical net-works semi-conduc-tor}

\clearpage
\appendices

\vspace*{0.4em} 

\section*{\huge \centering {STELLA: Guiding Large Language Models for Time Series Forecasting with Semantic Abstractions}}
\addcontentsline{toc}{section}{Appendix - Contents}  

\vspace{2.5em}  

\begin{center}
    {\Large \fontseries{b}\selectfont ————Appendix————}
\end{center}

\hypersetup{linkcolor=black}
\tableofcontents  
\hypersetup{linkcolor=blue!40}

\clearpage
\section{Showcases}
\label{sec:appendix_showcases}

This section presents a qualitative comparison to visually adjudicate the forecasting performance of our proposed model, STELLA, against the TimeLLM~\cite{jin2023time} baseline. The visualizations cover five benchmark datasets: ETTh1, ETTh2, ETTm1, ETTm2, and Exchange-Rate. Each case study utilizes a fixed input sequence of 96 steps to forecast two distinct prediction horizons: $H=96$ and $H=192$.

\begin{figure*}[htbp]
    \centering
    \includegraphics[width=\linewidth]{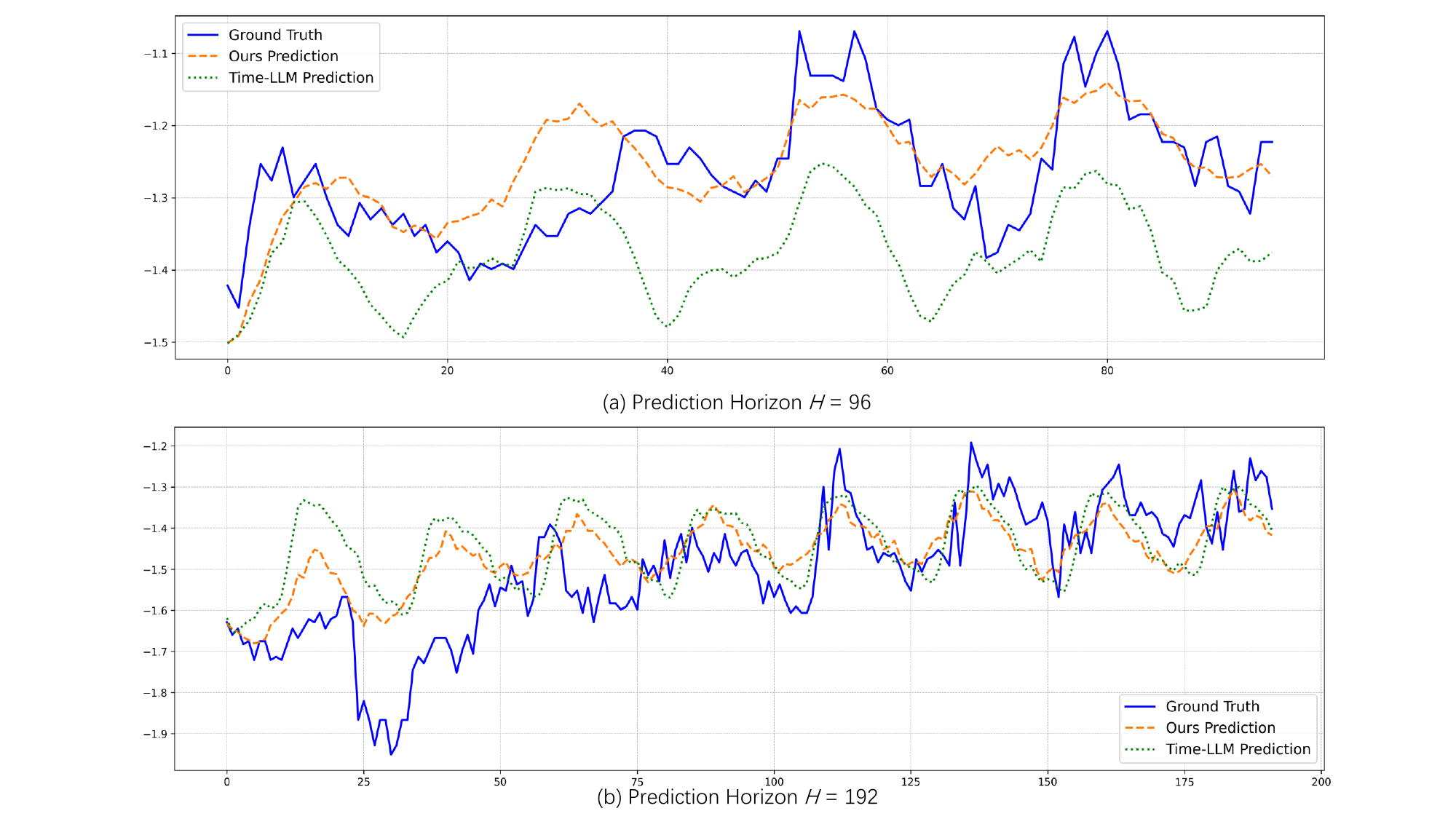}\\
    \caption{Qualitative forecasting results of our model (STELLA) against TimeLLM on the \textbf{ETTh1} dataset for prediction horizons $H=96$ (a) and $H=192$ (b).}
    \label{fig:showcase_etth1}
\end{figure*}

\begin{figure*}[htbp]
    \centering
    \includegraphics[width=\linewidth]{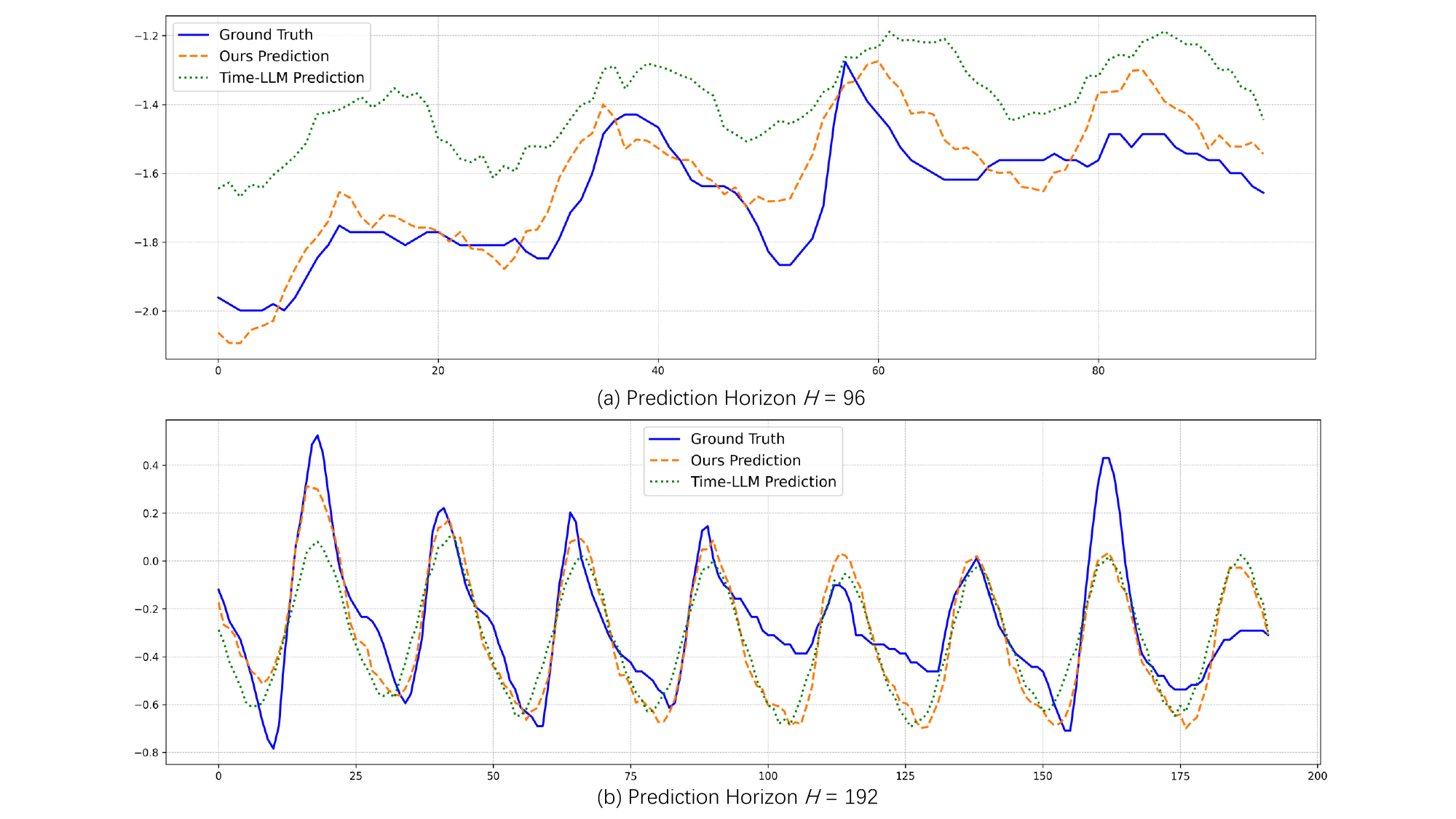}\\
    \caption{Qualitative forecasting results of our model (STELLA) against TimeLLM on the \textbf{ETTh2} dataset for prediction horizons $H=96$ (a) and $H=192$ (b).}
    \label{fig:showcase_etth2}
\end{figure*}

\begin{figure*}[htbp]
    \centering
    \includegraphics[width=\linewidth]{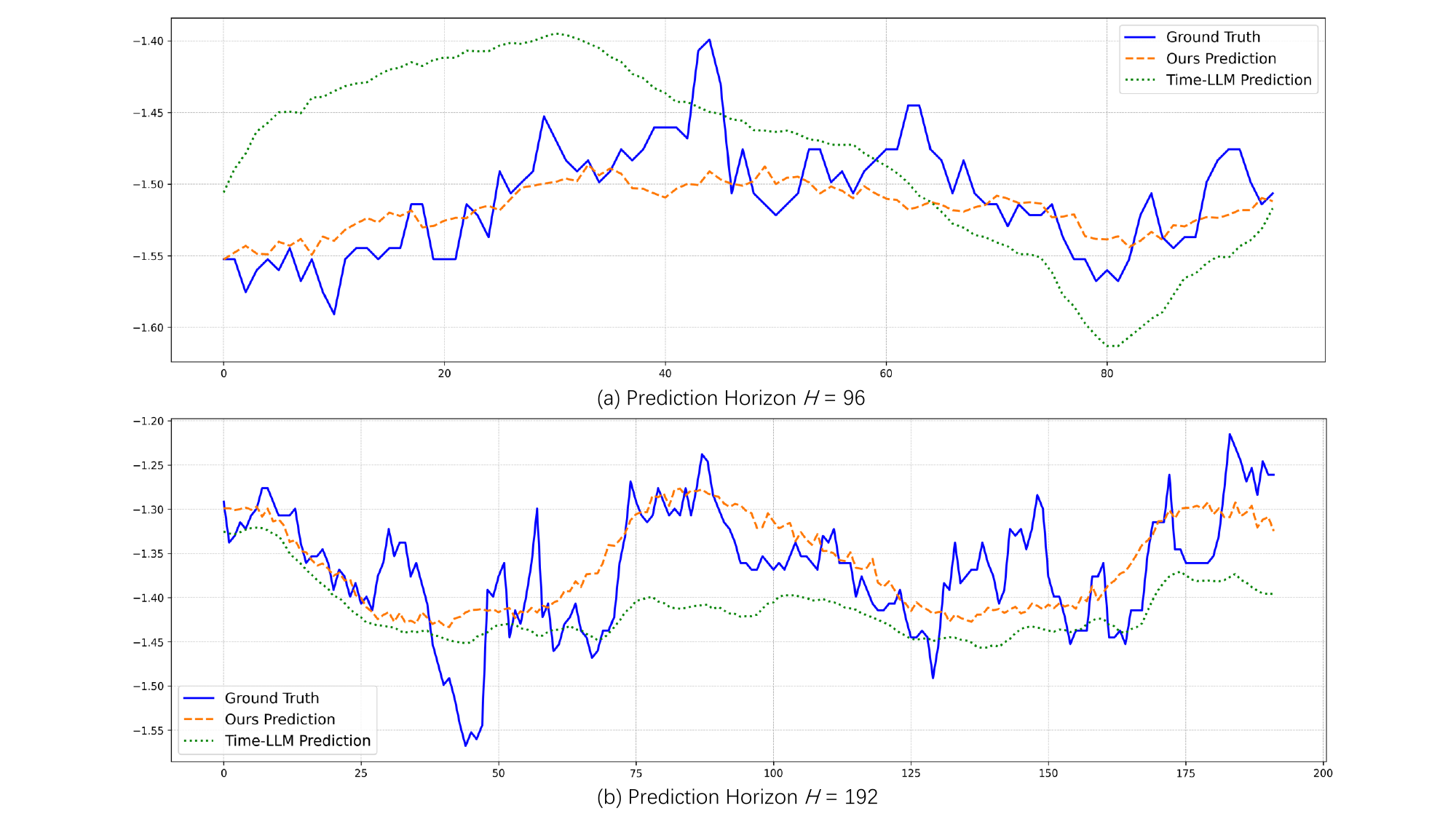}\\
    \caption{Qualitative forecasting results of our model (STELLA) against TimeLLM on the \textbf{ETTm1} dataset for prediction horizons $H=96$ (a) and $H=192$ (b).}
    \label{fig:showcase_ettm1}
\end{figure*}

\begin{figure*}[htbp]
    \centering
    \includegraphics[width=\linewidth]{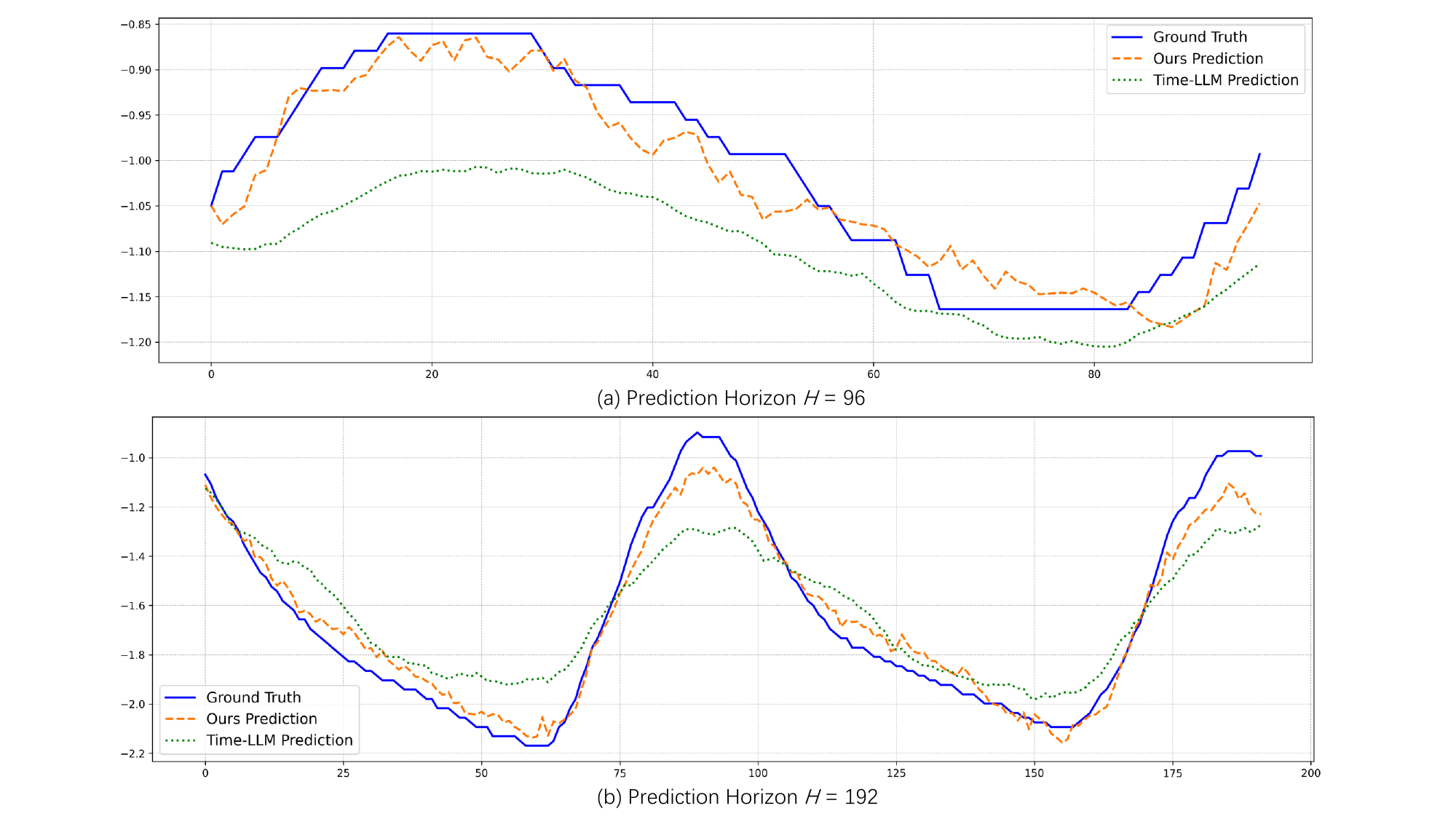}\\
    \caption{Qualitative forecasting results of our model (STELLA) against TimeLLM on the \textbf{ETTm2} dataset for prediction horizons $H=96$ (a) and $H=192$ (b).}
    \label{fig:showcase_ettm2}
\end{figure*}

\begin{figure*}[htbp]
    \centering
    \includegraphics[width=\linewidth]{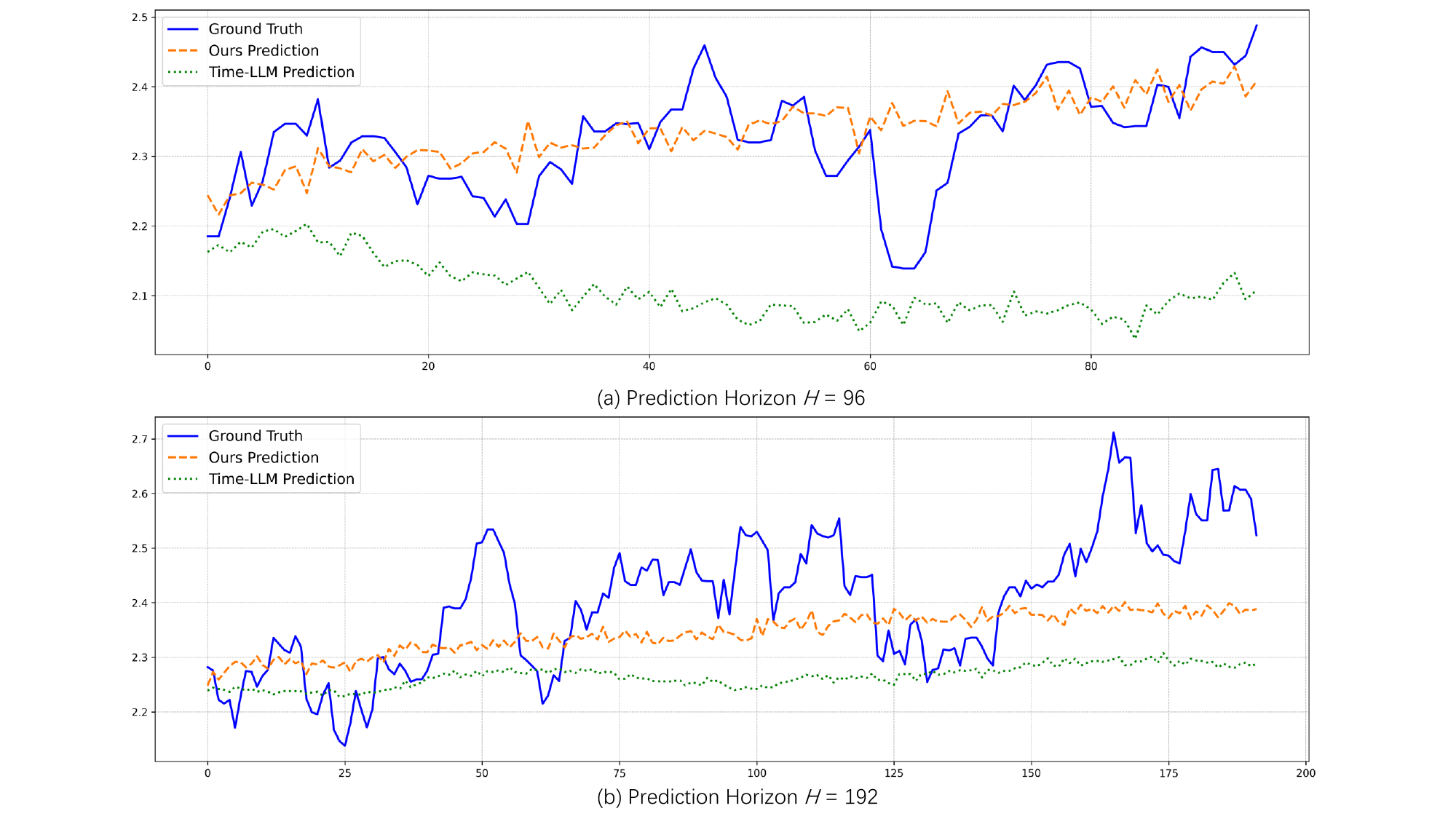}\\
    \caption{Qualitative forecasting results of our model (STELLA) against TimeLLM on the \textbf{Exchange-Rate} dataset for prediction horizons $H=96$ (a) and $H=192$ (b).}
    \label{fig:showcase_exchange}
\end{figure*}

The subsequent figures (Figures~\ref{fig:showcase_etth1} through~\ref{fig:showcase_exchange}) illustrate selected forecasting samples for a representative variate from each dataset. These visualizations underscore our model's superior capability to capture complex temporal dynamics—including long-term trends, periodicity, and abrupt shifts—in closer alignment with the ground truth. The consistent generation of more accurate and plausible forecasts across these diverse datasets, particularly over longer horizons, demonstrates the robustness and generalization capabilities of our proposed method.

\section{Experimental Details}
\subsection{Implementation}
\label{app:app_implementation}

We adopt the experimental setup proposed in \cite{wu2022timesnet}, and evaluate all baseline models using a unified evaluation framework provided by the official implementation (\url{https://github.com/thuml/Time-Series-Library}), ensuring fairness and consistency across comparisons.

For the backbone model, we utilize LLaMA2-7B \cite{touvron2023llama}, with only the first six transformer layers activated. Fine-tuning is performed via Low-Rank Adaptation (LoRA) \cite{hu2022lora}, where the rank is set to 4, the scaling factor (\(\alpha\)) to 8, and the dropout rate to 0.1. LoRA is applied specifically to the query and value projection layers within the self-attention mechanism, while all other model parameters remain frozen. In addition, we allow gradient updates for several normalization-related components, including \texttt{input\_layernorm}, \texttt{post\_attention\_layernorm}, and all LoRA-inserted modules.

LoRA is further applied to the \textbf{cross-attention layers within the Semantic Anchor Module (SAM)}, where a higher rank of 256 and a scaling factor of 512 are used, with the dropout rate fixed at 0.1. This adaptation targets both standard attention projections and their associated dimension-reduction layers, enabling more expressive and parameter-efficient prompt representations.

Moreover, we extend LoRA-based fine-tuning to the output MLP layers. In this stage, the rank is set to 64, the scaling factor to 128, and the dropout rate remains at 0.1. Fine-tuning is performed on the linear transformations within the residual blocks as well as the final output layer. To retain effective normalization, all RMSNorm layers are also unfrozen during training.

In our unified experimental configuration, the patch length is set to 16, with a stride of 16 to ensure non-overlapping segmentation. For our \textbf{Hierarchical Semantic Anchors}, the length of the \textbf{Corpus-level Semantic Prior (CSP)} is fixed at 10 tokens. The length of the \textbf{Fine-grained Behavioral Prompt (FBP)} is dynamically determined for each instance, calculated by dividing the input sequence length by the patch length.

Following the optimization protocol commonly adopted in Transformer-based models \cite{vaswani2017attention}, we incorporate a warmup phase to stabilize the initial stages of training. During this phase, the learning rate increases linearly from a small initial value to a target value over the first 4 epochs. After the warmup period, we employ an exponential decay schedule for the learning rate, gradually reducing it until training concludes. The total number of training epochs is capped at 100, with early stopping applied to prevent overfitting.

To enhance model robustness and mitigate overfitting, we incorporate dropout regularization. The dropout rate is treated as a tunable hyperparameter and is selected based on validation performance for each dataset or model configuration.

The optimization is performed using the Adam optimizer \cite{kingma2014adam}. All models are implemented in PyTorch \cite{paszke2019pytorch} and trained on multiple GPUs. Specifically, we conduct experiments on computing clusters equipped with 8 NVIDIA L20 GPUs (48GB each) and 8 NVIDIA A800 GPUs (80GB each), utilizing data parallelism to accelerate training.

\subsection{Baseline Introduction}

We introduce the baseline models selected for comparison in this study:

\begin{itemize}
    \item \textbf{GPT4TS~\cite{zhou2023one}}: GPT4TS represents time series data as patched tokens and fine-tunes a pre-trained GPT-2~\cite{radford2019language} to perform a variety of time series analysis tasks. This approach leverages the generative capacity of language models while adapting them to structured temporal data.

    \item \textbf{Time-LLM~\cite{jin2023time}}: Time-LLM performs time series analysis by fine-tuning pre-trained large language models (LLMs) through prefix prompting. Additionally, it employs a patch-based strategy that transforms time series data into discrete tokenized segments, enabling compatibility with natural language processing (NLP) input formats.

    \item \textbf{PatchTST~\cite{nie2022time}}: PatchTST is a Transformer-based model for time series forecasting that segments the input data into patches. It utilizes a channel-independent architecture to enhance predictive performance while significantly reducing computational complexity.

    \item \textbf{iTransformer~\cite{liu2023itransformer}}: iTransformer captures multivariate dependencies by applying attention mechanisms and feed-forward networks along the transposed dimensions of time series data. This design facilitates more effective modeling of cross-variable interactions.

    \item \textbf{Crossformer~\cite{zhang2023crossformer}}: Crossformer leverages a Two-Stage Attention mechanism and Hierarchical Encoder-Decoder architecture to jointly model temporal and variable dependencies in multivariate time series forecasting.

    \item \textbf{FEDformer~\cite{zhou2022fedformer}}: FEDformer enhances time series forecasting by incorporating seasonal-trend decomposition into the Transformer framework and leveraging frequency domain information to improve both prediction accuracy and computational efficiency.

    \item \textbf{Autoformer~\cite{wu2021autoformer}}: Autoformer introduces a decomposition-based architecture designed to achieve computational efficiency and high accuracy in long-term time series forecasting by employing auto-correlation mechanisms.
    
    \item \textbf{Informer~\cite{zhou2021informer}}: Informer proposes the ProbSparse self-attention mechanism and attention distillation strategy to efficiently model extensive input sequences, coupled with a generative-style decoder that significantly enhances inference efficiency in long-sequence time series forecasting tasks.

    \item \textbf{ETSformer~\cite{woo2022etsformer}}: ETSformer enhances time series forecasting by incorporating exponential smoothing, replacing conventional self-attention with specialized exponential smoothing attention and frequency-based attention mechanisms.

    \item \textbf{NSformer~\cite{liu2022non}}: NSformer introduces Series Stationarization and De-stationary Attention to effectively model non-stationary time series, enhancing forecasting accuracy by preserving intrinsic temporal dependencies.

    \item \textbf{Reformer~\cite{kitaev2020reformer}}: Reformer improves Transformer scalability on long sequences by using locality-sensitive hashing attention and reversible residual layers to reduce both computational and memory complexity while maintaining competitive performance.

    \item \textbf{TimesNet~\cite{wu2022timesnet}}: TimesNet reformulates 1D time series into 2D representations and employs an inception-based TimesBlock to jointly capture intra- and inter-period dependencies.

    \item \textbf{MICN~\cite{wang2023micn}}: MICN employs a multi-scale convolutional framework to efficiently model local and global dependencies in long-term time series forecasting with linear complexity.

    \item \textbf{Dlinear~\cite{zeng2023transformers}}: Dlinear adopts a decomposition-based framework that separates time series into trend and seasonal components, which are independently modeled using specialized linear layers.
    
    \item \textbf{TiDE~\cite{das2023long}}: TiDE is an MLP-based encoder-decoder architecture for long-term time-series forecasting, which combines the computational efficiency of linear models with the expressive capacity to capture covariate effects and non-linear temporal dependencies.

    \item \textbf{TCN~\cite{bai2018empirical}}: TCN is a convolutional architecture for sequence modeling that consistently outperforms standard recurrent networks such as LSTMs across a range of tasks, while providing longer effective memory and improved parallelism.

    \item \textbf{N-HiTS~\cite{challu2023nhits}}: N-HiTS is a long-horizon forecasting model that uses hierarchical interpolation and multi-rate sampling to achieve superior accuracy and efficiency over Transformer-based methods.    
    
    \item \textbf{N-BEATS~\cite{oreshkin2019n}}: N-BEATS is a deep residual architecture based on fully connected layers for univariate time series forecasting, offering high predictive accuracy, interpretability, and broad domain applicability without relying on time-series-specific inductive biases.
\end{itemize}

\subsection{Details of Dataset}

\subsubsection{Long-term Forecasting}

To rigorously assess the effectiveness of our approach in long-term forecasting scenarios, we conduct extensive experiments across seven widely used time series datasets. In accordance with the evaluation protocols proposed in \cite{wu2021autoformer, wu2022timesnet}, each dataset is chronologically partitioned into training, validation, and testing subsets. Specifically, a 6:2:2 split ratio is adopted for the ETT dataset, while a 7:1:2 ratio is employed for the remaining datasets. A detailed description of each dataset is presented below.

\begin{enumerate}[label=(\arabic*), resume]
  \item \textbf{ETT} (Electricity Transformer Temperature) \cite{zhou2021informer} is a dataset that records both temperature and power load measurements from electricity transformers located in two regions of China, covering the period from 2016 to 2018. It is available in two temporal resolutions: hourly (ETTh) and 15-minute intervals (ETTm). Specifically, the data are collected from two different electricity transformers, labeled as 1 and 2. Each transformer provides measurements at both temporal granularities, resulting in a total of four subsets: ETTm1, ETTm2, ETTh1, and ETTh2.
\end{enumerate}

\begin{enumerate}[label=(\arabic*), resume]
  \item \textbf{Weather} \cite{wu2021autoformer} dataset comprises 21 meteorological variables systematically collected across Germany over the course of 2020. With observations recorded every 10 minutes, it offers high temporal resolution for time-series analysis. Featuring key attributes such as air temperature and visibility, Weather provides a comprehensive foundation for modeling and understanding atmospheric variability and dynamics.
\end{enumerate}

\begin{enumerate}[label=(\arabic*), resume]
  \item \textbf{Exchange} dataset~\cite{lai2018modeling} is a widely used benchmark in time series research, particularly in exchange rate forecasting and economic-financial modeling. It contains daily exchange rates of eight foreign currencies against the US dollar from 1990 to 2010, comprising 7,588 time steps with eight variables—each representing a different currency. With its long time span and daily granularity, the dataset captures fine-grained fluctuations in currency values. Owing to its multi-currency structure and temporal depth, it is frequently employed to evaluate forecasting models such as ARIMA and neural networks. The \textit{Exchange} dataset provides a robust empirical foundation for analyzing exchange rate dynamics and macroeconomic trends, and holds significant value in econometrics and financial analytics.
\end{enumerate}

\begin{enumerate}[label=(\arabic*), resume]
  \item \textbf{Illness} dataset~\cite{cdc_fluview} constitutes a widely recognized benchmark for time series forecasting tasks in epidemiology, with a particular focus on modeling the temporal dynamics of influenza-like illness (ILI) in the United States. The dataset comprises 966 weekly observations collected over a 20-year period (2002–2021), and includes seven key epidemiological variables. Among these are age-specific case counts for individuals aged 0–4 and 5–24 years, as well as the aggregate number of ILI cases across all age groups. The availability of age-stratified data enables a nuanced analysis of disease transmission patterns across demographic cohorts, thereby enhancing the understanding of population-specific vulnerability and transmission dynamics. Due to its high temporal granularity and comprehensive coverage of critical public health indicators, the \textit{Illness} dataset is extensively utilized for forecasting seasonal disease trends, informing epidemic surveillance strategies, and evaluating the effectiveness of predictive models in public health policy and intervention planning.
\end{enumerate}

Detailed statistics for these datasets, including dimensionality, series lengths, dataset sizes, sampling frequency, and data types, are presented in Tab.~\ref{tab:longterm_dataset}.

As observed across datasets, there exists a significant imbalance in the number of available training samples. The ETTm1, ETTm2, and Weather datasets contain the largest amount of training data, each with over 30,000 samples. In contrast, ETTh1 and ETTh2 provide approximately 8,500 samples, Exchange has around 5,000, while the Illness dataset contains only about 600 training samples.

Following the experimental setup of \cite{liu2024unitime}, we adopt an oversampling strategy to mitigate the effect of such imbalance. Specifically, the Illness dataset is oversampled by a factor of 12 to alleviate data scarcity. The primary objective is to ensure that the model is sufficiently exposed to underrepresented domains and is not dominated by larger datasets during training. To maintain a fair comparison, the same oversampling configuration is consistently applied to all baseline models.

\begin{table*}[htbp]
    \centering
    \caption{The statistics of long-term forecasting datasets. The dimensionality refers to the number of time series variables, and the dataset sizes are structured as (training, validation, and testing).}
    \label{tab:longterm_dataset}
    \setlength{\heavyrulewidth}{1.1pt}  
    \setlength{\lightrulewidth}{0.5pt}  
    \setlength{\cmidrulewidth}{0.5pt}   
    \scriptsize
    \resizebox{\textwidth}{!}{  
    \begin{tabular}{c|c|c|c|c|cc}
        \toprule
        Datasets & Dim. & Series Length & Dataset Size & Frequency & Information \\
        \midrule
        ETTm1 & 7 & \{96, 192, 336, 720\} & (34465, 11521, 11521) & 15 min & Temperature \\
        ETTm2 & 7 & \{96, 192, 336, 720\} & (34465, 11521, 11521) & 15 min & Temperature \\
        ETTh1 & 7 & \{96, 192, 336, 720\} & (8545, 2881, 2881) & 1 hour & Temperature \\
        ETTh2 & 7 & \{96, 192, 336, 720\} & (8545, 2881, 2881) & 1 hour & Temperature \\
        Weather & 21 & \{96, 192, 336, 720\} & (36792, 5271, 10540) & 10 min & Weather \\
        Exchange & 8 & \{96, 192, 336, 720\} & (5120, 665, 1422) & 1 day & Finance \\
        Illness & 7 & \{24, 36, 48, 60\} & (617, 74, 170) & 7 day & Health \\
        \bottomrule
    \end{tabular}
    }
\end{table*}

\subsubsection{Short-term Forecasting}
  The \textbf{M4} benchmark \cite{makridakis2018m4} encompasses a collection of 100,000 time series drawn from a broad spectrum of application domains, ranging from business operations to economic forecasting. These time series are systematically organized into six distinct groups, each associated with a specific sampling frequency, spanning from annual to hourly intervals. Comprehensive statistical properties—such as the number of time steps and the temporal resolutions—are summarized in Table~\ref{tab:shortterm_dataset}.

  \begin{table*}[htbp]
    \centering
    \caption{The statistics of short-term forecasting datasets. The dimensionality refers to the number of time series variables, and the dataset sizes are structured as (training, validation, and testing).}
    \label{tab:shortterm_dataset} 
    \setlength{\heavyrulewidth}{1.1pt}  
    \setlength{\lightrulewidth}{0.5pt}  
    \setlength{\cmidrulewidth}{0.5pt}   
    \scriptsize
    \resizebox{\textwidth}{!}{  
    \begin{tabular}{c|c|c|c|c|cc}
        \toprule
        \textbf{Datasets} & \textbf{Dim.} & \textbf{Series Length} & \textbf{Dataset Size} & \textbf{Frequency} & \textbf{Information} \\
        \midrule
        M4-Yearly     & 1 & 6  & (23000, 0, 23000) & Yearly     & Demographic \\
        M4-Quarterly  & 1 & 8  & (24000, 0, 24000) & Quarterly  & Finance     \\
        M4-Monthly    & 1 & 18 & (48000, 0, 48000) & Monthly    & Industry    \\
        M4-Weekly     & 1 & 13 & (359, 0, 359)     & Weekly     & Macro       \\
        M4-Daily      & 1 & 14 & (4227, 0, 4227)   & Daily      & Micro       \\
        M4-Hourly     & 1 & 48 & (414, 0, 414)     & Hourly     & Other       \\
        \bottomrule
    \end{tabular}
    }
    \label{tab:m4_summary}
\end{table*}

\subsubsection{Few/Zero-shot Learning}

In the few- and zero-shot learning experiments, we adopt the same four datasets from the ETT collection as those used in the long-term forecasting study: ETTm1, ETTm2, ETTh1, and ETTh2.

\subsection{Evaluation Metrics}

To assess forecasting performance, we employ standard error-based metrics for both long-term and short-term forecasting tasks.

For long-term forecasting, we adopt the \textbf{mean squared error} (\textbf{MSE}) and \textbf{mean absolute error} (\textbf{MAE}). The reported values are normalized, accounting for global data standardization. Notably, the normalization scaler is computed exclusively from the training data.

For short-term forecasting on the \textbf{M4} benchmark, we follow the official evaluation protocol and utilize three commonly used metrics: \textbf{symmetric mean absolute percentage error} (\textbf{SMAPE}), \textbf{mean absolute scaled error} (\textbf{MASE}), and \textbf{overall weighted average} (\textbf{OWA}), as defined in \cite{oreshkin2019n}. Among them, \textbf{OWA} is a composite indicator specifically proposed for the M4 competition to jointly evaluate prediction accuracy and robustness.

The formal definitions of these metrics are given as follows:

\begin{align*}
\text{MSE} &= \frac{1}{H} \sum_{h=1}^H \left( \mathbf{Y}_h - \hat{\mathbf{Y}}_h \right)^2, \\
\text{MAE} &= \frac{1}{H} \sum_{h=1}^H \left| \mathbf{Y}_h - \hat{\mathbf{Y}}_h \right|, \\
\text{SMAPE} &= \frac{200}{H} \sum_{h=1}^H \frac{ \left| \mathbf{Y}_h - \hat{\mathbf{Y}}_h \right| }{ \left| \mathbf{Y}_h \right| + \left| \hat{\mathbf{Y}}_h \right| }, \\
\text{MAPE} &= \frac{100}{H} \sum_{h=1}^H \frac{ \left| \mathbf{Y}_h - \hat{\mathbf{Y}}_h \right| }{ \left| \mathbf{Y}_h \right| }, \\
\text{MASE} &= \frac{1}{H} \sum_{h=1}^H \frac{ \left| \mathbf{Y}_h - \hat{\mathbf{Y}}_h \right| }{ \frac{1}{H-s} \sum_{j=s+1}^H \left| \mathbf{Y}_j - \mathbf{Y}_{j-s} \right| }, \\
\text{OWA} &= \frac{1}{2} \left[ \frac{\text{SMAPE}}{\text{SMAPE}_{\text{Na\"ive2}}} + \frac{\text{MASE}}{\text{MASE}_{\text{Na\"ive2}}} \right],
\end{align*}

\noindent
where \( s \) denotes the periodicity of the time series data, and \( H \) represents the prediction horizon. \( \mathbf{Y}_h \) and \( \hat{\mathbf{Y}}_h \) denote the ground truth and the predicted value at time step \( h \in \{1, 2, \dots, H\} \), respectively.

\section{Full Long-term Forecasting Results}
\label{app:app_full_long_term_results}

Due to the limited space in the main text, a more comprehensive comparison of baseline models is provided in Table~\ref{tab:full_results_part1} and its continuation Table~\ref{tab:full_results_part2}. We carefully select a diverse set of representative baselines from recent advances in time series forecasting, covering the following categories: LLM-based models, including GPT4TS \cite{zhou2023one} and TimeLLM \cite{jin2023time}; Transformer-based models, including PatchTST \cite{nie2022time}, iTransformer \cite{liu2023itransformer}, Crossformer \cite{zhang2023crossformer}, FEDformer \cite{zhou2022fedformer}, Autoformer \cite{wu2021autoformer}, Informer \cite{zhou2021informer}, NSformer \cite{liu2022non}, and Reformer \cite{kitaev2020reformer}; CNN-based models, including TimesNet \cite{wu2022timesnet} and MICN \cite{wang2023micn}; and MLP-based models, including DLinear \cite{zeng2023transformers} and TiDE \cite{das2023long}.

\begin{table*}[htbp]
    \centering
    \caption{Full multivariate long-term forecasting results for all datasets across different prediction lengths. The input sequence length $T$ is set to 36 for the Illness dataset and 96 for all others. The prediction lengths are set to $\{24, 36, 48, 60\}$ for Illness, and $\{96, 192, 336, 720\}$ for the remaining datasets. The table reports Mean Squared Error (MSE) and Mean Absolute Error (MAE) for each method at each forecasting horizon. \textbf{Avg.} represents the average performance over all four prediction horizons. The \textbf{\textcolor{red}{best}} and \textcolor{softblue}{\uline{second-best}} results for each forecasting horizon are highlighted in \textbf{\textcolor{red}{bold}} and \textcolor{softblue}{\uline{underlined}}, respectively. The \textbf{1st Count} column reports the number of times each method achieves the \textbf{\textcolor{red}{best}} performance. (Part 1)}
    \label{tab:full_results_part1}
    \setlength{\heavyrulewidth}{1.1pt}  
    \setlength{\lightrulewidth}{0.5pt}  
    \setlength{\cmidrulewidth}{0.5pt}   
    \resizebox{\textwidth}{!}{  
    \begin{tabular}{c|c|cc|cc|cc|cc|cc|cc|cc|cc}
    \toprule
        \multicolumn{2}{c}{Categories}  & \multicolumn{6}{c}{LLM-Based}  & \multicolumn{10}{c}{Transformer-based}  \\
        \toprule
        \multicolumn{2}{c}{Models} & \multicolumn{2}{c}{\makecell{\textbf{STELLA}\\\textbf{(Ours)}}}  & \multicolumn{2}{c}{\makecell{GPT4TS\\ \cite{zhou2023one}}}  & \multicolumn{2}{c}{\makecell{TimeLLM\\ \cite{jin2023time}}}   & \multicolumn{2}{c}{\makecell{PatchTST\\ \cite{nie2022time}}}  &   \multicolumn{2}{c}{\makecell{iTransformer\\ \cite{liu2023itransformer}}}  & \multicolumn{2}{c}{\makecell {Crossformer\\ \cite{zhang2023crossformer}}} &  \multicolumn{2}{c}{\makecell{FEDformer\\ \cite{zhou2022fedformer}}} & \multicolumn{2}{c}{\makecell{Autoformer\\ \cite{wu2021autoformer}}}   \\
        \toprule
        \multicolumn{2}{c}{Metric} & MSE & MAE & MSE & MAE & MSE & MAE & MSE & MAE & MSE & MAE & MSE & MAE & MSE & MAE & MSE & MAE \\
        \toprule
        \multirow{5}{*}{\rotatebox{90}{ETTm1}} & 96 &  \textbf{\textcolor{red}{0.310}}  & \textbf{\textcolor{red}{0.343}}  & 0.335  & 0.369  & 0.359  & 0.381  & 0.344  & 0.373  & 0.341  & 0.376  & 0.360  & 0.401  & 0.379  & 0.419  & 0.505  & 0.475 \\ 
        ~ & 192 & \textbf{\textcolor{red}{0.358}} & \textbf{\textcolor{red}{0.371}}  & 0.374  & \textbf{\textcolor{red}{0.385}}  & 0.383  & 0.393  & 0.367  & \textcolor{softblue}{\uline{0.386}} & 0.382  & 0.395  & 0.402  & 0.440  & 0.426  & 0.441  & 0.553  & 0.496 \\ 
        ~ & 336 & \textbf{\textcolor{red}{0.391}}  & \textbf{\textcolor{red}{0.404}}   & 0.407  &  \textcolor{softblue}{\uline{0.406}}  & 0.416  & 0.414  &  \textcolor{softblue}{\uline{0.392}} & 0.407  & 0.418  & 0.418  & 0.543  & 0.528  & 0.445  & 0.459  & 0.621  & 0.537 \\
        ~ & 720 & \textbf{\textcolor{red}{0.457}} &  \textbf{\textcolor{red}{0.432}} & 0.469  &  \textcolor{softblue}{\uline{0.442}}  & 0.483  & 0.449  & \textcolor{softblue}{\uline{0.464}} & \textcolor{softblue}{\uline{0.442}} & 0.487  & 0.456  & 0.704  & 0.642  & 0.543  & 0.490  & 0.671  & 0.561 \\
        \cmidrule(lr){2-18} 
        ~ & Avg. & \textbf{\textcolor{red}{0.379}}  & \textbf{\textcolor{red}{0.388}}  & 0.396  & \textcolor{softblue}{\uline{0.401}}  & 0.410  & 0.409  & \textcolor{softblue}{\uline{0.392}} & 0.402  & 0.407  & 0.411  & 0.502  & 0.503  & 0.448  & 0.452  & 0.588  & 0.517  \\
        \midrule
        \multirow{5}{*}{\rotatebox{90}{ETTm2}} & 96 &  \textbf{\textcolor{red}{0.171}}  & \textbf{\textcolor{red}{0.249}}  & 0.190  & 0.275  & 0.193  & 0.280  & \textbf{\textcolor{red}{0.177}}  & \textcolor{softblue}{\uline{0.260}}  & 0.185  & 0.272  & 0.273  & 0.356  & 0.203  & 0.287  & 0.255  & 0.339  \\
        ~ & 192 &  \textbf{\textcolor{red}{0.236}} &  \textbf{\textcolor{red}{0.293}}  & 0.253  & 0.313  & 0.257  & 0.318  &  \textcolor{softblue}{\uline{0.246}}  & 0.305  & 0.253  & 0.313  & 0.426  & 0.487  & 0.269  & 0.328  & 0.281  & 0.340 \\
        ~ & 336 & \textbf{\textcolor{red}{0.296}}  & \textcolor{softblue}{\uline{0.331}} & 0.321  & \textbf{\textcolor{red}{0.320}}   & 0.317  & 0.353  & 0.305  & 0.343  & 0.315  & 0.350  & 1.013  & 0.714  & 0.325  & 0.366  & 0.339  & 0.372  \\
        ~ & 720 &  \textbf{\textcolor{red}{0.397}} &   \textbf{\textcolor{red}{0.396}}   & 0.411  & 0.406  & 0.419  & 0.411  & 0.410  & 0.405  & 0.413  & 0.406  & 3.154  & 1.274  & 0.421  & 0.415  & 0.433  & 0.432 \\
        \cmidrule(lr){2-18} 
        ~ & Avg. &  \textbf{\textcolor{red}{0.275}}  &  \textbf{\textcolor{red}{0.317}}  & 0.294  & 0.329  & 0.297  &  0.341  &   \textcolor{softblue}{\uline{0.285}}  & 0.328  & 0.292  & 0.335  & 1.217  & 0.708  & 0.305  & 0.349  & 0.327  & 0.371  \\
        \midrule
        \multirow{5}{*}{\rotatebox{90}{ETTh1}} & 96 & \textbf{\textcolor{red}{0.368}}  & \textbf{\textcolor{red}{0.390}}  & 0.398  & 0.424  & 0.398  & 0.410  & 0.404  & 0.413  & 0.386  & 0.404  & 0.420  & 0.439  &  \textcolor{softblue}{\uline{0.376}} & 0.419  & 0.449  & 0.459  \\
        ~ & 192 & \textbf{\textcolor{red}{0.406}}  & \textbf{\textcolor{red}{0.421}} & 0.449  & 0.427  & 0.451  & 0.440  & 0.454  & 0.440  & 0.441  & 0.436  & 0.540  & 0.519  &  \textcolor{softblue}{\uline{0.420}} & 0.448  & 0.500  & 0.482 \\        
        ~ & 336 & \textbf{\textcolor{red}{0.446}}  & \textbf{\textcolor{red}{0.442}} & 0.492  & 0.466  & 0.508  & 0.471  & 0.497  & 0.462  & 0.489  & 0.461  & 0.722  & 0.648  & \textcolor{softblue}{\uline{0.459}} & 0.465  & 0.521  & 0.496  \\
        ~ & 720 &  \textbf{\textcolor{red}{0.445}}  & \textbf{\textcolor{red}{0.448}} & 0.487  & 0.483  & 0.483  & 0.478  & 0.496  & 0.481  & 0.508  & 0.493  & 0.799  & 0.685  & 0.506  & 0.507  & 0.514  & 0.512  \\
        \cmidrule(lr){2-18} 
        ~ & Avg. & \textbf{\textcolor{red}{0.416}} & \textbf{\textcolor{red}{0.425}}  & 0.457  & 0.450  & 0.460  & 0.449  & 0.463  & 0.449  & 0.455  & 0.448  & 0.620  & 0.572  & \textcolor{softblue}{\uline{0.440}} & 0.460  & 0.496  & 0.487 \\
        \midrule
        \multirow{5}{*}{\rotatebox{90}{ETTh2}} & 96 &  \textbf{\textcolor{red}{0.292}}   & \textbf{\textcolor{red}{0.336}}  & 0.312  & 0.360  & \textcolor{softblue}{\uline{0.295}}  &  \textcolor{softblue}{\uline{0.348}}  & 0.312  & 0.358  & 0.300  & 0.349  & 0.745  & 0.584  & 0.358  & 0.397  & 0.346  & 0.388 \\
        ~ & 192 & \textbf{\textcolor{red}{0.362}}  & \textbf{\textcolor{red}{0.383}}  & 0.387  & 0.405  & 0.386  & 0.404  & 0.397  & 0.408  & \textcolor{softblue}{\uline{0.379}} & \textcolor{softblue}{\uline{0.398}}  & 0.877  & 0.656  & 0.429  & 0.439  & 0.456  & 0.452\\
        ~ & 336 & \textbf{\textcolor{red}{0.388}}  &  \textbf{\textcolor{red}{0.414}} & 0.424  & 0.437  & 0.421  & 0.435  & 0.435  & 0.440  & \textcolor{softblue}{\uline{0.418}}  & \textcolor{softblue}{\uline{0.429}}  & 1.043  & 0.731  & 0.496  & 0.487  & 0.482  & 0.486 \\
        ~ & 720 & \textbf{\textcolor{red}{0.414}}  & \textbf{\textcolor{red}{0.433}}  & 0.433  & 0.453  &  \textcolor{softblue}{\uline{0.422}}   &  \textcolor{softblue}{\uline{0.445}}  & 0.436  & 0.449  & 0.428  & 0.445  & 1.104  & 0.763  & 0.463  & 0.474  & 0.515  & 0.511   \\
        \cmidrule(lr){2-18} 
        ~ & Avg. & \textbf{\textcolor{red}{0.364}} & \textbf{\textcolor{red}{0.391}}  & 0.389  & 0.414  & \textcolor{softblue}{\uline{0.381}}   & 0.408  & 0.395  & 0.414  & \textcolor{softblue}{\uline{0.381}}  & \textcolor{softblue}{\uline{0.405}}  & 0.942  & 0.684  & 0.437  & 0.449  & 0.450  & 0.459 \\
        \midrule
        \multirow{5}{*}{\rotatebox{90}{Weather}}  & 96 & \textcolor{softblue}{\uline{0.161}}  &  \textbf{\textcolor{red}{0.198}}  & 0.203  & 0.244  & 0.182  & 0.223  & 0.177  & 0.218  & 0.174  & \textcolor{softblue}{\uline{0.214}} & \textbf{\textcolor{red}{0.158}}  & 0.230  & 0.217  & 0.296  & 0.266  & 0.336 \\
        ~ & 192 & \textcolor{softblue}{\uline{0.207}}  & \textbf{\textcolor{red}{0.241}} & 0.247  & 0.277  & 0.231  & 0.263  & 0.222  & 0.259  & 0.221  &  \textcolor{softblue}{\uline{0.254}} & \textbf{\textcolor{red}{0.206}}  & 0.277  & 0.276  & 0.336  & 0.307  & 0.367  \\
        ~ & 336 & \textbf{\textcolor{red}{0.263}}  & \textbf{\textcolor{red}{0.285}}   & 0.297  & 0.311  & 0.283  & 0.300  & 0.277  & 0.297  & 0.278  &  \textcolor{softblue}{\uline{0.296}} & \textcolor{softblue}{\uline{0.272}} & 0.335  & 0.339  & 0.380  & 0.359  & 0.395 \\
        ~ & 720 & \textcolor{softblue}{\uline{0.341}} & \textbf{\textcolor{red}{0.335}}  & 0.368  & 0.356  & 0.360  & 0.350  & 0.352  & \textcolor{softblue}{\uline{0.347}}  & 0.358  & 0.349  & 0.398  & 0.418  & 0.403  & 0.428  & 0.419  & 0.428 \\
        \cmidrule(lr){2-18} 
        ~ & Avg. & \textcolor{softblue}{\uline{0.243}}   & \textbf{\textcolor{red}{0.265}}  & 0.279  & 0.297  & 0.264  & 0.284  & 0.257  & 0.280  & 0.257  & \textcolor{softblue}{\uline{0.279}}   & 0.259  & 0.315  & 0.309  & 0.360  & 0.338  & 0.382  \\
        \midrule
        \multirow{5}{*}{\rotatebox{90}{Exchange}} & 96 & \textbf{\textcolor{red}{0.084}}  &  \textbf{\textcolor{red}{0.204}}   & 0.091  & 0.212  & 0.087* & \textcolor{softblue}{\uline{0.206*}}& 0.109  & 0.236  & \textcolor{softblue}{\uline{0.086}}  & \textcolor{softblue}{\uline{0.206}}  & 0.232* & 0.359* & 0.148  & 0.278  & 0.197  & 0.323   \\
        ~ & 192 &  \textbf{\textcolor{red}{0.172}}   &  \textbf{\textcolor{red}{0.298}} & 0.183  & 0.304  & 0.179* & 0.302* & 0.205  & 0.327  & 0.177  &   \textcolor{softblue}{\uline{0.299}}  & 0.439* & 0.509* & 0.271  & 0.380  & 0.300  & 0.369\\
        ~ & 336 & 0.315  & \textbf{\textcolor{red}{0.407}}  & 0.328  & \textbf{\textcolor{red}{0.407}}  & 0.332* & 0.417* & 0.356  & 0.436  & 0.331  & 0.417  & 0.906* & 0.800* & 0.460  & 0.500  & 0.509  & 0.524  \\
        ~ & 720 & \textcolor{softblue}{\uline{0.785}}   & \textcolor{softblue}{\uline{0.666}}   & 0.880  & 0.704  & 0.837* & 0.688* & 0.888  & 0.716  & 0.847  & 0.691  & 1.332* & 0.943* & 1.195  & 0.841  & 1.447  & 0.941  \\ 
        \cmidrule(lr){2-18} 
        ~ & Avg. &  \textcolor{softblue}{\uline{0.339}}   & \textbf{\textcolor{red}{0.394}}    & 0.371  & 0.409  & 0.359* & 0.403* & 0.390  & 0.429  & 0.360  & 0.403  & 0.727* & 0.653* & 0.519  & 0.500  & 0.613  & 0.539 \\ 
        \midrule
        \multirow{5}{*}{\rotatebox{90}{Illness}} & 24 &  \textbf{\textcolor{red}{1.844}}   & \textbf{\textcolor{red}{0.813}}     & 2.732  & 1.100  & 2.058* & 0.890* & 2.335  & 0.989  & 2.181* & \textcolor{softblue}{\uline{0.907*}} & 4.322* & 1.429* & 3.228  & 1.260  & 3.483  & 1.287  \\
        ~ & 36 & 1.979 &  \textbf{\textcolor{red}{0.847}} & 2.664  & 1.063  & 2.099* & 0.889* & 2.561  & 1.035  & 2.000* & 0.901* & 4.321* & 1.408* & 2.679  & 1.080  & 3.103  & 1.148   \\
        ~ & 48 & \textbf{\textcolor{red}{1.769}} & \textbf{\textcolor{red}{0.806}}   & 2.617  & 1.041  & \textcolor{softblue}{\uline{1.954*}}  & \textcolor{softblue}{\uline{0.860*}}  & 2.465  & 1.022  & 2.388* & 1.005* & 4.492* & 1.429* & 2.622  & 1.078  & 2.669  & 1.085   \\ 
        ~ & 60 & \textbf{\textcolor{red}{1.683}}  &\textbf{\textcolor{red}{0.790}}   & 2.478  & 1.035  & \textcolor{softblue}{\uline{1.890*}}  & \textcolor{softblue}{\uline{0.881*}} & 2.189  & 0.997  & 2.162* & 0.959* & 4.590* & 1.450* & 2.857  & 1.157  & 2.770  & 1.125 \\
        \cmidrule(lr){2-18} 
        ~ & Avg. &  \textbf{\textcolor{red}{1.819}}  & \textbf{\textcolor{red}{0.814}}   & 2.623  & 1.060  & \textcolor{softblue}{\uline{2.000*}}   &  \textcolor{softblue}{\uline{0.880*}}   & 2.388  & 1.011  & 2.183* & 0.943* & 4.431* & 1.429* & 2.847  & 1.144  & 3.006  & 1.161 \\ 
        \midrule
        \multicolumn{2}{c}{1\textsuperscript{st}Count} & \multicolumn{2}{c}{\textbf{\textcolor{red}{60}}} & \multicolumn{2}{c|}{3}  & \multicolumn{2}{c}{0}  &  \multicolumn{2}{c}{1} &  \multicolumn{2}{c}{0}  &  \multicolumn{2}{c}{2}  &  \multicolumn{2}{c}{0}  &  \multicolumn{2}{c}{0}  \\ 
        \bottomrule
      \end{tabular}}
\vspace{2pt}
\parbox{\textwidth}{\scriptsize\textit{* indicates that the model did not report results on this dataset in the original paper. The results were reproduced using the code provided by the authors, ensuring that the experimental settings (including hyperparameters, data splits, etc.) were consistent with those described in the original paper. Other results are from TimesNet \cite{wu2022timesnet} and iTransformer \cite{liu2023itransformer}.}}
\end{table*}

\begin{table*}[htbp]
    \centering
    \caption{Full multivariate long-term forecasting results (continued). All settings are the same as in Table~\ref{tab:full_results_part1}. (Part 2)}
    \label{tab:full_results_part2}
    \setlength{\heavyrulewidth}{1.1pt}  
    \setlength{\lightrulewidth}{0.5pt}  
    \setlength{\cmidrulewidth}{0.5pt}   
    \resizebox{\textwidth}{!}{  
    \begin{tabular}{c|c|cc|cc|cc|cc|cc|cc|cc|cc}
    \toprule
        \multicolumn{2}{c}{Categories}  & \multicolumn{2}{c}{LLM-Based}  & \multicolumn{8}{c}{Transformer-based}  & \multicolumn{4}{c}{CNN-based} & \multicolumn{2}{c}{MLP-based}  \\
        \toprule
        \multicolumn{2}{c}{Models} & \multicolumn{2}{c}{\makecell{\textbf{STELLA}\\\textbf{(Ours)}}}   &\multicolumn{2}{c}{\makecell{Informer\\ \cite{zhou2021informer}}}   & \multicolumn{2}{c}{\makecell{NSformer\\ \cite{liu2022non}}}  & \multicolumn{2}{c}{\makecell{Reformer\\ \cite{kitaev2020reformer}}}  &  \multicolumn{2}{c}{\makecell{TimesNet\\ \cite{wu2022timesnet}}}  & \multicolumn{2}{c}{\makecell{MICN \\ \cite{wang2023micn}}} & \multicolumn{2}{c}{\makecell{DLinear\\ \cite{zeng2023transformers}}}& \multicolumn{2}{c}{\makecell{TiDE\\ \cite{das2023long}}}  \\ 
        \toprule
        \multicolumn{2}{c}{Metric} & MSE & MAE & MSE & MAE & MSE & MAE & MSE & MAE & MSE & MAE & MSE & MAE & MSE & MAE & MSE & MAE \\ 
        \toprule
        \multirow{5}{*}{\rotatebox{90}{ETTm1}} & 96 & \textbf{\textcolor{red}{0.310}}  & \textbf{\textcolor{red}{0.343}}  & 0.672  & 0.571  & 0.386  & 0.398  & 0.538  & 0.528  & 0.338  & 0.375  &  \textcolor{softblue}{\uline{0.316}} & \textcolor{softblue}{\uline{0.362}}  & 0.345  & 0.372  & 0.352  & 0.373  \\ 
        ~ & 192 & \textbf{\textcolor{red}{0.358}} & \textbf{\textcolor{red}{0.371}}   & 0.795  & 0.669  & 0.459  & 0.444  & 0.658  & 0.592  & 0.374  & 0.387  &  \textcolor{softblue}{\uline{0.363}}  & 0.390  & 0.380  & 0.389  & 0.389  & 0.391  \\ 
        ~ & 336 & \textbf{\textcolor{red}{0.391}}  & \textbf{\textcolor{red}{0.404}}  & 1.212  & 0.871  & 0.495  & 0.464  & 0.898  & 0.721  & 0.410  & 0.411  & 0.408  & 0.426  & 0.413  & 0.413  & 0.423  & 0.413  \\ 
        ~ & 720 &  \textbf{\textcolor{red}{0.457}} &  \textbf{\textcolor{red}{0.432}}   & 1.166  & 0.823  & 0.585  & 0.516  & 1.102  & 0.841  & 0.478  & 0.450  & 0.481  & 0.476  & 0.474  & 0.453  & 0.485  & 0.448  \\ 
        \cmidrule(lr){2-18} 
        ~ & Avg. & \textbf{\textcolor{red}{0.379}}  & \textbf{\textcolor{red}{0.388}} & 0.961  & 0.734  & 0.481  & 0.456  & 0.799  & 0.671  & 0.400  & 0.406  & \textcolor{softblue}{\uline{0.392}}   & 0.414  & 0.403  & 0.407  & 0.412  & 0.406  \\
        \midrule
        \multirow{5}{*}{\rotatebox{90}{ETTm2}} & 96 & \textbf{\textcolor{red}{0.171}}  & \textbf{\textcolor{red}{0.249}}  & 0.365  & 0.453  & 0.192  & 0.274  & 0.658  & 0.619  & 0.187  & 0.267  & \textcolor{softblue}{\uline{0.179}}  & 0.275  & 0.193  & 0.292  & 0.181  & 0.264  \\
        ~ & 192 & \textbf{\textcolor{red}{0.236}} &  \textbf{\textcolor{red}{0.293}}   & 0.533  & 0.563  & 0.280  & 0.339  & 1.078  & 0.827  & 0.249  & 0.309  & 0.307  & 0.376  & 0.284  & 0.362  & \textcolor{softblue}{\uline{0.246}}  &  \textcolor{softblue}{\uline{0.304 }} \\
        ~ & 336 & \textbf{\textcolor{red}{0.296}}  & \textcolor{softblue}{\uline{0.331}} & 1.363  & 0.887  & 0.334  & 0.361  & 1.549  & 0.972  & 0.321  & 0.351  & 0.325  & 0.388  & 0.369  & 0.427  &  \textcolor{softblue}{\uline{0.307}} & 0.341  \\
        ~ & 720 &  \textbf{\textcolor{red}{0.397}} &   \textbf{\textcolor{red}{0.396}}   & 3.379  & 1.338  & 0.417  & 0.413  & 2.631  & 1.242  & 0.408  & 0.403  & 0.502  & 0.490  & 0.554  & 0.522  & \textcolor{softblue}{\uline{0.407}} & \textcolor{softblue}{\uline{0.397}} \\
        \cmidrule(lr){2-18} 
        ~ & Avg. & \textbf{\textcolor{red}{0.275}}  &  \textbf{\textcolor{red}{0.317}}  & 1.410  & 0.810  & 0.306  & 0.347  & 1.479  & 0.915  & 0.291  & 0.333  & 0.328  & 0.382  & 0.350  & 0.401  & 0.289  &   \textcolor{softblue}{\uline{0.326}}   \\
        \midrule
        \multirow{5}{*}{\rotatebox{90}{ETTh1}} & 96 & \textbf{\textcolor{red}{0.368}}  & \textbf{\textcolor{red}{0.390}}   & 0.865  & 0.713  & 0.513  & 0.491  & 0.837  & 0.728  & 0.384  & 0.402  & 0.421  & 0.431  & 0.386  & 0.400  & 0.384  & \textcolor{softblue}{\uline{0.393}}  \\ 
        ~ & 192 &  \textbf{\textcolor{red}{0.406}}  & \textbf{\textcolor{red}{0.421}}  & 1.008  & 0.792  & 0.534  & 0.504  & 0.923  & 0.766  & 0.436  & 0.429  & 0.474  & 0.487  & 0.437  & 0.432  & 0.436  & \textcolor{softblue}{\uline{0.422}}   \\ 
        ~ & 336 & \textbf{\textcolor{red}{0.446}}  & \textbf{\textcolor{red}{0.442}}   & 1.107  & 0.809  & 0.588  & 0.535  & 1.097  & 0.835  & 0.491  & 0.469  & 0.569  & 0.551  & 0.481  & 0.459  & 0.480  &  \textcolor{softblue}{\uline{0.445}}  \\ 
        ~ & 720 & \textbf{\textcolor{red}{0.445}}  & \textbf{\textcolor{red}{0.448}}  & 1.181  & 0.865  & 0.643  & 0.616  & 1.257  & 0.889  & 0.521  & 0.500  & 0.770  & 0.672  & 0.519  & 0.516  &  \textcolor{softblue}{\uline{0.481}}  & \textcolor{softblue}{\uline{0.469}}   \\ 
        \cmidrule(lr){2-18} 
        ~ & Avg. & \textbf{\textcolor{red}{0.416}} & \textbf{\textcolor{red}{0.425}}  & 1.040  & 0.795  & 0.570  & 0.537 & 1.029  & 0.805  & 0.458  & 0.450  & 0.558  & 0.535  & 0.456  & 0.452  & 0.445  & \textcolor{softblue}{\uline{0.432}}  \\ 
        \midrule
        \multirow{5}{*}{\rotatebox{90}{ETTh2}} & 96 & \textbf{\textcolor{red}{0.292}}   & \textbf{\textcolor{red}{0.336}}  & 3.755  & 1.525  & 0.476  & 0.458  & 2.626  & 1.317  & 0.340  & 0.374  & 0.299  & 0.364  & 0.333  & 0.387  & 0.400  & 0.440  \\
        ~ & 192 & \textbf{\textcolor{red}{0.362}}  & \textbf{\textcolor{red}{0.383}}  & 5.602  & 1.931  & 0.512  & 0.493  & 11.120  & 2.979  & 0.402  & 0.414  & 0.441  & 0.454  & 0.477  & 0.476  & 0.528  & 0.509  \\
        ~ & 336 &  \textbf{\textcolor{red}{0.388}}  &  \textbf{\textcolor{red}{0.414}} & 4.721  & 1.835  & 0.552  & 0.551  & 9.323  & 2.769  & 0.452  & 0.452  & 0.654  & 0.567  & 0.594  & 0.541  & 0.643  & 0.571  \\
        ~ & 720 & \textbf{\textcolor{red}{0.414}}  & \textbf{\textcolor{red}{0.433}} & 3.647  & 1.625  & 0.562  & 0.560  & 3.874  & 1.697  & 0.462  & 0.468  & 0.956  & 0.716  & 0.831  & 0.657  & 0.874  & 0.679  \\ 
        \cmidrule(lr){2-18} 
        ~ & Avg. & \textbf{\textcolor{red}{0.364}} & \textbf{\textcolor{red}{0.391}}  & 4.431  & 1.729  & 0.526  & 0.516  & 6.736  & 2.191  & 0.414  & 0.427  & 0.587  & 0.525  & 0.559  & 0.551  & 0.611  & 0.550  \\
        \midrule
        \multirow{5}{*}{\rotatebox{90}{Weather}}  & 96 &  \textcolor{softblue}{\uline{0.161}}   & 0.198 & 0.300  & 0.384  & 0.173  & 0.223  & 0.689  & 0.596  & 0.172  & 0.220  & 0.161  & 0.229  & 0.196  & 0.255  & 0.202  & 0.261  \\
        ~ & 192 & \textcolor{softblue}{\uline{0.207}}  & \textbf{\textcolor{red}{0.241}}   & 0.598  & 0.544  & 0.245  & 0.285  & 0.752  & 0.638  & 0.219  & 0.261  & 0.220  & 0.281  & 0.237  & 0.296  & 0.242  & 0.298  \\
        ~ & 336 &  \textbf{\textcolor{red}{0.263}}  & \textbf{\textcolor{red}{0.285}}    & 0.578  & 0.523  & 0.321  & 0.338  & 0.639  & 0.596  & 0.280  & 0.306  & 0.278  & 0.331  & 0.283  & 0.335  & 0.287  & 0.335 \\ 
        ~ & 720 & \textcolor{softblue}{\uline{0.341}} & \textbf{\textcolor{red}{0.335}}  & 1.059  & 0.741  & 0.414  & 0.410  & 1.130  & 0.792  & 0.365  & 0.359  & \textbf{\textcolor{red}{0.311}} & 0.356  & 0.345  & 0.381  & 0.351  & 0.386  \\ 
        \cmidrule(lr){2-18}
        ~ & Avg. & \textcolor{softblue}{\uline{0.243}}   & \textbf{\textcolor{red}{0.265}} & 0.634  & 0.548  & 0.288  & 0.314  & 0.803  & 0.656  & 0.259  & 0.287  & \textbf{\textcolor{red}{0.242}}   & 0.299  & 0.265  & 0.317  & 0.271  & 0.320  \\ 
        \midrule
        \multirow{5}{*}{\rotatebox{90}{Exchange}} & 96 &  \textbf{\textcolor{red}{0.084}}  &  \textbf{\textcolor{red}{0.204}}   & 0.847  & 0.752  & 0.111  & 0.237  & 1.065  & 0.829  & 0.107  & 0.234  & 0.102  & 0.235  & 0.088  & 0.218  & \textbf{\textcolor{red}{0.084*}}  & 0.208*  \\ 
        ~ & 192 &  \textbf{\textcolor{red}{0.172}}   &  \textbf{\textcolor{red}{0.298}}   & 1.204  & 0.895  & 0.219  & 0.335  & 1.188  & 0.906  & 0.226  & 0.344  &  \textbf{\textcolor{red}{0.172}}  & 0.316  & \textcolor{softblue}{\uline{0.176}}  & 0.315  & 0.179* & 0.313* \\ 
        ~ & 336 & 0.315  &  \textbf{\textcolor{red}{0.407}}  & 1.672  & 1.036  & 0.421  & 0.476  & 1.357  & 0.976  & 0.367  & 0.448  & \textbf{\textcolor{red}{0.272}}  & \textbf{\textcolor{red}{0.407}} & 0.313  & 0.427  & \textcolor{softblue}{\uline{0.300*}} & \textcolor{softblue}{\uline{0.411*}}  \\
        ~ & 720 & \textcolor{softblue}{\uline{0.785}}   & \textcolor{softblue}{\uline{0.666}}  & 2.478  & 1.310  & 1.092  & 0.769  & 1.510  & 1.016  & 0.964  & 0.746  &  \textbf{\textcolor{red}{0.714}}   &  \textbf{\textcolor{red}{0.658}}  & 0.839  & 0.695  & 0.793* & 0.671* \\  
        \cmidrule(lr){2-18} 
        ~ & Avg. & \textcolor{softblue}{\uline{0.339}}   & \textbf{\textcolor{red}{0.394}}  & 1.550  & 0.998  & 0.461  & 0.454  & 1.280  & 0.932  & 0.416  & 0.443  & \textbf{\textcolor{red}{0.315}} & 0.404  & 0.354  & 0.414  &  \textcolor{softblue}{\uline{0.339*}} & \textcolor{softblue}{\uline{0.400*}} \\  
        \midrule
        \multirow{5}{*}{\rotatebox{90}{Illness}} & 24 &  \textbf{\textcolor{red}{1.844}}   & \textbf{\textcolor{red}{0.813}}    & 5.764  & 1.677  &  \textcolor{softblue}{\uline{2.294}}  & 0.945  & 4.400  & 1.382  & 2.317  & 0.934  & 2.684  & 1.112  & 2.398  & 1.040  & 4.504* & 1.584* \\
        ~ & 36 & 1.979 &  \textbf{\textcolor{red}{0.847}}   & 4.755  & 1.467  & \textbf{\textcolor{red}{1.825}} & \textcolor{softblue}{\uline{0.848}}  & 4.783  & 1.448  &  \textcolor{softblue}{\uline{1.972}} & 0.920  & 2.667  & 1.608  & 2.646  & 1.088  & 3.859* & 1.423* \\ 
        ~ & 48 &  \textbf{\textcolor{red}{1.769}} & \textbf{\textcolor{red}{0.806}}   & 4.763  & 1.469  & 2.010  & 0.900  & 4.832  & 1.465  & 2.238  & 0.940  & 2.558  & 1.052  & 2.614  & 1.086  & 3.799* & 1.406* \\
        ~ & 60 & \textbf{\textcolor{red}{1.683}}  &\textbf{\textcolor{red}{0.790}}   & 5.264  & 1.564  & 2.178  & 0.963  & 4.882  & 1.483  & 2.027  & 0.928  & 2.747  & 1.110  & 2.804  & 1.146  & 3.917* & 1.403* \\ 
        \cmidrule(lr){2-18} 
        ~ & Avg. &  \textbf{\textcolor{red}{1.819}}  & \textbf{\textcolor{red}{0.814}}   & 5.137  & 1.544  & 2.077  & 0.914  & 4.724  & 1.445  & 2.139  & 0.931  & 2.664  & 1.221  & 2.616  & 1.090  & 4.020* & 1.454* \\  
        \midrule
        \multicolumn{2}{c}{1\textsuperscript{st}Count} & \multicolumn{2}{c}{\textbf{\textcolor{red}{60}}} & \multicolumn{2}{c}{0}  & \multicolumn{2}{c}{1}  &  \multicolumn{2}{c}{0}  &  \multicolumn{2}{c}{0}  &  \multicolumn{2}{c}{\textcolor{softblue}{\uline{8}}}  &  \multicolumn{2}{c}{0}  &  \multicolumn{2}{c}{1} \\ 
        \bottomrule
      \end{tabular}}
\end{table*}

\section{Full Short-term Forecasting Results}
\label{app:app_full_short_term_results}

A detailed comparison of our proposed STELLA model with a broad range of baseline methods for short-term forecasting is provided in Table~\ref{tab:shortterm_full}. The compared baselines include GPT4TS \cite{zhou2023one}, TimeLLM \cite{jin2023time}, PatchTST \cite{nie2022time}, ETSformer \cite{woo2022etsformer}, FEDformer \cite{zhou2022fedformer}, Autoformer \cite{wu2021autoformer}, TimesNet \cite{wu2022timesnet}, TCN \cite{bai2018empirical}, N-HiTS \cite{challu2023nhits}, N-BEATS \cite{oreshkin2019n}, DLinear \cite{zeng2023transformers}, LSSL \cite{gu2021efficiently}, and LSTM \cite{hochreiter1997long}.

\begin{table*}[htbp]
    \centering
    \caption{Full Results for Short-Term Forecasting on the M4 Dataset. The input and prediction lengths are set to [12, 96] and [6, 48], respectively. \textbf{\textcolor{red}{best}} and \textcolor{softblue}{\uline{second-best}} results are highlighted accordingly. \textbf{1st Count} indicates the number of times each method achieves the \textbf{\textcolor{red}{best}} performance.}
     \label{tab:shortterm_full}
    \footnotesize
    \setlength{\heavyrulewidth}{1.1pt}
    \setlength{\lightrulewidth}{0.5pt}
    \setlength{\cmidrulewidth}{0.5pt}
    \resizebox{\textwidth}{!}{
    \renewcommand{\arraystretch}{1.3}
    \begin{tabular}{@{}p{0.2cm}|c|ccccccccccccccc@{}}
        \toprule
        \multicolumn{2}{c}{Categories} & \multicolumn{3}{c}{LLM-Based}  &  \multicolumn{4}{c}{Transformer-based}  &   \multicolumn{2}{c}{CNN-based} & \multicolumn{3}{c}{MLP-based}   & \multicolumn{2}{c}{Other}  \\ 
        \toprule
        \multicolumn{2}{c}{Models} & \makecell{\textbf{STELLA}\\\textbf{(Ours)}}  &  \makecell{TimeLLM\\ \cite{jin2023time}} & \makecell{GPT4TS\\ \cite{zhou2023one}} & \makecell{PatchTST\\ \cite{nie2022time}} &  \makecell{ETSformer \\ \cite{woo2022etsformer}} & \makecell{FEDformer\\ \cite{zhou2022fedformer}} & \makecell{Autoformer\\ \cite{wu2021autoformer}} & \makecell{TimesNet  \\ \cite{wu2022timesnet}}& \makecell{TCN \\ \cite{bai2018empirical}} & \makecell{N-HiTS\\ \cite{challu2023nhits}}  &  \makecell{N-BEATS \\ \cite{oreshkin2019n}} & \makecell{DLinear\\ \cite{zeng2023transformers}}  & \makecell{LSSL \\ \cite{gu2021efficiently}} & \makecell{LSTM\\ \cite{hochreiter1997long}} \\
        \toprule
        \multirow{3}{*}{\rotatebox{90}{Yearly}}
          & SMAPE & \textbf{\textcolor{red}{13.348}}  & 13.419  & 13.531  & 13.477  & 18.009  & 13.728  & 13.974  &  \textcolor{softblue}{\uline{13.387}}   & 14.920  & 13.418  & 13.436  & 16.965  & 61.675  & 176.040  \\
          & MASE & \textbf{\textcolor{red}{2.978}}  & 3.005  & 3.015  & 3.019  & 4.487  & 3.048  & 3.134  &  \textcolor{softblue}{\uline{2.996}}  & 3.364  & 3.045  & 3.043  & 4.283  & 19.953  & 31.033  \\ 
          & OWA & \textbf{\textcolor{red}{0.783 }} & 0.789  & 0.793  & 0.792  & 1.115  & 0.803  & 0.822  & \textcolor{softblue}{\uline{0.786}}  & 0.880  & 0.793  & 0.794  & 1.058  & 4.397  & 9.290  \\ 
        \midrule
        \multirow{3}{*}{\rotatebox{90}{Quarterly}}
          & SMAPE & \textbf{\textcolor{red}{9.997 }}  & 10.110  & 10.177  & 10.380  & 13.376  & 10.792  & 11.338  & \textcolor{softblue}{\uline{10.100}}   & 11.122  & 10.202  & 10.124  & 12.145  & 65.999  & 172.808  \\ 
          & MASE & \textbf{\textcolor{red}{1.164}}  & 1.178  & 1.194  & 1.233  & 1.906  & 1.283  & 1.365  & 1.182  & 1.360  & 1.194  &  \textcolor{softblue}{\uline{1.169}}  & 1.520  & 17.662  & 19.753  \\ 
          & OWA & \textbf{\textcolor{red}{0.878}} & 0.889  & 0.898  & 0.921  & 1.302  & 0.958  & 1.012  & 0.890  & 1.001  & 0.899  & \textcolor{softblue}{\uline{0.886}}  & 1.106  & 9.436  & 15.049  \\   
        \midrule
        \multirow{3}{*}{\rotatebox{90}{Monthly}}
          & SMAPE & \textbf{\textcolor{red}{12.608}}  & 12.980  & 12.894  & 12.959  & 14.588  & 14.260  & 13.958  & 12.679  & 15.626  & 12.791  & \textcolor{softblue}{\uline{12.677}} & 13.514  & 64.664  & 143.237  \\ 
          & MASE & \textbf{\textcolor{red}{0.932}}  & 0.963  & 0.956  & 0.970  & 1.368  & 1.102  & 1.103  & \textcolor{softblue}{\uline{0.933}}  & 1.274  & 0.969  & 0.937  & 1.037  & 16.245  & 16.551  \\ 
          & OWA & \textbf{\textcolor{red}{0.875}}  & 0.903  & 0.897  & 0.905  & 1.149  & 1.012  & 1.002  &  \textcolor{softblue}{\uline{0.878}}  & 1.141  & 0.899  & 0.880  & 0.956  & 9.879  & 12.747  \\
        \midrule
        \multirow{3}{*}{\rotatebox{90}{Others}}
          & SMAPE &  \textbf{\textcolor{red}{4.651}}   & \textcolor{softblue}{\uline{4.795}}   & 4.940  & 4.952  & 7.267  & 4.954  & 5.485  & 4.891  & 7.186  & 5.061  & 4.925  & 6.709  & 121.844  & 186.282  \\ 
          & MASE & \textbf{\textcolor{red}{3.112}} & \textcolor{softblue}{\uline{3.178}}    & 3.228  & 3.347  & 5.240  & 3.264  & 3.865  & 3.302  & 4.677  & 3.216  & 3.391  & 4.953  & 91.650  & 119.294  \\ 
          & OWA &  \textbf{\textcolor{red}{0.980}}  & \textcolor{softblue}{\uline{1.006}}  & 1.029  & 1.049  & 1.591  & 1.036  & 1.187  & 1.035  & 1.494  & 1.040  & 1.053  & 1.487  & 27.273  & 38.411  \\  
        \midrule
        \multirow{3}{*}{\rotatebox{90}{Average}}
            & SMAPE & \textbf{\textcolor{red}{11.754}}   & 11.983  & 11.991  & 12.059  & 14.718  & 12.840  & 12.909  & \textcolor{softblue}{\uline{11.829}}  & 13.961  & 11.927  & 11.851  & 13.639  & 67.156  & 160.031  \\ 
            & MASE & \textbf{\textcolor{red}{1.567}}    & 1.595  & 1.600  & 1.623  & 2.408  & 1.701  & 1.771  & \textcolor{softblue}{\uline{1.585}}  & 1.945  & 1.613  & 1.599  & 2.095  & 21.208  & 25.788  \\ 
            & OWA & \textbf{\textcolor{red}{0.843}} & 0.859  & 0.861  & 0.869  & 1.172  & 0.918  & 0.939  & \textcolor{softblue}{\uline{0.851}} & 1.023  & 0.861  & 0.855  & 1.051  & 8.021  & 12.642  \\ 
        \midrule
        \multicolumn{2}{l}{1\textsuperscript{st}Count} &  \textbf{\textcolor{red}{15}}  & 0 & 0 & 0 & 0 & 0 & 0 & 0 & 0 & 0 & 0 & 0 & 0 & 0 \\
        \bottomrule
    \end{tabular}
    }
\end{table*}

\section{Full Few-shot Forecasting Results}
\label{app:app_fewshot_results}
Following the experimental protocol outlined in the main text, we assess the performance of our model in the few-shot forecasting setting, where only 10\% of the training data is utilized for each dataset. This scenario poses a substantial challenge, aiming to evaluate the model's capacity to generalize under limited supervision. The complete results across all prediction lengths \( H \in \{96, 192, 336, 720\} \) are presented in Table~\ref{tab:full_few_shot}.

\begin{table*}[htbp]
    \centering    
    \caption{Full results for few-shot learning on 10\% of the training data from the ETT datasets with varying prediction lengths $H \in \{96, 192, 336, 720\}$. The input sequence length is fixed at 96 across all baselines. \textbf{Avg.} represents the average performance over all four prediction horizons. The \textbf{\textcolor{red}{best}} and \textcolor{softblue}{\uline{second-best}} results for each forecasting horizon are highlighted in \textbf{\textcolor{red}{bold}} and \textcolor{softblue}{\uline{underlined}}, respectively. The \textbf{1st Count} column reports the number of times each method achieves the \textbf{\textcolor{red}{best}} performance.}
    \label{tab:full_few_shot}
    \setlength{\heavyrulewidth}{1.1pt}
    \setlength{\lightrulewidth}{0.5pt}
    \setlength{\cmidrulewidth}{0.5pt}
    \resizebox{\textwidth}{!}{
    \begin{tabular}{cc|cc|cc|cc|cc|cc|cc|cc|cc|cc|cc}
    \toprule
        \multicolumn{2}{c}{Categories}  & \multicolumn{6}{c}{LLM-Based}  & \multicolumn{6}{c}{Transformer-based}  & \multicolumn{4}{c}{CNN-based} & \multicolumn{4}{c}{MLP-based}  \\
    \toprule
        \multicolumn{2}{c}{Models} 
        & \multicolumn{2}{c}{\makecell{\textbf{STELLA}\\\textbf{(Ours)}}}   
        & \multicolumn{2}{c}{\makecell{TimeLLM\\ \cite{jin2023time}}} 
        & \multicolumn{2}{c}{\makecell{GPT4TS\\ \cite{zhou2023one}}} 
        & \multicolumn{2}{c}{\makecell{PatchTST\\ \cite{nie2022time}}} 
        & \multicolumn{2}{c}{\makecell {Crossformer\\ \cite{zhang2023crossformer}}}  
        & \multicolumn{2}{c}{\makecell{FEDformer\\ \cite{zhou2022fedformer}} } 
        & \multicolumn{2}{c}{\makecell {TimesNet \\ \cite{wu2022timesnet}}}
        & \multicolumn{2}{c}{\makecell{MICN \\ \cite{wang2023micn}}}
        & \multicolumn{2}{c}{\makecell{DLinear\\ \cite{zeng2023transformers}}} 
        & \multicolumn{2}{c}{\makecell{TiDE\\ \cite{das2023long}}} \\
    \toprule
        \multicolumn{2}{c}{Metric} 
        & MSE & MAE 
        & MSE & MAE 
        & MSE & MAE 
        & MSE & MAE 
        & MSE & MAE 
        & MSE & MAE 
        & MSE & MAE 
        & MSE & MAE 
        & MSE & MAE 
        & MSE & MAE \\
    \toprule
        \multirow{5}{*}{\rotatebox{90}{ETTm1}}  & 96 & \textcolor{softblue}{\uline{0.521}} & \textbf{\textcolor{red}{0.449}}  & 0.587 & 0.491 & 0.615 & 0.497 & 0.558 & 0.478 & 1.037 & 0.705 & 0.604 & 0.530 & 0.583 & 0.503 & 0.677 & 0.585 & 0.552 & 0.488 & \textbf{\textcolor{red}{0.501}} & \textcolor{softblue}{\uline{0.458}} \\ 
        ~ & 192 & \textcolor{softblue}{\uline{0.506}} &   \textbf{\textcolor{red}{0.454}} & 0.606 & 0.490 & 0.597 & 0.492 & 0.539 & 0.471 & 1.170 & 0.778 & 0.641 & 0.546 & 0.608 & 0.515 & 0.784 & 0.627 & 0.546 & 0.487 &  \textbf{\textcolor{red}{0.493}} & \textcolor{softblue}{\uline{0.456}} \\ 
        ~ & 336 & \textcolor{softblue}{\uline{0.537}}  & \textbf{\textcolor{red}{0.477}} & 0.719 & 0.555 & 0.597 & 0.501 & 0.558 & 0.488 & 1.463 & 0.913 & 0.768 & 0.606 & 0.733 & 0.572 & 0.972 & 0.684 & 0.567 & 0.501 & \textbf{\textcolor{red}{0.516}}& \textbf{\textcolor{red}{0.477}} \\ 
        ~ & 720 & 0.609  & 0.513  & 0.632 & 0.514 & 0.623 & 0.513 & \textcolor{softblue}{\uline{0.574}}  & \textcolor{softblue}{\uline{0.498}}  & 1.693 & 0.997 & 0.771 & 0.606 & 0.768 & 0.548 & 1.449 & 0.800 & 0.606 & 0.522 & \textbf{\textcolor{red}{0.553}} & \textbf{\textcolor{red}{0.488}} \\ 
        \cmidrule(lr){2-22}
        ~ & Avg. &  \textcolor{softblue}{\uline{0.543}}  & \textcolor{softblue}{\uline{0.473}}   & 0.636 & 0.512 & 0.608 & 0.500 & 0.557 & 0.483 & 1.340 & 0.848 & 0.696 & 0.572 & 0.673 & 0.534 & 0.970 & 0.674 & 0.567 & 0.499 & \textbf{\textcolor{red}{0.515}} & \textbf{\textcolor{red}{0.469}}  \\ 
        \midrule
        \multirow{5}{*}{\rotatebox{90}{ETTm2}}  & 96 &   \textbf{\textcolor{red}{0.186}}  &  \textbf{\textcolor{red}{0.266}} & 0.189 & 0.270  & \textcolor{softblue}{\uline{0.187}}  & \textbf{\textcolor{red}{0.266}} & 0.189 & \textcolor{softblue}{\uline{0.268}}  & 1.397 & 0.866 & 0.222 & 0.314 & 0.214 & 0.288 & 0.389 & 0.448 & 0.225 & 0.320  & 0.191 & 0.269 \\ 
        ~ & 192 & \textbf{\textcolor{red}{0.246}}    & \textbf{\textcolor{red}{0.306}}   & 0.264 & 0.319 & 0.253 & 0.308 & \textcolor{softblue}{\uline{0.248}} & \textcolor{softblue}{\uline{0.307}} & 1.757 & 0.987 & 0.284 & 0.351 & 0.271 & 0.325 & 0.622 & 0.575 & 0.291 & 0.362 & 0.256 & 0.310  \\ 
        ~ & 336 & \textbf{\textcolor{red}{0.306}}    & \textbf{\textcolor{red}{0.344}}   & 0.327 & 0.358 & 0.332 & 0.353 & \textcolor{softblue}{\uline{0.311}} & \textcolor{softblue}{\uline{0.346}} & 2.075 & 1.086 & 0.392 & 0.419 & 0.329 & 0.356 & 1.055 & 0.755 & 0.354 & 0.402 & 0.321 & 0.349 \\ 
        ~ & 720 & \textbf{\textcolor{red}{0.428}}  &  \textbf{\textcolor{red}{0.412}} & 0.454 & 0.428 & 0.438 & 0.417 & \textcolor{softblue}{\uline{0.435}}  & \textcolor{softblue}{\uline{0.418}}  & 2.712 & 1.253 & 0.527 & 0.485 & 0.473 & 0.448 & 2.226 & 1.087 & 0.446 & 0.447 & 0.446 & 0.421 \\ 
        \cmidrule(lr){2-22}
        ~ & Avg. & \textbf{\textcolor{red}{0.291}}  & \textbf{\textcolor{red}{0.332}}  & 0.308 & 0.343 & 0.303 & 0.336 & \textcolor{softblue}{\uline{0.295}}  & \textcolor{softblue}{\uline{0.334}} & 1.985 & 1.048 & 0.356 & 0.392 & 0.321 & 0.354 & 1.073 & 0.716 & 0.329 & 0.382 & 0.303 & 0.337 \\ 
        \midrule
        \multirow{5}{*}{\rotatebox{90}{ETTh1}}  & 96 & 0.512  & 0.472  & 0.500 & 0.464 & \textcolor{softblue}{\uline{0.462}} &  \textcolor{softblue}{\uline{0.449}}  & \textbf{\textcolor{red}{0.433}} &  \textbf{\textcolor{red}{0.428}} & 1.129 & 0.775 & 0.651 & 0.563 & 0.855 & 0.625 & 0.689 & 0.592 & 0.590  & 0.515 & 0.642 & 0.545 \\ 
        ~ & 192 & 0.569  &  \textcolor{softblue}{\uline{0.492}}   & 0.590 & 0.516 & \textcolor{softblue}{\uline{0.551}}   & 0.495 & \textbf{\textcolor{red}{0.509}}  & \textbf{\textcolor{red}{0.474}}  & 1.832 & 0.922 & 0.666 & 0.562 & 0.791 & 0.589 & 1.160 & 0.748 & 0.634 & 0.541 & 0.761 & 0.595 \\ 
        ~ & 336 & 0.632  &  \textbf{\textcolor{red}{0.508}}  & 0.638 & 0.542 & \textcolor{softblue}{\uline{0.630}}  & 0.539 &  \textbf{\textcolor{red}{0.572}}  & \textcolor{softblue}{\uline{0.509}}   & 2.022 & 0.973 & 0.767 & 0.602 & 0.939 & 0.648 & 1.747 & 0.899 & 0.659 & 0.554 & 0.789 & 0.610  \\ 
        ~ & 720 & \textcolor{softblue}{\uline{0.755}}   & \textcolor{softblue}{\uline{0.617}}   & 1.344 & 0.816 & 1.113 & 0.738 & 1.221 & 0.773 & 1.903 & 0.986 & 0.918 & 0.703 & 0.876 & 0.641 & 2.024 & 1.019 &  \textbf{\textcolor{red}{0.708}}   & \textbf{\textcolor{red}{0.598}}   & 0.927 & 0.667 \\ 
        \cmidrule(lr){2-22}
        ~ & Avg. & \textbf{\textcolor{red}{0.617}}   &  \textbf{\textcolor{red}{0.522}}  & 0.765 & 0.584 & 0.689 & 0.555 & 0.683 & 0.645 & 1.744 & 0.914 & 0.750  & 0.607 & 0.865 & 0.625 & 1.405 & 0.814 &  \textcolor{softblue}{\uline{0.647}}   & \textcolor{softblue}{\uline{0.552}} & 0.779 & 0.604 \\
        \midrule
        \multirow{5}{*}{\rotatebox{90}{ETTh2}}  & 96 &  \textbf{\textcolor{red}{0.303}}  &  \textbf{\textcolor{red}{0.350}}    & 0.329 & 0.365 & \textcolor{softblue}{\uline{0.327}}  & 0.359 & 0.314 &  \textcolor{softblue}{\uline{0.354}} & 2.482 & 1.206 & 0.359 & 0.404 & 0.372 & 0.405 & 0.510 & 0.502 & 0.361 & 0.407 & 0.337 & 0.379 \\ 
        ~ & 192 & \textbf{\textcolor{red}{0.377}}   & \textbf{\textcolor{red}{0.393}}   & 0.414 & 0.413 &   \textcolor{softblue}{\uline{0.403}}   &   \textcolor{softblue}{\uline{0.405}}   & 0.420  & 0.415 & 3.136 & 1.372 & 0.460  & 0.461 & 0.483 & 0.463 & 1.809 & 1.036 & 0.444 & 0.453 & 0.424 & 0.427 \\
        ~ & 336 & \textbf{\textcolor{red}{0.434}}  & \textcolor{softblue}{\uline{0.435}}   & 0.579 & 0.506 & 0.568 & 0.499 & 0.543 & 0.489 & 2.925 & 1.331 & 0.569 & 0.530  & 0.541 & 0.496 & 3.250 & 1.419 & 0.509 & 0.501 &\textcolor{softblue}{\uline{0.435}}   & \textbf{\textcolor{red}{0.426}} \\ 
        ~ & 720 & 0.541  & 0.508  & 1.034 & 0.711 & 1.020  & 0.725 & 0.926 & 0.691 & 4.014 & 1.603 & 0.827 & 0.707 & 0.510  & 0.491 & 4.564 & 1.676 &  \textbf{\textcolor{red}{0.453}} & \textbf{\textcolor{red}{0.471}}  & \textcolor{softblue}{\uline{0.489}} &  \textcolor{softblue}{\uline{0.480}}  \\ 
        \cmidrule(lr){2-22}
        ~ & Avg. & \textbf{\textcolor{red}{0.414}}  &  \textbf{\textcolor{red}{0.421}}  & 0.589 & 0.498 & 0.579 & 0.497 & 0.550  & 0.487 & 3.139 & 1.378 & 0.553 & 0.525 & 0.476 & 0.463 & 2.533 & 1.158 & 0.441 & 0.458 &  \textcolor{softblue}{\uline{0.421}}  &   \textcolor{softblue}{\uline{0.428}}  \\ 
        \midrule
        \multicolumn{2}{l}{1\textsuperscript{st}Count} & \multicolumn{2}{c}{\textcolor{red}{23}} & \multicolumn{2}{c}{0} & \multicolumn{2}{c}{1} & \multicolumn{2}{c}{5} & \multicolumn{2}{c}{0} & \multicolumn{2}{c}{0} & \multicolumn{2}{c}{0} & \multicolumn{2}{c}{0} & \multicolumn{2}{c}{4} & \multicolumn{2}{c}{\textcolor{softblue}{\uline{9}}} \\
    \bottomrule
     \end{tabular}}
\end{table*}

\section{Full Zero-shot Forecasting Results}
\label{app:app_zeroshot_results}

To further evaluate the cross-domain generalization ability of our model, we present the complete results of the zero-shot forecasting experiments in Table~\ref{tab:zero_shot}. In this setting, the model is trained on one ETT dataset and directly tested on a different target dataset, without access to its training samples. We use the notation “A~$\rightarrow$~B” to denote transfer from source dataset A to target dataset B (e.g., ETTh1~$\rightarrow$~ETTm2), following the protocol described in the main text. Results are reported across all prediction lengths \( H \in \{96, 192, 336, 720\} \).

\begin{table*}[htbp]
    \centering
    \caption{Full results for zero-shot forecasting on the ETT datasets. Each row represents a cross-dataset evaluation setting, where the model is trained on one ETT dataset and evaluated on another (e.g., ETTh1~$\rightarrow$~ETTm2). The input sequence length is fixed at 96 for all baselines. All results are averaged over four prediction horizons, $H \in \{96, 192, 336, 720\}$. \textbf{Avg.} represents the average performance over all four prediction horizons. The \textbf{\textcolor{red}{best}} and \textcolor{softblue}{\uline{second-best}} results are highlighted accordingly. \textbf{1st Count} indicates the number of times each method achieves the \textbf{\textcolor{red}{best}} performance.}
    \label{tab:zero_shot}
    \setlength{\heavyrulewidth}{1.1pt}
    \setlength{\lightrulewidth}{0.5pt}
    \setlength{\cmidrulewidth}{0.5pt}
    \resizebox{\textwidth}{!}{
    \begin{tabular}{cc|cc|cc|cc|cc|cc|cc|cc|cc|cc|cc}
    \toprule
        \multicolumn{2}{c}{Categories}  & \multicolumn{6}{c}{LLM-Based}  & \multicolumn{6}{c}{Transformer-based}  & \multicolumn{4}{c}{CNN-based} & \multicolumn{4}{c}{MLP-based}  \\
    \toprule
        \multicolumn{2}{c}{Models} 
        & \multicolumn{2}{c}{\makecell{\textbf{STELLA}\\\textbf{(Ours)}}}   
        & \multicolumn{2}{c}{\makecell{TimeLLM\\ \cite{jin2023time}}} 
        & \multicolumn{2}{c}{\makecell{GPT4TS\\ \cite{zhou2023one}}} 
        & \multicolumn{2}{c}{\makecell{PatchTST\\ \cite{nie2022time}}} 
        & \multicolumn{2}{c}{\makecell {Crossformer\\ \cite{zhang2023crossformer}}}  
        & \multicolumn{2}{c}{\makecell{FEDformer\\ \cite{zhou2022fedformer}} } 
        & \multicolumn{2}{c}{\makecell {TimesNet \\ \cite{wu2022timesnet}}}
        & \multicolumn{2}{c}{\makecell{MICN \\ \cite{wang2023micn}}}
        & \multicolumn{2}{c}{\makecell{DLinear\\ \cite{zeng2023transformers}}} 
        & \multicolumn{2}{c}{\makecell{TiDE\\ \cite{das2023long}}} \\
    \toprule
        \multicolumn{2}{c}{Metric} 
        & MSE & MAE 
        & MSE & MAE 
        & MSE & MAE 
        & MSE & MAE 
        & MSE & MAE 
        & MSE & MAE 
        & MSE & MAE 
        & MSE & MAE 
        & MSE & MAE 
        & MSE & MAE \\
    \toprule
        \multirow{5}{*}{\rotatebox{90}{h1->m1}}  
        & 96 & \textbf{\textcolor{red}{0.697}} & \textbf{\textcolor{red}{0.545}} & 0.804 & 0.565 & 0.809 & 0.563 & 0.908 & 0.596 & 0.856 & 0.649 & \textcolor{softblue}{\uline{0.731}} & 0.561 & 0.764 & 0.563 & 0.832 & 0.621 & 0.735 & 0.554 & 0.748 & \textcolor{softblue}{\uline{0.551}} \\ 
        ~ & 192 & \textbf{\textcolor{red}{0.711}} & \textbf{\textcolor{red}{0.554}} & 0.827 & 0.593 & 0.799 & 0.567 & 0.927 & 0.616 & 0.906 & 0.684 & \textcolor{softblue}{\uline{0.746}} & 0.573 & 0.798 & \textcolor{softblue}{\uline{0.562}} & 1.288 & 0.854 & 0.752 & 0.570 & 0.779 & 0.571 \\ 
        ~ & 336 & \textbf{\textcolor{red}{0.723}} & \textbf{\textcolor{red}{0.564}} & 0.835 & 0.600 & 0.803 & \textcolor{softblue}{\uline{0.577}} & 0.920 & 0.621 & 1.104 & 0.796 & 0.775 & 0.596 & 0.790 & 0.584 & 1.721 & 0.972 & \textcolor{softblue}{\uline{0.749}} & 0.579 & 0.775 & 0.580 \\ 
        ~ & 720 & \textbf{\textcolor{red}{0.746}} & \textbf{\textcolor{red}{0.580}} & 0.922 & 0.644 & \textcolor{softblue}{\uline{0.783}} & \textcolor{softblue}{\uline{0.589}} & 0.822 & 0.608 & 1.131 & 0.816 & 0.808 & 0.625 & 0.827 & 0.594 & 1.915 & 1.036 & 0.805 & 0.606 & 0.795 & 0.595 \\ 
        \cmidrule(lr){2-22}
        ~ & Avg. & \textbf{\textcolor{red}{0.719}} & \textbf{\textcolor{red}{0.561}} & 0.847 & 0.600 & 0.798 & \textcolor{softblue}{\uline{0.574}} & 0.894 & 0.610 & 0.999 & 0.736 & 0.765 & 0.588 & 0.794 & 0.575 & 1.439 & 0.870 & \textcolor{softblue}{\uline{0.760}} & 0.577 & 0.774 & \textcolor{softblue}{\uline{0.574}} \\ 
        \midrule
        \multirow{5}{*}{\rotatebox{90}{h1->m2}}  
        & 96 & \textbf{\textcolor{red}{0.212}} & \textbf{\textcolor{red}{0.297}} & \textbf{\textcolor{red}{0.212}} & \textcolor{softblue}{\uline{0.298}} & 0.218 & 0.304 & 0.219 & 0.305 & 0.611 & 0.588 & 0.257 & 0.345 & 0.245 & 0.322 & 0.496 & 0.556 & 0.239 & 0.343 & \textcolor{softblue}{\uline{0.215}} & 0.299 \\ 
        ~ & 192 & \textbf{\textcolor{red}{0.275}} & \textbf{\textcolor{red}{0.332}} & \textcolor{softblue}{\uline{0.277}} & 0.338 & 0.279 & 0.338 & 0.280 & 0.341 & 0.789 & 0.685 & 0.318 & 0.380 & 0.293 & 0.346 & 1.798 & 1.137 & 0.320 & 0.397 & \textcolor{softblue}{\uline{0.277}} & \textcolor{softblue}{\uline{0.335}} \\ 
        ~ & 336 & \textbf{\textcolor{red}{0.333}} & \textbf{\textcolor{red}{0.366}} & \textcolor{softblue}{\uline{0.336}} & 0.371 & 0.342 & 0.376 & 0.341 & 0.376 & 1.469 & 0.927 & 0.375 & 0.417 & 0.361 & 0.382 & 2.929 & 1.472 & 0.409 & 0.453 & 0.337 & \textcolor{softblue}{\uline{0.370}} \\ 
        ~ & 720 & \textbf{\textcolor{red}{0.427}} & \textbf{\textcolor{red}{0.416}} & 0.435 & 0.424 & 0.431 & 0.419 & 0.432 & 0.426 & 1.612 & 0.957 & 0.480 & 0.472 & 0.460 & 0.432 & 4.489 & 1.782 & 0.629 & 0.565 & \textcolor{softblue}{\uline{0.429}} & \textcolor{softblue}{\uline{0.418}} \\ 
        \cmidrule(lr){2-22}
        ~ & Avg. & \textbf{\textcolor{red}{0.312}} & \textbf{\textcolor{red}{0.353}} & 0.315 & 0.357 & 0.317 & 0.359 & 0.318 & 0.362 & 1.120 & 0.789 & 0.357 & 0.403 & 0.339 & 0.370 & 2.428 & 1.236 & 0.399 & 0.439 & \textcolor{softblue}{\uline{0.314}} & \textcolor{softblue}{\uline{0.355}} \\ 
        \midrule
        \multirow{5}{*}{\rotatebox{90}{h2->m1}}  
        & 96 & \textbf{\textcolor{red}{0.715}} & \textbf{\textcolor{red}{0.546}} & 0.891 & 0.587 & 0.985 & 0.604 & 0.815 & \textcolor{softblue}{\uline{0.560}} & 1.032 & 0.620 & \textcolor{softblue}{\uline{0.734}} & 0.578 & 1.205 & 0.678 & 0.743 & 0.577 & 0.762 & 0.567 & 0.819 & 0.566 \\ 
        ~ & 192 & \textbf{\textcolor{red}{0.720}} & \textbf{\textcolor{red}{0.548}} & 0.850 & \textcolor{softblue}{\uline{0.583}} & 0.872 & 0.600 & 0.900 & 0.606 & 1.176 & 0.676 & \textcolor{softblue}{\uline{0.723}} & 0.594 & 1.159 & 0.670 & 0.750 & 0.588 & 0.785 & 0.588 & 0.845 & 0.586 \\ 
        ~ & 336 & \textbf{\textcolor{red}{0.729}} & \textbf{\textcolor{red}{0.571}} & 0.853 & 0.594 & 0.926 & 0.614 & 0.906 & 0.602 & 1.199 & 0.718 & \textcolor{softblue}{\uline{0.750}} & \textcolor{softblue}{\uline{0.590}} & 1.197 & 0.689 & 0.764 & 0.606 & 0.767 & 0.594 & 0.834 & 0.595 \\ 
        ~ & 720 & \textbf{\textcolor{red}{0.758}} & \textbf{\textcolor{red}{0.582}} & 0.879 & 0.616 & 0.899 & 0.624 & 0.866 & 0.619 & 1.373 & 0.832 & \textcolor{softblue}{\uline{0.760}} & \textcolor{softblue}{\uline{0.592}} & 1.583 & 0.784 & 0.801 & 0.634 & 0.800 & 0.627 & 0.867 & 0.616 \\ 
        \cmidrule(lr){2-22}
        ~ & Avg. & \textbf{\textcolor{red}{0.730}} & \textbf{\textcolor{red}{0.562}} & 0.868 & 0.595 & 0.920 & 0.610 & 0.871 & 0.596 & 1.195 & 0.711 & \textcolor{softblue}{\uline{0.741}} & \textcolor{softblue}{\uline{0.588}} & 1.286 & 0.705 & 0.764 & 0.601 & 0.778 & 0.594 & 0.841 & 0.590 \\ 
        \midrule
        \multirow{5}{*}{\rotatebox{90}{h2->m2}}  
        & 96 & \textbf{\textcolor{red}{0.216}} & \textbf{\textcolor{red}{0.300}} & 0.228 & \textcolor{softblue}{\uline{0.311}} & 0.235 & 0.316 & 0.288 & 0.345 & 0.821 & 0.634 & 0.261 & 0.347 & 0.244 & 0.324 & 0.327 & 0.414 & 0.264 & 0.366 & \textcolor{softblue}{\uline{0.226}} & 0.315 \\ 
        ~ & 192 & \textbf{\textcolor{red}{0.275}} & \textbf{\textcolor{red}{0.333}} & \textcolor{softblue}{\uline{0.283}} & \textcolor{softblue}{\uline{0.341}} & 0.287 & 0.346 & 0.344 & 0.375 & 1.732 & 1.018 & 0.313 & 0.370 & 0.331 & 0.374 & 0.450 & 0.485 & 0.394 & 0.452 & 0.289 & 0.348 \\ 
        ~ & 336 & \textbf{\textcolor{red}{0.335}} & \textbf{\textcolor{red}{0.368}} & 0.343 & 0.376 & 0.361 & 0.391 & 0.438 & 0.425 & 2.587 & 1.393 & 0.401 & 0.431 & 0.386 & 0.405 & 0.526 & 0.526 & 0.506 & 0.513 & \textcolor{softblue}{\uline{0.339}} & \textcolor{softblue}{\uline{0.372}} \\ 
        ~ & 720 & \textbf{\textcolor{red}{0.428}} & \textbf{\textcolor{red}{0.417}} & 0.437 & 0.424 & 0.444 & 0.433 & 0.611 & 0.588 & 3.034 & 1.452 & 0.487 & 0.472 & 0.485 & 0.458 & 0.806 & 0.652 & 0.822 & 0.655 & \textcolor{softblue}{\uline{0.433}} & \textcolor{softblue}{\uline{0.422}} \\ 
        \cmidrule(lr){2-22}
        ~ & Avg. & \textbf{\textcolor{red}{0.313}} & \textbf{\textcolor{red}{0.354}} & 0.322 & \textcolor{softblue}{\uline{0.363}} & 0.331 & 0.371 & 0.420 & 0.433 & 2.043 & 1.124 & 0.365 & 0.405 & 0.361 & 0.390 & 0.527 & 0.519 & 0.496 & 0.496 & \textcolor{softblue}{\uline{0.321}} & 0.364 \\ 
        \midrule
        \multicolumn{2}{l}{1\textsuperscript{st}Count} & \multicolumn{2}{c}{\textcolor{red}{40}} & \multicolumn{2}{c}{\textcolor{softblue}{\uline{1}}} & \multicolumn{2}{c}{0} & \multicolumn{2}{c}{0} & \multicolumn{2}{c}{0} & \multicolumn{2}{c}{0} & \multicolumn{2}{c}{0} & \multicolumn{2}{c}{0} & \multicolumn{2}{c}{0} & \multicolumn{2}{c}{0} \\
    \bottomrule
     \end{tabular}}
\end{table*}

\section{Parameter Sensitivity Analysis}
\label{sec:appendix_sensitivity}
To elucidate the influence of our hierarchical prompt design, we conduct a systematic sensitivity analysis on two pivotal hyperparameters: the length of the Fine-grained Behavioral Prompt ($G_{pl}$) and that of the Corpus-level Semantic Prior ($G_{pl}^{\mathrm{data}}$). The experiments are performed on the ETTh1 dataset, with forecasting performance evaluated using MSE and MAE, where lower values signify higher accuracy.

\begin{figure*}[htbp]
    \centering
    \includegraphics[width=\textwidth]{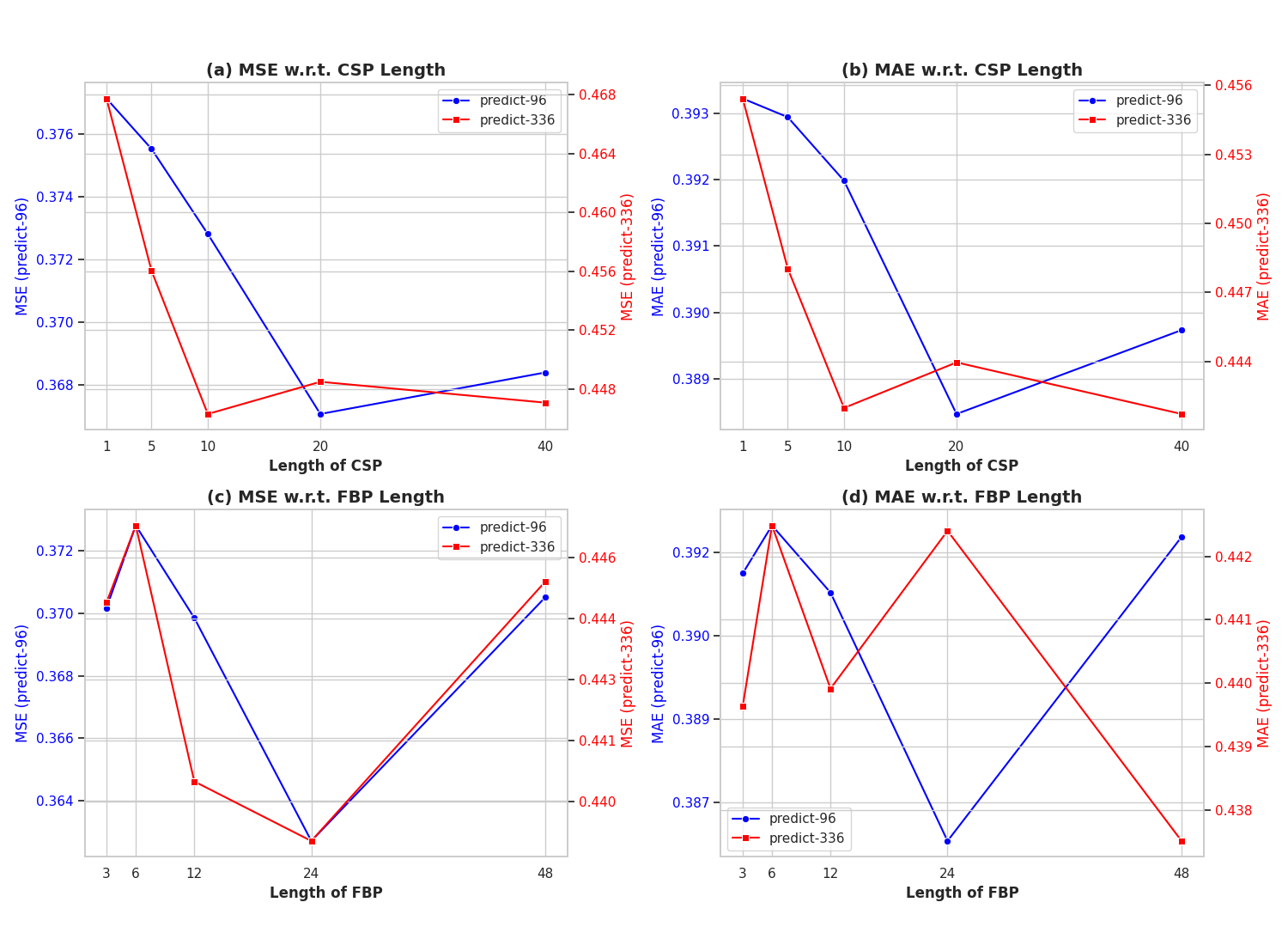}
    \caption{Parameter sensitivity analysis on the ETTh1 dataset for prediction horizons of 96 and 336: 
    (a) and (b) show the effect of the CSP length on MSE and MAE, respectively; 
    (c) and (d) show the effect of the FBP length on MSE and MAE, respectively.}
    \label{fig:parameter_sensitivity}
\end{figure*}

Our analysis first addresses the Fine-grained Behavioral Prompt (FBP), which is engineered to distill instance-specific behavioral signatures into a compact, learnable vector. Consequently, determining its optimal length, $G_{pl}$, is a critical design choice with significant performance implications. We establish a baseline length of $G_{pl}=6$, a value derived directly from the model's input structure (a 96-step sequence with a patch size of 16). To ensure a rigorous analysis, we selected the test range `{3, 6, 12, 24, 48}`. This choice is underpinned by a dual rationale: (i) it represents a systematic multiplicative scaling (from 0.5x to 8x) of the input-aligned baseline, and (ii) it strategically aligns with the number of patches corresponding to various forecast horizons (e.g., 12 patches for a 192-step forecast). This principled selection ensures our analysis spans a comprehensive spectrum of operational regimes, from highly information-constrained states to those with extensive expressive capacity. As illustrated in Figure~\ref{fig:parameter_sensitivity}(c-d), the model's performance exhibits a nuanced, non-monotonic relationship with the FBP length. Following a slight performance degradation from a highly compressed prompt ($G_{pl}=3$) to the baseline ($G_{pl}=6$), the forecasting error diminishes substantially as the length increases to 12 and 24. This trend strongly indicates that a more expressive FBP is instrumental for achieving high accuracy. However, extending the prompt further to $G_{pl}=48$ leads to a general performance decline, suggesting that an excessively long prompt introduces attentional noise or promotes overfitting to spurious textual details. Our analysis thus identifies an optimal operational range of 12-24 for the FBP, confirming the necessity of a carefully tuned balance between expressive power and architectural conciseness.

In contrast, the analysis for the Corpus-level Semantic Prior length, $G_{pl}^{\mathrm{data}}$, reveals a clearer, more consistent trend, as shown in Figure~\ref{fig:parameter_sensitivity}(a-b). Forecasting error steadily decreases as the CSP length increases from 1 to 10, demonstrating the value of a sufficiently rich, high-level contextual anchor. The improvements then begin to plateau between lengths of 10 and 20, signaling a point of diminishing returns. Beyond this threshold, increasing the length to 40 consistently, albeit moderately, degrades performance. This suggests that an overly verbose global prior can introduce semantic redundancy, which can slightly obscure the more critical, dynamic signals. These results allow us to confidently identify an optimal range of 10-20 for the CSP.

In summary, this sensitivity analysis validates that both prompt mechanisms operate within a well-defined optimal range, refuting any naive assumption of monotonic improvement with length. For both the instance-specific FBP and the global CSP, expressive capacity is beneficial up to a threshold, beyond which the risk of introducing noise and redundancy leads to a decline in forecasting performance. This underscores the pivotal role of balancing the richness of semantic guidance with architectural conciseness in our model's design.

\section{Additional Disentanglement Visualizations}
\label{sec:appendix_viz}

To supplement the UMAP visualizations presented in the main body, this section provides complementary results from Principal Component Analysis (PCA) and t-Distributed Stochastic Neighbor Embedding (t-SNE). PCA offers a linear perspective on the data's variance, while t-SNE is particularly effective at revealing local clustering structures. These visualizations further test the hypothesis that our semantic guidance mechanism produces a well-structured latent space.

\begin{figure*}[!tbp]
    \centering

    \subfloat[PCA of Component Representations\label{fig:pca_components}]{%
        \resizebox{0.47\textwidth}{!}{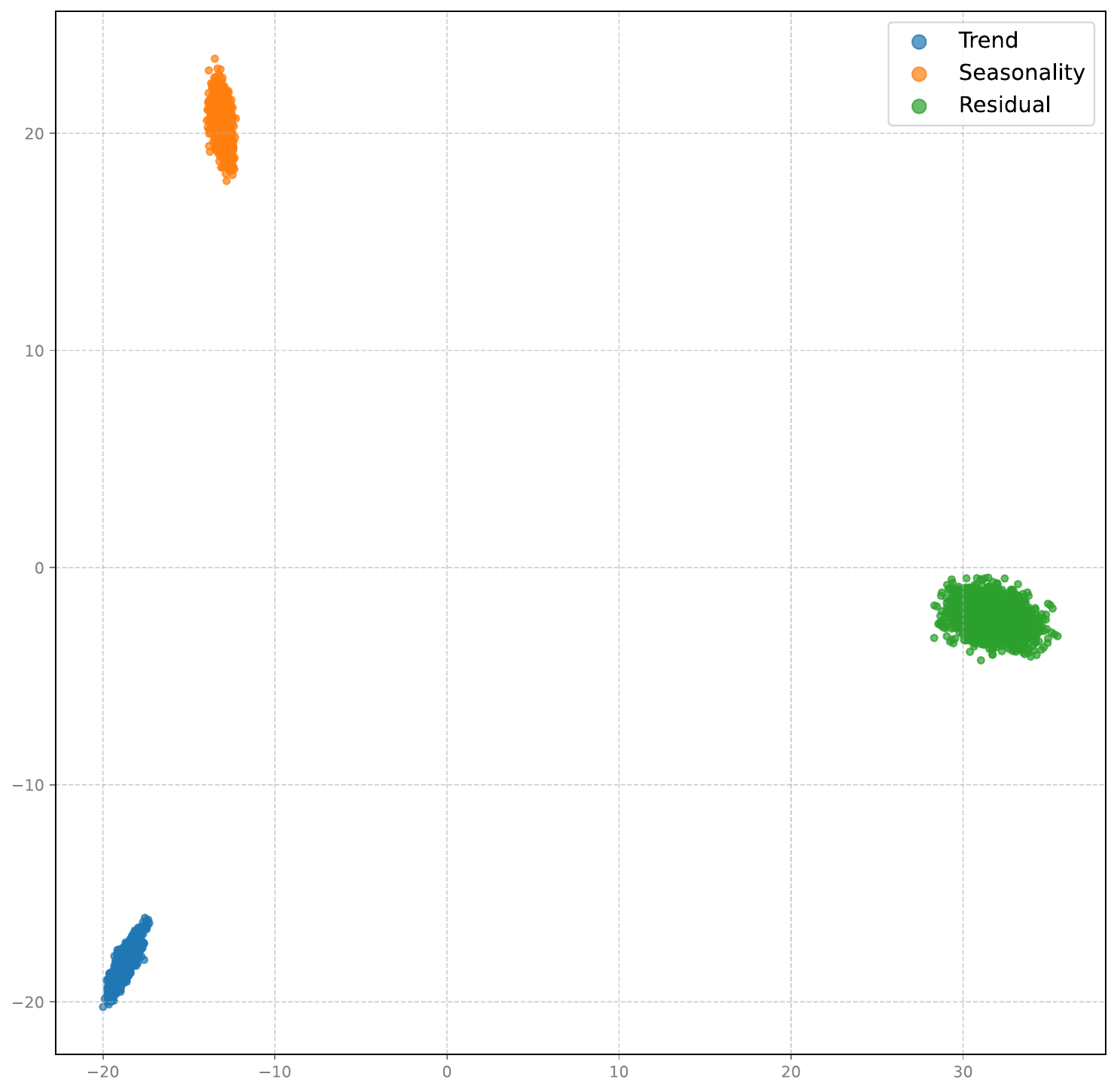}
    }
    \hfill 
    \subfloat[t-SNE of Component Representations\label{fig:tsne_components}]{%
        \resizebox{0.47\textwidth}{!}{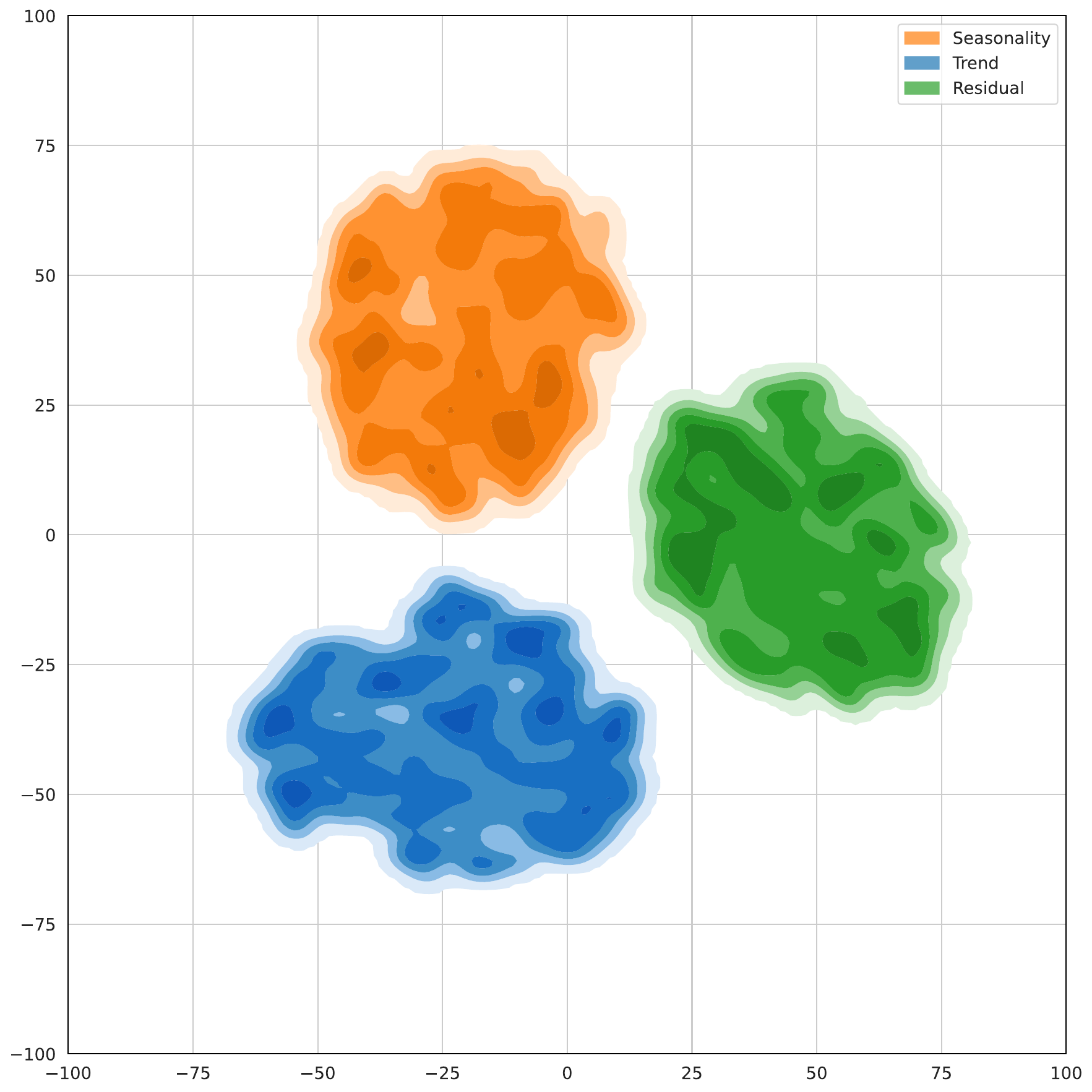}
    }

    \vspace{1.5em} 

    \subfloat[PCA of Hierarchical Semantic Anchors\label{fig:pca_anchors}]{%
        \resizebox{0.47\textwidth}{!}{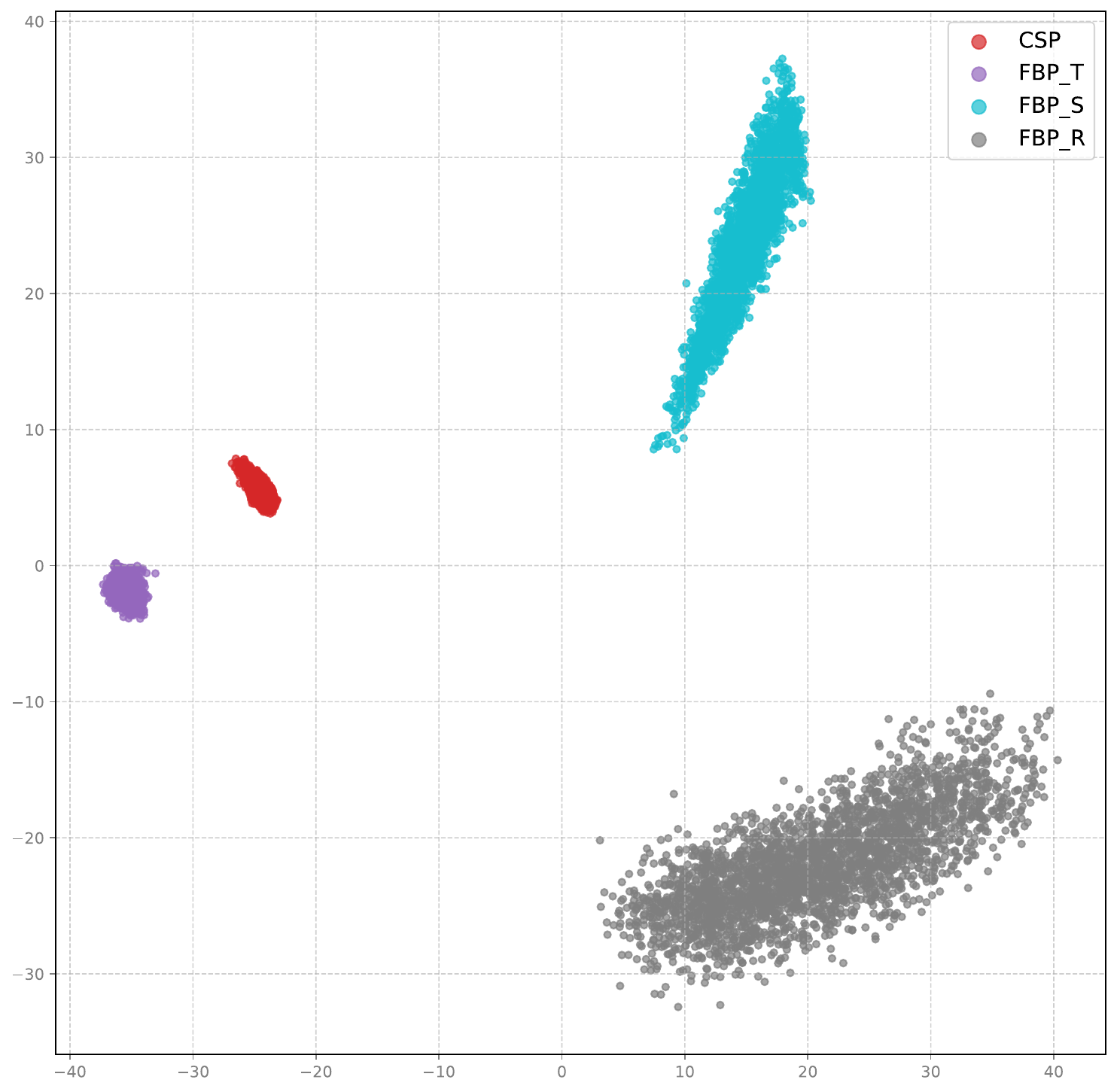}
    }
    \hfill 
    \subfloat[t-SNE of Hierarchical Semantic Anchors\label{fig:tsne_anchors}]{%
        \resizebox{0.47\textwidth}{!}{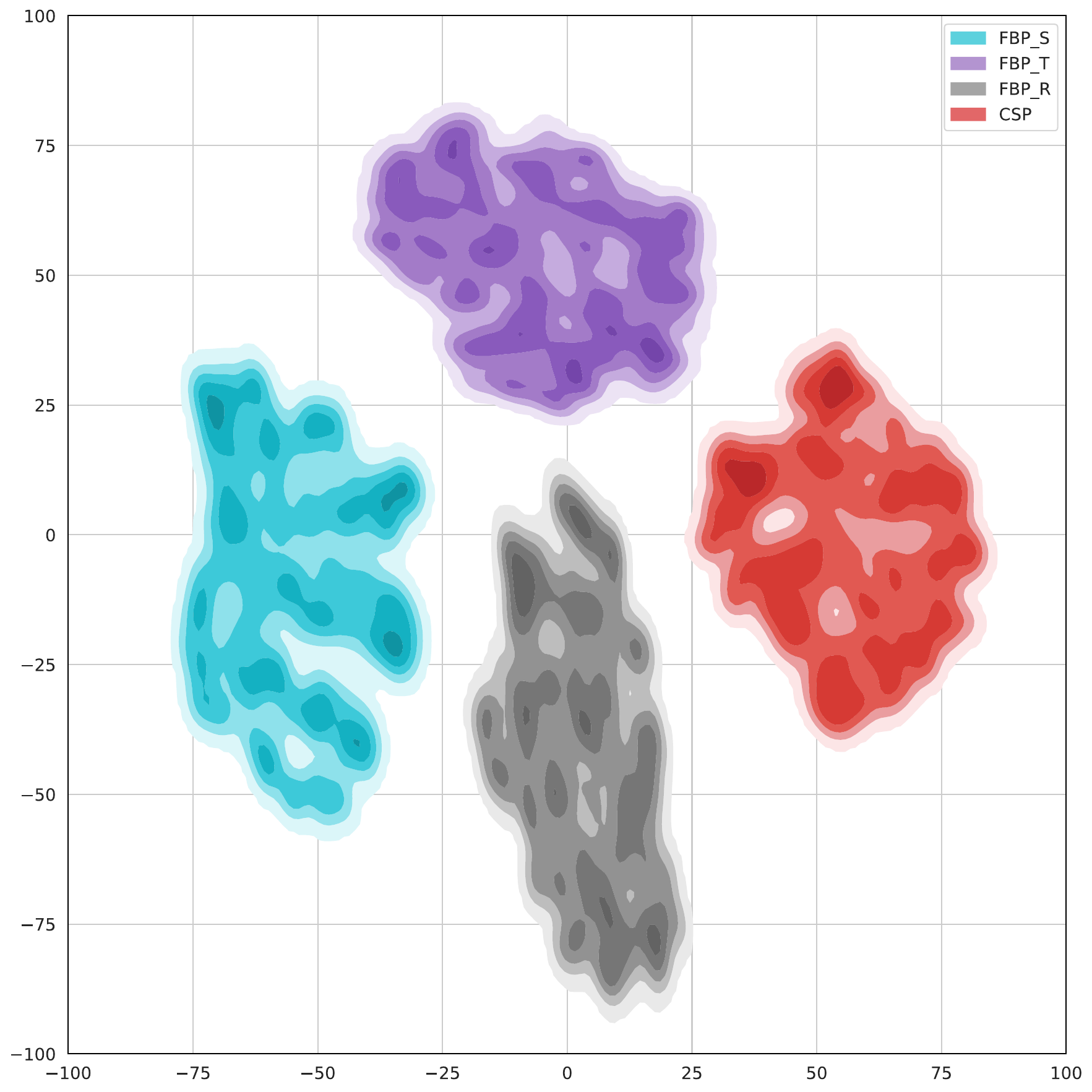}
    }
    
    \caption{Complementary visualizations using PCA and t-SNE further corroborate the semantic-driven disentanglement. (a, c) The linear projections from PCA confirm the separability of both components and their guiding anchors. (b, d) The non-linear manifold learning from t-SNE reveals even tighter and more distinct clustering. These results consistently demonstrate a robust separation in the learned latent space, reinforcing the findings in the main text.}
    \label{fig:appendix_disentanglement_viz}
\end{figure*}

As shown in Figure~\ref{fig:appendix_disentanglement_viz}, both methods confirm the strong separation of learned embeddings. For the final component representations (Figures~\ref{fig:pca_components} and~\ref{fig:tsne_components}), both linear (PCA) and non-linear (t-SNE) projections show distinct clusters. This separation is mirrored in the embeddings of the Hierarchical Semantic Anchors (Figures~\ref{fig:pca_anchors} and~\ref{fig:tsne_anchors}). These observations are consistent with the UMAP results, reinforcing the conclusion that our model achieves robust, semantic-driven disentanglement.

\section{Broader Impacts}
\subsection*{G Broader Impacts}

This work presents the STELLA framework for time series forecasting, aiming to enhance predictive accuracy and generalization. It has potential applications in critical domains such as weather prediction, energy management, and finance, offering tangible societal benefits.At the same time, we acknowledge potential risks, including data privacy concerns and biased predictions that may unfairly impact certain groups.To address these issues, we advocate for transparent model development, ethical data handling, and continuous monitoring during deployment to ensure fairness and accountability.

\balance
\clearpage



\clearpage


\small

\end{document}